%% file: iclr2020_conference.tex
\documentclass{article} 
\usepackage{iclr2020_conference,times}

\input{math_commands.tex}

\usepackage{hyperref}
\usepackage{url}
\usepackage{todonotes}
\usepackage{booktabs}
\usepackage{multirow}
\usepackage{graphicx}
\usepackage{makecell}
\usepackage{pifont}
\usepackage{longtable}
\usepackage{wrapfig}
\usepackage{listings}
\newcommand{\cmark}{\textcolor{teal}{\ding{51}}}%
\newcommand{\xmark}{\textcolor{red}{\ding{55}}}%
\usepackage{caption}    
\usepackage{subcaption}

\lstdefinestyle{jsonstyle}{
    basicstyle=\ttfamily\footnotesize,
    breakatwhitespace=false,         
    breaklines=true,                 
    captionpos=b,                    
    keepspaces=true,                 
    numbers=left,                    
    numbersep=5pt,                  
    showspaces=false,                
    showstringspaces=false,
    showtabs=false,                  
    tabsize=2
}
 
\title{SLM Lab: A Comprehensive Benchmark and Modular Software Framework for Reproducible Deep Reinforcement Learning}

\author{Wah Loon Keng  \\
Mountain View, CA, USA \\
\texttt{kengzwl@gmail.com} \\
\And
Laura Graesser \\
Mountain View, CA, USA \\
\texttt{lhgraesser@gmail.com} \\
\And
Milan Cvitkovic \\
Department of Computing and Mathematical Sciences \\
California Institute of Technology \\
Pasadena, CA, USA \\
\texttt{mcvitkov@caltech.edu}
}

\iclrfinalcopy 
\begin{document}

\maketitle

\begin{abstract}
We introduce SLM Lab, a software framework for reproducible reinforcement learning (RL) research. SLM Lab implements a number of popular RL algorithms, provides synchronous and asynchronous parallel experiment execution, hyperparameter search, and result analysis. RL algorithms in SLM Lab are implemented in a modular way such that differences in algorithm performance can be confidently ascribed to differences between algorithms, not between implementations.  In this work we present the design choices behind SLM Lab and use it to produce a comprehensive single-codebase RL algorithm benchmark.  In addition, as a consequence of SLM Lab's modular design, we introduce and evaluate a discrete-action variant of the Soft Actor-Critic algorithm (Haarnoja et al., 2018) and a hybrid synchronous/asynchronous training method for RL agents.
\end{abstract}

\section{Introduction}

Progress in reinforcement learning (RL) research proceeds only as quickly as researchers can implement new algorithms and publish reproducible empirical results.  But it is no secret that modern RL algorithms are hard to implement correctly \citep{tucker2018the}, and many empirical results are challenging to reproduce \citep{Henderson2017DeepRL, islam-rep, ale2}.  Addressing these problems is aided by providing better software tools to the RL research community.

In this work we introduce SLM Lab,\footnote{\url{https://github.com/kengz/SLM-Lab}} 
a software framework for reinforcement learning research designed for reproducibility and extensibility. SLM Lab is the first open source library that includes algorithm implementations, parallelization, hyperparameter search, and experiment analysis in one framework.

After presenting the design and organization of SLM Lab, we demonstrate the correctness of its implementations by producing a comprehensive performance benchmark of RL algorithms across 77 environments.  To our knowledge this is the largest single-codebase RL algorithm comparison in the literature. In addition, we leverage the modular design of SLM Lab to introduce a variant of the SAC algorithm for use in discrete-action-space environments and a hybrid synchronous/asynchronous training scheme for RL algorithms.


\section{SLM Lab}

\subsection{Library Organization}\label{section:organization}

\begin{figure}[h]
\begin{center}
\includegraphics[width=0.95\linewidth]{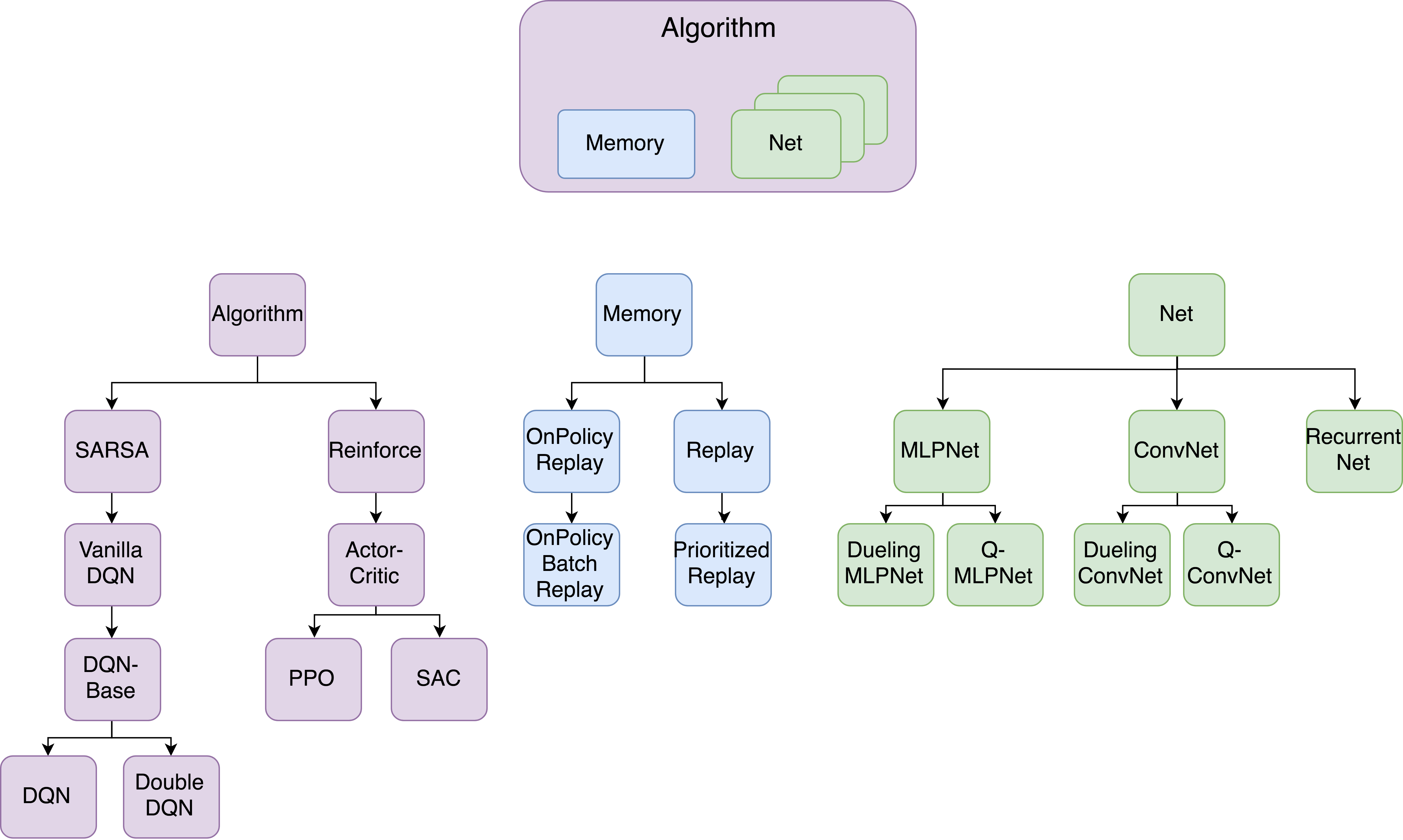}
\end{center}
\caption{SLM Lab classes and their inheritance structure.}
\label{fig:class_diagram}
\end{figure}

Modularity is the central design choice in SLM Lab, as depicted in Figure \ref{fig:class_diagram}. Reinforcement learning algorithms in SLM Lab are built around three base classes:
\begin{itemize}
  \item \texttt{Algorithm}: Handles interaction with the environment, implements an action policy, computes the algorithm-specific loss functions, and runs the training step.
  \item \texttt{Net}: Implements the deep networks that serve as the function approximators for an \texttt{Algorithm}.
  \item \texttt{Memory}: Provides the data storage and retrieval necessary for training.
\end{itemize}

The deep learning components in SLM Lab are implemented using PyTorch \cite{pytorch}. The \texttt{Net} and \texttt{Memory} classes abstract network training details, data storage, and data retrieval, simplifying algorithm implementations. Furthermore, many \texttt{Algorithm} classes are natural extensions of each other.

Modular code is critical for deep RL research because many RL algorithms are extensions of other RL algorithms. If two RL algorithms differ in only a small way, but a researcher compares their performance by running a standalone implementation of each algorithm, they cannot know whether differences in algorithm performance are due to meaningful differences between the algorithms or merely due to quirks in the two implementations. \cite{Henderson2017DeepRL} showcase this, demonstrating significant performance differences between different implementations of the same algorithm. 

Modular code is also important for research progress.  It makes it as simple as possible for a researcher to implement --- and reliably evaluate --- new RL algorithms.  And for the student of RL, modular code is easier to read and learn from due to its brevity and organization into digestible components.

\begin{figure}[h!]
\begin{center}
\includegraphics[width=0.95\linewidth]{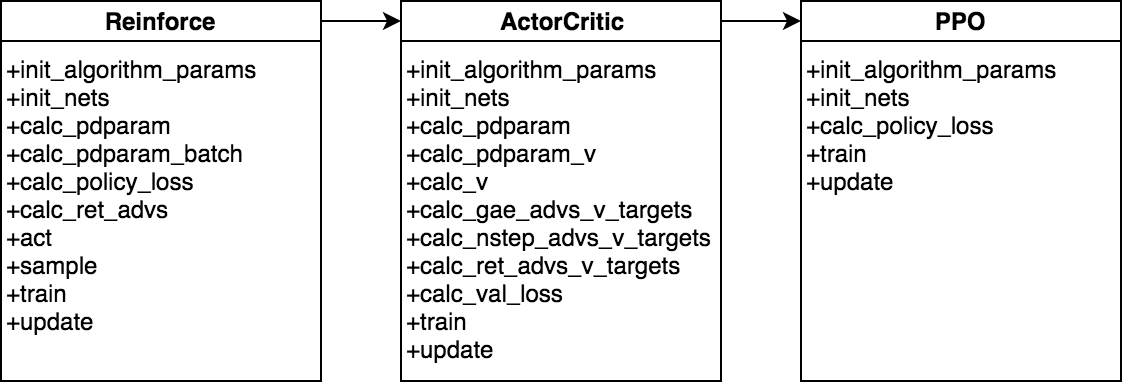}
\end{center}
\caption{\texttt{Reinforce}, \texttt{ActorCritic}, and \texttt{PPO} class methods in SLM Lab. + indicates that a method is added or overridden in the class.}
\label{fig:reinforce_ac_ppo}
\end{figure}

Proximal Policy Optimization (PPO) \citep{ppo} is a good example. When considered as a stand alone algorithm, PPO has a number of different components. However, it differs from the Actor-Critic algorithm only in how it computes the policy loss, runs the training loop, and by needing to maintain an additional actor network during training.  Figure \ref{fig:reinforce_ac_ppo} shows how this similarity is reflected in the SLM Lab implementation of PPO.

The result is that the \verb|PPO| class in SLM Lab has five overridden methods and contains only about 140 lines of code.  Implementing it was straightforward once the \verb|ActorCritic| class was implemented and thoroughly tested.  More importantly, we can be sure that the performance difference between Actor-Critic and PPO observed in experiments using SLM Lab, shown below in Section \ref{sec:results}, are due to something in the 140 lines of code that differentiate \verb|ActorCritic| and \verb|PPO|, and not to other implementation differences.

Another example of modularity in SLM Lab is that, thanks to the consistent API shared between \texttt{Algorithm}, \texttt{Net}, and \texttt{Memory} subclasses, synchronous parallelization using vector environments \citep{baselines} can be combined with asynchronous parallelization of an individual RL agent's learning algorithm.  This multi-level parallelization is further discussed in Section \ref{sec:hybrid}, where we demonstrate its performance benefits.

\subsection{Experiment Organization}\label{sec:EO}

Reinforcement learning algorithms vary greatly in their performance across different environments, hyperparameter settings, and even within a single environment due to inherent randomness. SLM Lab is designed to easily allow users to study all these types of variability.


SLM Lab organizes experiments in the following hierarchy, shown in Figure \ref{fig:experiment_org}:
\begin{figure}[h!]
  \centering
  \begin{subfigure}[b]{0.3\textwidth}
  \includegraphics[width=\textwidth]{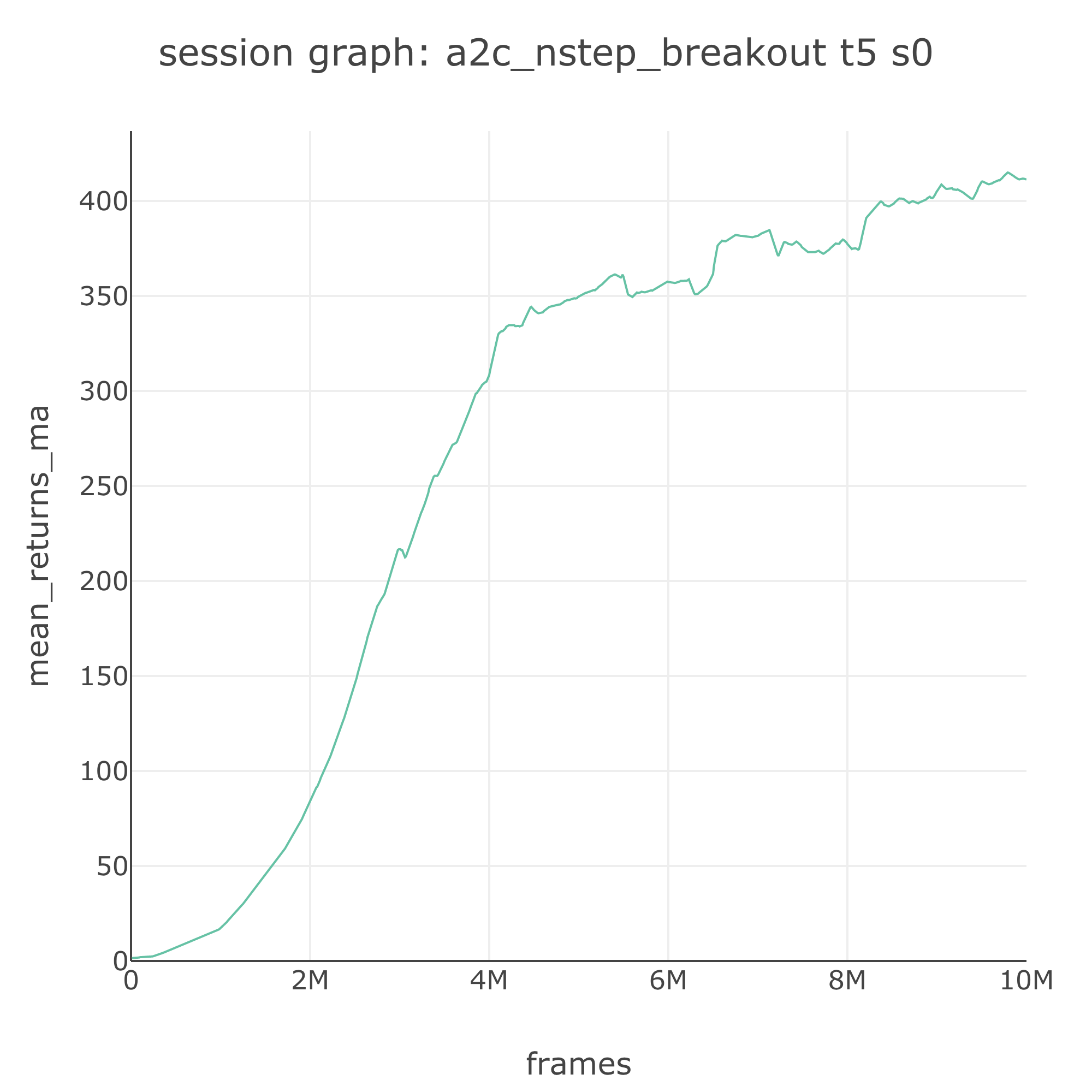}
  \caption{Session}
  \end{subfigure}
  \begin{subfigure}[b]{0.3\textwidth}
  \includegraphics[width=\textwidth]{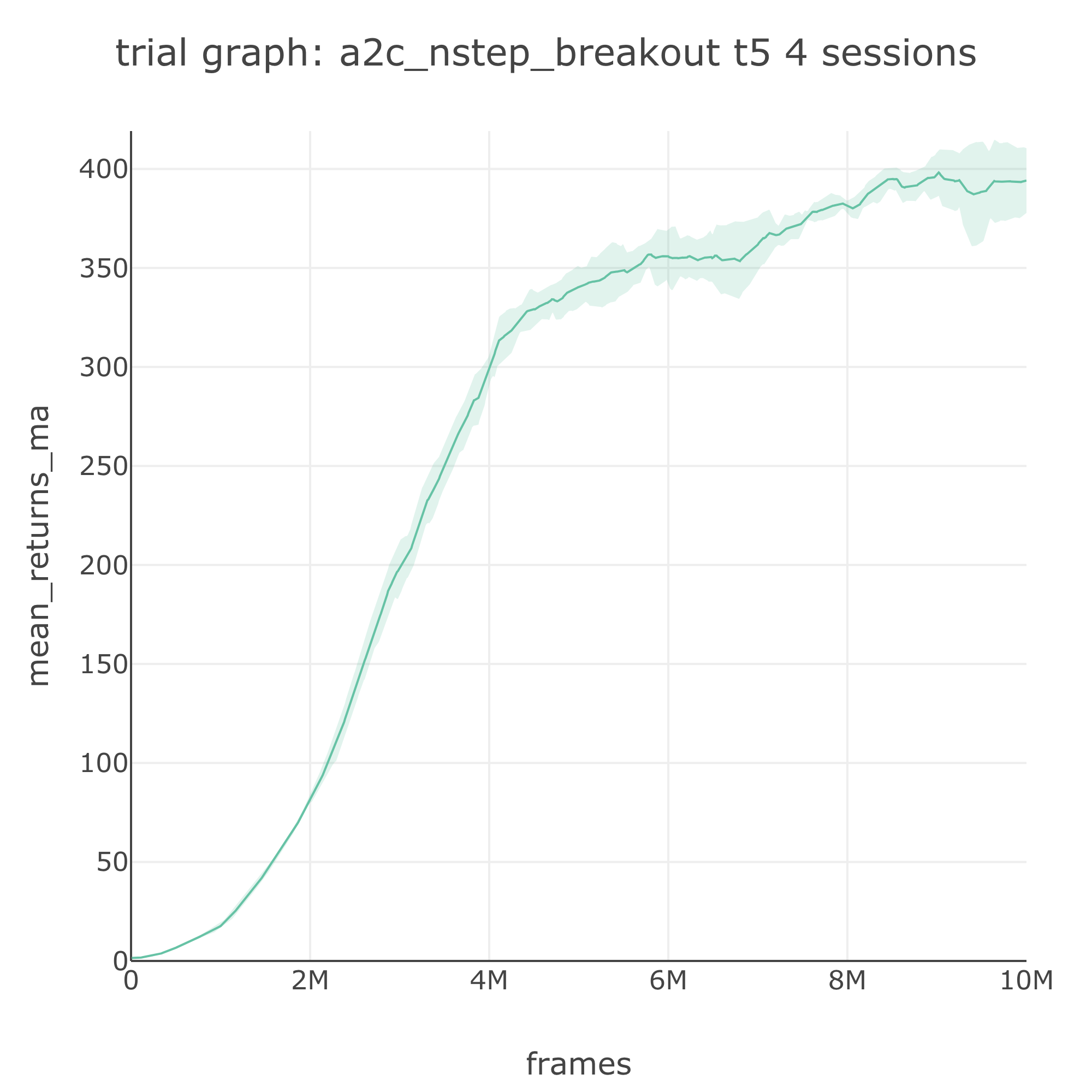}
  \caption{Trial}
  \end{subfigure}
  \begin{subfigure}[b]{0.3\textwidth}
  \includegraphics[width=\textwidth]{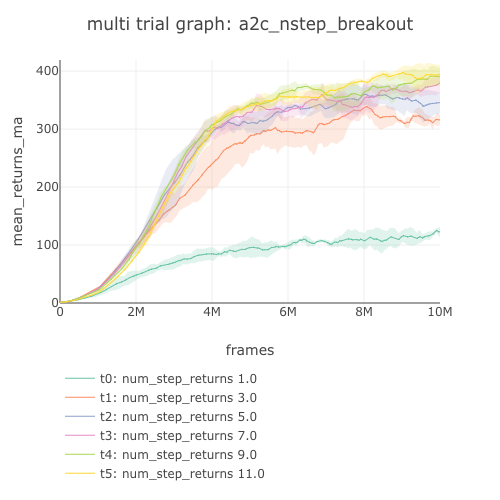}
  \caption{Experiment}
  \end{subfigure}%
 
  \caption{Experiment organization in SLM Lab.  A Session is a single run of an algorithm on an environment.  A Trial is a collection of Sessions.  An Experiment is a collection of Trials with different algorithms and/or environments.}
  \label{fig:experiment_org}
\end{figure}
\begin{enumerate}
    \item \textbf{Session} The lowest level of the SLM Lab experiment framework: a single training run of one agent on one environment with one set of hyperparameters, all with a fixed random seed.
    \item \textbf{Trial} A trial consists of multiple Sessions, with the Sessions varying only in the random seed.
    \item \textbf{Experiment} Generates different sets of hyperparameters (according to a spec file, see below) and runs a Trial for each one. It can be thought of as a study, e.g. ``What values of $n$ of A2C n-step returns provide the fastest, most stable solution, if the other variables are held constant?''
\end{enumerate}

SLM Lab automatically produces plots for Sessions, Trials, and Experiments for any combination of environments and algorithms. It also logs and tracks metrics during training such as rewards, loss, exploration and entropy variables, model weights and biases, action distributions, frames-per-second and wall-clock time. The metrics are also visualized using TensorBoard \citep{tensorflow2015-whitepaper}. Hyperparameter search is implemented using Ray Tune \citep{liaw2018tune}, and the results are automatically analyzed and presented hierarchically in increasingly granular detail.

\subsection{Reproducibility}

The complexity of RL algorithms makes reproducing RL results challenging \citep{Henderson2017DeepRL, ale2}. Every RL researcher knows the difficulty of trying to reproduce an algorithm from its description in a paper alone. Even if code is published along with RL research results, the key algorithm design choices (or mistakes \citep{tucker2018the}) are often buried in the algorithm implementation and not exposed naturally to the user.

In SLM Lab, every configurable hyperparameter for an algorithm is specified in a \emph{spec file}. The spec file is a JSON file containing a git SHA and all the information required to reproduce a Session, Trial, or Experiment as per Section \ref{sec:EO}. Reproducing the entirety of an RL experiment merely requires checking out the code at the git SHA and running the saved spec file.

The main entries in a spec file are given below. Examples of spec files are given in full in Supplementary Section \ref{sec:sup_spec}.
\begin{enumerate}
    \item \textbf{agent -} A list (to allow for multi-agent settings in the future), each element of which contains the spec for a single agent.  Each agent spec contains the details for its components as described in Section \ref{section:organization}:
    \begin{enumerate}
        \item \textbf{algorithm.} The main parameters specific to the algorithm, such as the policy type, exploration strategy (e.g. $\epsilon$-greedy, Boltzmann), algorithm coefficients, rate decays, and training schedules.
        \item \textbf{memory.} Specifies which memory to use as appropriate to the algorithm along with any specific memory hyperparameters such as the batch size and the memory size.
        \item \textbf{net.} The type of neural network, its hidden layer architecture, activations, gradient clipping, loss function, optimizer, rate decays, update method, and CUDA usage.
    \end{enumerate}
    \item \textbf{env -} A list (to allow for multi-environment settings in the future), each element of which specifies an environment.  Each environment spec includes an optional maximum time step per episode, the total time steps (frames) in a Session, the state and reward preprocessing methods, and the number of environments in a vector environment.
    \item  \textbf{body -} Specifies how (multi-)agents connect to (multi-)environments.
    \item \textbf{meta -} The high-level configuration of how the experiment is to be run. It gives the number of Trials and Sessions to run, the evaluation and logging frequency, and a toggle to activate asynchronous training.
    \item \textbf{search -} The hyperparameters to search over and the methods used to sample them. Any variables in the spec file can be searched over, including environment variables.
\end{enumerate}

\section{Results}\label{sec:results}

Reporting benchmark results is essential for validating algorithm implementations. SLM Lab maintains a benchmark page and a public directory containing all of the experiment data, models, plots, and spec files associated with the reported results.\footnote{\url{https://github.com/kengz/SLM-Lab/blob/master/BENCHMARK.md}}
We welcome contributions to this benchmark page via a Pull Request.

We tested the algorithms implemented in SLM Lab on 77 environments: 62 Atari games and 11 Roboschool environments available through OpenAI gym \citep{gym} and 4 Unity environments \citep{unity-ml}. These environments span discrete (Table \ref{discrete-benchmark}) and continuous (Table \ref{continuous-benchmark}) control problems with high- and low-dimensional state spaces.

The results we report in each of the tables are the score per episode at the end of training averaged over the previous 100 training checkpoints. Agents were checkpointed every 10k (Atari) or 1k (Roboschool, Unity) training frames. This measure is less sensitive to rapid increases or decreases in performance, instead reflecting average performance over a substantial number of training frames. 



To our knowledge, the results we present below are a more comprehensive performance comparison than has been previously published for a single codebase.  A full set of learning curves as well as a full table of results for the Atari environments are provided in the supplementary materials.

\subsection{Experiment details}

A complete set of spec files listing all hyperparameters for all algorithms and experiments are included in the \verb|slm_lab/spec/benchmark| directory of SLM Lab as well in the experiment data released along with the results. Example spec files listing all of the hyperparameters for PPO and DDQN + PER on the Atari and Roboschool environments are included in Supplementary Section \ref{sec:sup_spec}.

For the Atari environments, agents were trained for 10M frames (40M accounting for skipped frames). For the Roboschool environments, agents were trained for 2M frames except for RoboschoolHumanoid (50M frames), RoboschoolHumanoidFlagrun (100M) and RoboschoolHumanoidFlagrunHarder (100M). For the Unity environments, agents were trained for 2M frames. Training was parallelized either synchronously or with a hybrid of synchronous and asynchronous methods.


Our results for PPO and A2C on Atari games are comparable the results published by \cite{ppo}. The results on DQN and DDQN + PER on Atari games are mixed: at the same number of training frames\footnote{Estimated using the training curves provided in Figure 7 of \cite{per}} we sometimes match or exceed the reported results and sometimes perform worse. This is likely due to two hyperparameter differences. We used a replay memory of size 200,000 compared to 1M in \cite{mnih2015humanlevel}, \cite{doubledqn}, and \cite{per}. The final output layer of the network is smaller fully-connected layer with 256 instead of 512 units. Finally, our results for SAC confirm the strength of this algorithm compared to PPO for continuous control problems. However the absolute performance is typically worse than the published results from \cite{sac2}. Due to computational constraints, SAC was trained with a replay buffer of 0.2M elements and combined experience replay \citep{cer} compared with 1M elements in \cite{sac2} and this is potentially a significant difference.


\begin{table}[h!]
\caption{Episode score at the end of training attained by SLM Lab implementations on discrete-action control problems. Reported episode scores are the average over the last 100 checkpoints, and then averaged over 4 Sessions. A Random baseline with score averaged over 100 episodes is included. Results marked with `*' were trained using the hybrid synchronous/asynchronous version of SAC to parallelize and speed up training time. For SAC, Breakout, Pong and Seaquest were trained for 2M frames instead of 10M frames.}
\label{discrete-benchmark}
\begin{center}
\begin{tabular}{r|ccccccc}
\toprule
\bf Environment & \multicolumn{7}{c}{\bf Algorithm} \\
  & Random & DQN & DDQN+PER & A2C (GAE) & A2C (n-step) & PPO & SAC \\
\midrule
Breakout & 1.26 & 80.88 & 182 & 377 & 398 & \textbf{443} & 3.51* \\
Pong & -20.4 & 18.48 & 20.5 & 19.31 & 19.56 & \textbf{20.58} & 19.87* \\
Seaquest & 106 & 1185 & \textbf{4405} & 1070 & 1684 & 1715 & 171* \\
Qbert & 157 & 5494 & 11426 & 12405 & \textbf{13590} & 13460 & 923* \\
LunarLander & -162 & 192 & 233 & 25.21 & 68.23 & 214 & \textbf{276} \\
UnityHallway & -0.99 & -0.32 & 0.27 & 0.08 & -0.96 & \textbf{0.73} & 0.01 \\
UnityPushBlock & -1.00 & 4.88 & 4.93 & 4.68 & 4.93 & \textbf{4.97} & -0.70 \\
\bottomrule
\end{tabular}
\end{center}
\end{table}

\begin{table}[h!]
\caption{Episode score at the end of training attained by SLM Lab implementations on continuous control problems. Reported episode scores are the average over the last 100 checkpoints, and then averaged over 4 Sessions. A Random baseline with score averaged over 100 episodes is included. Results marked with `*' require 50M-100M frames, so we use the hybrid synchronous/asynchronous version of SAC to parallelize and speed up training time.}
\label{continuous-benchmark}
\begin{center}
\begin{tabular}{r|ccccc}
\toprule
\bf Environment & \multicolumn{5}{c}{\bf Algorithm} \\
 & Random & A2C (GAE) & A2C (n-step) & PPO & SAC \\
\midrule
\multirow{8}{*}{\rotatebox{90}{}} RoboschoolAnt & 52.30 & 787 & 1396 & 1843 & \textbf{2915} \\
RoboschoolAtlasForwardWalk & 40.23 & 59.87 & 88.04 & 172 & \textbf{800} \\
RoboschoolHalfCheetah & 0.83 & 712 & 439 & 1960 & \textbf{2497} \\
RoboschoolHopper & 21.22 & 710 & 285 & 2042 & \textbf{2045} \\
RoboschoolInvertedDoublePendulum & 285 & 996 & 4410 & 8076 & \textbf{8085} \\
RoboschoolInvertedPendulum & 23.5 & \textbf{995} & 978 & 986 & 941 \\
RoboschoolReacher & -9.24 & 12.9 & 10.16 & 19.51 & \textbf{19.99} \\
RoboschoolWalker2d & 16.06 & 280 & 220 & 1660 & \textbf{1894} \\
RoboschoolHumanoid & -36.16 & 99.31 & 54.58 & 2388 & \textbf{2621}* \\
RoboschoolHumanoidFlagrun & -6.15 & 73.57 & 178 & 2014 & \textbf{2056}* \\
RoboschoolHumanoidFlagrunHarder & -8.64 & -429 & 253 & \textbf{680} & 280* \\
Unity3DBall & 1.24 & 33.48 & 53.46 & 78.24 & \textbf{98.44} \\
Unity3DBallHard & 1.17 & 62.92 & 71.92 & 91.41 & \textbf{97.06} \\
\bottomrule
\end{tabular}
\end{center}
\end{table}

\subsection{Soft Actor-Critic for discrete environments}

All published results for the Soft Actor-Critic (SAC) algorithm \citep{sac1, sac2} are for continuous control environments.  However, nothing in its algorithmic description makes it unsuitable in principle for use in discrete action-space environments.  As a consequence of the modular structure of SLM Lab, it was straightforward to design and implement a discrete variant of SAC.  We did so by using policy \texttt{Net}s that produced parameters of a Gumbel-Softmax distribution \citep{gumbel-softmax, cont-relax} from which discrete actions were sampled.  The results of this discrete variant of SAC are in Table \ref{discrete-benchmark}.

On environments for which its training converged, we found SAC to be of comparable or better sample efficiency than all of the other algorithms that we tested.
For example, SAC achieves an average score over the previous 100 checkpoints of around 20 after 1M frames on Atari Pong, whereas all other algorithms that we tested require at least 2M frames to achieve the same result (plot shown in Supplementary Section \ref{sec:sup-atari-bench}). 
However, even though SAC trained successfully on Pong and Lunar Lander, we were not able to successfully train it on all other Atari environments. We also note that while SAC is sample efficient it is more computationally expensive than the other algorithms, which presents an obstacle for extensive performance tuning.

\subsection{Hybrid synchronous and asynchronous training}\label{sec:hybrid}

\begin{figure}[h]
\begin{center}
\includegraphics[width=0.95\linewidth]{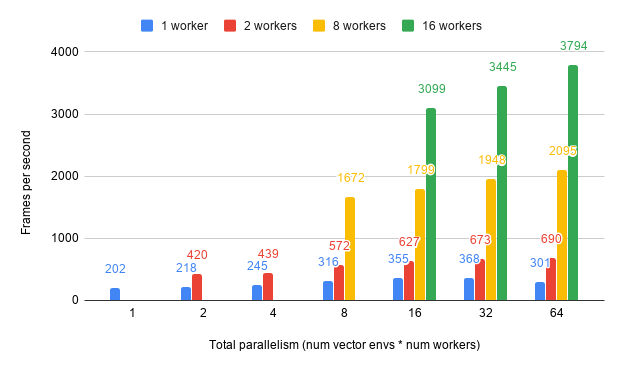}
\end{center}
\caption{Average frames per second as number of vector environments and Hogwild! workers are varied. Each setting was trained using PPO with the same hyperparameters on the RoboschoolAnt environment.}
\label{fig:fps}
\end{figure}

Synchronous and asynchronous parallelization can be combined in SLM Lab to accelerate training, as shown in Figure \ref{fig:fps}.  SLM Lab implements synchronous parallelization within Sessions (Section \ref{sec:EO}) using vector environments \citep{baselines} and asynchronous parallelization within Trials (Section \ref{sec:EO}) using multiple workers, one per Session. There are two available methods for Trial level parallelization; Hogwild!\ \citep{hogwild} or a server-worker model in which workers periodically push gradients to a central network and pull copies of the updated parameters.

If training is constrained by data sampling from the environment, then increasing the number of vector environments (synchronous parallelization) speeds up training. But this speed-up saturates as the training step becomes a bottleneck and the environment waits for the agent to train. In the example shown in Figure \ref{fig:fps} the frames per seconds (fps) increases from around 200 for a single worker and 1 environment to around 360 for 1 worker and 16 environments, and saturates thereafter. Once fps becomes constrained by the training step, it is beneficial to add workers (asynchronous parallelization) to effectively parallelize the parameter updates. A hybrid of 16 workers each with 4 environments resulted in the maximum fps of around 3800.

\section{Related Work}

\subsection{Reproducibility in Reinforcement Learning}

The instability of RL algorithms \citep{sac1}, randomness in agent policies and the environment \citep{Henderson2017DeepRL, islam-rep}, as well as differences in hyperparameter tuning \citep{islam-rep} and implementations \citep{Henderson2017DeepRL, tucker2018the} all contribute to the challenge of reproducing RL results.

Consequently, the importance of comprehensively documenting all hyperparameters along with published results and software is well recognized within the research community \citep{ale2, dopamine}.









\subsection{Software for Reinforcement Learning}

To date more than twenty reinforcement-learning-themed open source software libraries have been released. They can be organized into two categories: those implementing RL algorithms and those implement RL environments.\footnote{A few libraries such as OpenSpiel~\citep{openspiel} and ELF~\citep{facebook-elf} implement both.} SLM Lab is an algorithm-focused library with built-in integration with the OpenAI gym \citep{gym}, OpenAI Roboschool, VizDoom \citep{vizdoom}, and Unity ML-Agents \citep{unity-ml} environment libraries.




Table \ref{lib-comp} summarizes the algorithm-focused reinforcement learning software libraries.

\begin{table}[h!]
\caption{Comparison of RL software libraries.  Algorithm acronyms are explained in Supplementary Section A.1. REINFORCE is excluded as are less well-known algorithms. ``Benchmark'' indicates whether the library reports the performance of their implementations. ``Config'' indicates whether hyperparameters are specified separately from the implementation and run scripts; ``split'' indicates that the configuration is divided across multiple files, ``partial'' indicates that some but not all hyperparameters are included. ``Parallel'' denotes whether training for any algorithms can be parallelized. ``HPO'' denotes support for hyperparameter optimization.  ``Plot'' denotes whether the library provides any methods for visualizing results. }
\label{lib-comp}
\begin{center}
\begin{tabular}[t]{r|p{4.5cm}ccccc}
\toprule
\bf Library & \multicolumn{1}{c}{\bf Algorithms} & \bf Benchmark & \bf Config & \bf Parallel & \bf HPO & \bf Plot   \\ \midrule
a2c-ppo-acktr-gail & A2C, ACKTR, GAIL, PPO &\cmark & \xmark & \cmark & \xmark & \cmark \\
Baselines & A2C, ACER, ACKTR, DDPG, DQN, GAIL, HER, PPO, TRPO & \cmark & \xmark & \cmark & \xmark & \cmark \\
Catalyst & CRA, TD3, SAC, DDPG, QR & \cmark & \cmark & \cmark & \xmark & \cmark \\
ChainerRL & A3C, ACER, C51, DDPG, DQN+, IQN, NSQ, PPO, Rainbow, TRPO, TD3, SAC & \cmark & \xmark & \cmark & \xmark & \cmark \\
coach & A3C, ACER, C51, DDPG, DQN+, QR-DQN, NAF, NEC, NSQ, PPO, Rainbow, SAC, TD3 & \cmark & split & \cmark & \xmark & \cmark \\
DeepRL & A2C, C51, DDPG, DQN+, NSQ, OC, PPO, QR-DQN, TD3 & \cmark & split & \xmark & \xmark & \cmark \\
Dopamine & DQN, C51, Rainbow & \cmark & split & \cmark & \xmark & \cmark \\
ELF & A3C, DQN, MCTS, TRPO & \cmark & \xmark & \cmark & \xmark & \xmark \\
Keras-RL & DDPG, DQN+, NAF, CEM, SARSA & \xmark & \xmark & \cmark & \xmark & \cmark \\
Horizon & C51, DQN+, SAC, TD3 & \xmark & \cmark & \cmark & \xmark & \cmark \\
MAgent & A2C, DQN & \xmark & \xmark & \cmark & \xmark & \cmark \\
OpenSpiel & A2C, DQN, MCTS & \xmark & \xmark & \xmark & \xmark & \xmark\\
pytorch-rl & A2C, DDPG, DQN+, HER & \xmark & \xmark & \xmark & \xmark & \xmark \\
reaver & A2C, PPO & \cmark & split & \cmark & \xmark & \cmark \\
RLgraph & A2C, Ape-X, DQN+, DQFD, IMPALA, PPO, SAC & \cmark & \cmark & \cmark & \xmark & \xmark \\
RLkit & DQN+, HER, SAC, TDM, TD3 & \xmark & \xmark & \xmark & \xmark & \cmark \\
RLLib & A3C, Ape-X, ARS, DDPG, DQN, ES, IMPALA, PPO, SAC & \cmark & partial & \cmark & \xmark & \xmark \\
rlpyt & A2C, DDPG, DQN+, CAT-DQN, PPO, TD3, SAC & \cmark & \xmark & \cmark & \xmark & \xmark \\
Softlearning & SAC & \xmark & \cmark & \cmark & \xmark & \xmark \\
Stable Baselines & A2C, ACER, ACKTR, DDPG, DQN, GAIL, HER, PPO, SAC, TD3, TRPO & \cmark & \xmark & \cmark & \cmark & \cmark \\
TensorForce & A3C, DQN+, DQFD, NAF, PPO, TRPO & \xmark & \xmark & \cmark & \cmark & \xmark \\
TF-Agents & DDPG, DQN+, PPO, SAC, TD3 & \xmark & \xmark & \xmark & \xmark & \cmark \\
TRFL & - & \xmark & \xmark & \xmark & \xmark & \xmark \\
vel & A2C, ACER, DDPG, DQN+, Rainbow, PPO, TRPO & \cmark & split & \cmark & \xmark & \xmark \\
\midrule
SLM Lab & A2C, A3C, CER, DQN+, PPO, SAC & \cmark & \cmark & \cmark & \cmark & \cmark \\
\bottomrule
\end{tabular}
\end{center}
\end{table}

Libraries such as Catalyst \citep{catalyst}, ChainerRL \citep{chainerrl, chainer}, coach \citep{coach}, DeepRL \citep{sz-deeprl}, OpenAI baselines \citep{baselines}, RLgraph \citep{rlgraph}, RLkit \citep{rlkit}, rlpyt \citep{rlpyt}, RLLib \citep{ray-rllib}, Stable Baselines \citep{stable-baselines}, TensorForce \citep{tensorforce}, TF-Agents \citep{TFAgents}, and vel \citep{vel} implement a wide variety of algorithms and are intended to be applied to a variety of RL problems. Most of these libraries also provide some benchmark results for the implemented algorithms to validate their performance. These can be thought of as \textit{generalist} RL libraries and are the most closely related to SLM Lab.

Other libraries focus on specific algorithms (Dopamine \citep{dopamine}, Softlearning \citep{softlearn}, a2c-ppo-acktr-gail \citep{kostrikov}, Keras-RL \citep{kerasrl}), problems (OpenSpiel \citep{openspiel}, ELF \citep{facebook-elf}, MAgent \citep{magent}, reaver \citep{reaver}), components such as loss functions (TRFL \citep{deepmind-trfl} or scaling training (Horizon \citep{facebook-horizon}).

The use of configuration files to specify hyperparameters varies. Catalyst, coach, DeepRL, Dopamine, Horizon, reaver, Softlearning, RLgraph, RLLib, and vel use configuration files and provide a number of configured examples. In most cases the network architecture is excluded from the main configuration file and specified elsewhere. Catalyst, Horizon and RLgraph include all hyperparameters in a single configuration file and are the most similar to SLM Lab. However, RLgraph does not include environment information, and none use the same file for configuring hyperparameter search.

Almost all libraries include some methods for parallelizing agent training, especially for on-policy methods. However most do not include hyperparameter optimization as a feature. The two exceptions are Stable Baselines \citep{stable-baselines} and Tensorforce \citep{tensorforce}.

Finally, many libraries include some tools for visualizing and plotting results. Notably coach \citep{coach} provides an interactive dashboard for exploring a variety of metrics which are automatically tracked during training.

\bibliography{iclr2020_conference}
\bibliographystyle{iclr2020_conference}

\appendix
\section{Supplementary Materials}

\subsection{Algorithm acronyms}\label{sec:sup-ac}

The expanded acronymns for all of the algorithms listed in Table \ref{lib-comp} are given below:
\begin{itemize}
    \item A2C: Advantage Actor-Critic
    \item ACER: Actor-Critic with Experience Replay
    \item ACKTR: Actor-Critic using Kronecker-Factored Trust Region
    \item Ape-X: Distributed Prioritized Experience Replay
    \item ARS: Augmented Random Search
    \item C51: Categorical DQN
    \item CAT-DQN: Categorical DQN
    \item CRA: Categorical Return Approximation
    \item CEM: Cross Entropy Method
    \item CER: Combined Experience Replay
    \item DDPG: Deep Deterministic Policy Gradients
    \item DQN: Deep Q Networks
    \item DQN+: Deep Q Network modifications, including some or all of the following: Double DQN, Dueling DQN, Prioritized Experience Replay
    \item DQFD: Deep Q-Learning from Demonstrations
    \item ES: Evolutionary Strategies
    \item GAIL: Generative Adversarial Imitation learning
    \item HER: Hindsight Experience Replay
    \item IMPALA: Importance Weighted Actor-Learner Architecture
    \item IQN: Implicit Quantile Networks
    \item MCTS: Monte Carlo Tree Search
    \item NAF: Normalized Advantage Functions
    \item NEC: Neural Episodic Control
    \item NSQ: n-step Q-learning
    \item OC: Option-Critic
    \item PPO: Proximal Policy Optimization
    \item QR: Quantile Regression
    \item QR-DQN: Quantile Regression DQN
    \item TD3: Twin Delayed Deep Deterministic Policy Gradient
    \item TRPO: Trust Region Policy Optimization
    \item SAC: Soft Actor-Critic
\end{itemize}

\subsection{Atari results}

The table below presents results for 62 Atari games. All agents were trained for 10M frames (40M including skipped frames). Reported results are the episode score at the end of training, averaged over the previous 100 evaluation checkpoints with each checkpoint averaged over 4 Sessions. Agents were checkpointed every 10k training frames. A Random baseline with score averaged over 100 episodes is included.

\begin{longtable}{r|cccccc}
\toprule
\bf Environment & \multicolumn{6}{c}{\bf Algorithm} \\
\bf (Atari, Discrete) & Random & DQN & DDQN+PER & A2C (GAE) & A2C (n-step) & PPO \\
\midrule
Adventure & -0.89 & -0.94 & -0.92 & -0.77 & -0.85 & \textbf{-0.3} \\
AirRaid & 486 & 1876 & 3974 & \textbf{4202} & 3557 & 4028 \\
Alien & 97.0 & 822 & 1574 & 1519 & \textbf{1627} & 1413 \\
Amidar & 1.8 & 90.95 & 431 & 577 & 418 & \textbf{795} \\
Assault & 308 & 1392 & 2567 & 3366 & 3312 & \textbf{3619} \\
Asterix & 307 & 1253 & \textbf{6866} & 5559 & 5223 & 6132 \\
Asteroids & 1331 & 439 & 426 & \textbf{2951} & 2147 & 2186 \\
Atlantis & 29473 & 68679 & 644810 & \textbf{2747371} & 2259733 & 2148077 \\
BankHeist & 13.4 & 131 & 623 & 855 & 1170 & \textbf{1183} \\
BattleZone & 3080 & 6564 & 6395 & 4336 & 4533 & \textbf{13649} \\
BeamRider & 355 & 2799 & \textbf{5870} & 2659 & 4139 & 4299 \\
Berzerk & 212 & 319 & 401 & \textbf{1073} & 763 & 860 \\
Bowling & 24.14 & 30.29 & \textbf{39.5} & 24.51 & 23.75 & 31.64 \\
Boxing & -0.91 & 72.11 & 90.98 & 1.57 & 1.26 & \textbf{96.53} \\
Breakout & 1.26 & 80.88 & 182 & 377 & 398 & \textbf{443} \\
Carnival & 905 & 4280 & \textbf{4773} & 2473 & 1827 & 4566 \\
Centipede & 2888 & 1899 & 2153 & 3909 & 4202 & \textbf{5003} \\
ChopperCommand & 735 & 1083 & \textbf{4020} & 3043 & 1280 & 3357 \\
CrazyClimber & 2452 & 46984 & 88814 & 106256 & 109998 & \textbf{116820} \\
Defender & 546810 & 281999 & 313018 & \textbf{665609} & 657823 & 534639 \\
DemonAttack & 346 & 1705 & 19856 & 23779 & 19615 & \textbf{121172} \\
DoubleDunk & -16.48 & -21.44 & -22.38 & \textbf{-5.15} & -13.3 & -6.01 \\
ElevatorAction & 9851 & 32.62 & 17.91 & \textbf{9966} & 8818 & 6471 \\
Enduro & 0.0 & 437 & 959 & 787 & 0.0 & \textbf{1926} \\
FishingDerby & -93.96 & -88.14 & -1.7 & 16.54 & 1.65 & \textbf{36.03} \\
Freeway & 0.0 & 24.46 & 30.49 & 30.97 & 0.0 & \textbf{32.11} \\
Frostbite & 68.0 & 98.8 & \textbf{2497} & 277 & 261 & 1062 \\
Gopher & 276 & 1095 & \textbf{7562} & 929 & 1545 & 2933 \\
Gravitar & 219 & 87.34 & 258 & 313 & \textbf{433} & 223 \\
Hero & 706 & 1051 & 12579 & 16502 & \textbf{19322} & 17412 \\
IceHockey & -9.87 & -14.96 & -14.24 & \textbf{-5.79} & -6.06 & -6.43 \\
Jamesbond & 13.0 & 44.87 & \textbf{702} & 521 & 453 & 561 \\
JourneyEscape & -18095 & -4818 & -2003 & \textbf{-921} & -2032 & -1094 \\
Kangaroo & 54.0 & 1965 & \textbf{8897} & 67.62 & 554 & 4989 \\
Krull & 1747 & 5522 & 6650 & 7785 & 6642 & \textbf{8477} \\
KungFuMaster & 865 & 2288 & 16547 & 31199 & 25554 & \textbf{34523} \\
MontezumaRevenge & 0.0 & 0.0 & 0.02 & 0.08 & 0.19 & \textbf{1.08} \\
MsPacman & 170 & 1175 & 2215 & 1965 & 2158 & \textbf{2350} \\
NameThisGame & 2088 & 3915 & 4474 & 5178 & 5795 & \textbf{6386} \\
Phoenix & 1324 & 2909 & 8179 & 16345 & 13586 & \textbf{30504} \\
Pitfall & -301 & -68.83 & -73.65 & -101 & \textbf{-31.13} & -35.93 \\
Pong & -20.4 & 18.48 & 20.5 & 19.31 & 19.56 & \textbf{20.58} \\
Pooyan & 515 & 1958 & 2741 & 2862 & 2531 & \textbf{6799} \\
PrivateEye & -731 & \textbf{784} & 303 & 93.22 & 78.07 & 50.12 \\
Qbert & 157 & 5494 & 11426 & 12405 & \textbf{13590} & 13460 \\
Riverraid & 1554 & 953 & \textbf{10492} & 8308 & 7565 & 9636 \\
RoadRunner & 35.0 & 15237 & 29047 & 30152 & 31030 & \textbf{32956} \\
Robotank & 1.78 & 3.43 & \textbf{9.05} & 2.98 & 2.27 & 2.27 \\
Seaquest & 106 & 1185 & \textbf{4405} & 1070 & 1684 & 1715 \\
Skiing & -17361 & -14094 & \textbf{-12883} & -19481 & -14234 & -24713 \\
Solaris & 2097 & 612 & 1396 & 2115 & \textbf{2236} & 1892 \\
SpaceInvaders & 164 & 451 & 670 & 733 & 750 & \textbf{797} \\
StarGunner & 645 & 3565 & 38238 & 44816 & 48410 & \textbf{60579} \\
Tennis & -24.0 & -23.78 & \textbf{-10.33} & -22.42 & -19.06 & -11.52 \\
TimePilot & 3151 & 2819 & 1884 & 3331 & 3440 & \textbf{4398} \\
Tutankham & 15.45 & 35.03 & 159 & 161 & 175 & \textbf{211} \\
UpNDown & 125 & 2043 & 11632 & 89769 & 18878 & \textbf{262208} \\
Venture & 0.0 & 4.56 & 9.61 & 0.0 & 0.0 & \textbf{11.84} \\
VideoPinball & 25365 & 8056 & \textbf{79730} & 35371 & 40423 & 58096 \\
WizardOfWor & 784 & 869 & 328 & 1516 & 1247 & \textbf{4283} \\
YarsRevenge & 3369 & 5816 & 15698 & \textbf{27097} & 11742 & 10114 \\
Zaxxon & 0.0 & 442 & 54.28 & 64.72 & 24.7 & \textbf{641} \\
\bottomrule
\end{longtable}

\subsection{Learning curves for all algorithms and environments}\label{sec:sup-atari-bench}

Each learning curve depicts the episode score, averaged over the previous 100 evaluation checkpoints, with each checkpoint averaged over 4 Sessions. The shaded area depicts +/- one standard deviation calculated over 4 Sessions. 

\begin{longtable}{cccc}
\includegraphics[width=1.22in]{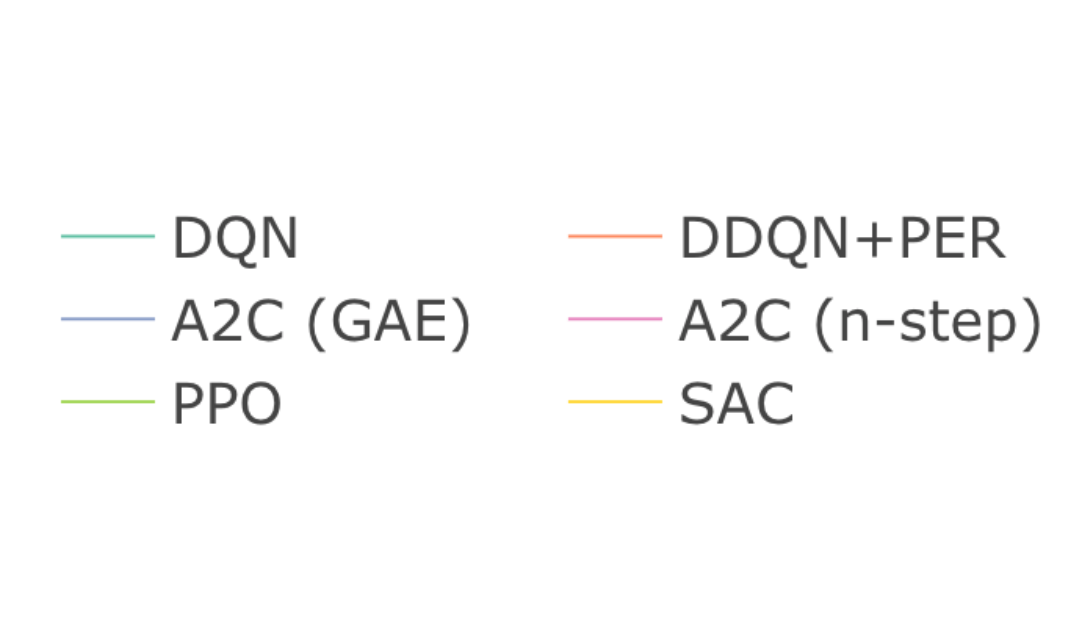} &
\includegraphics[width=1.22in]{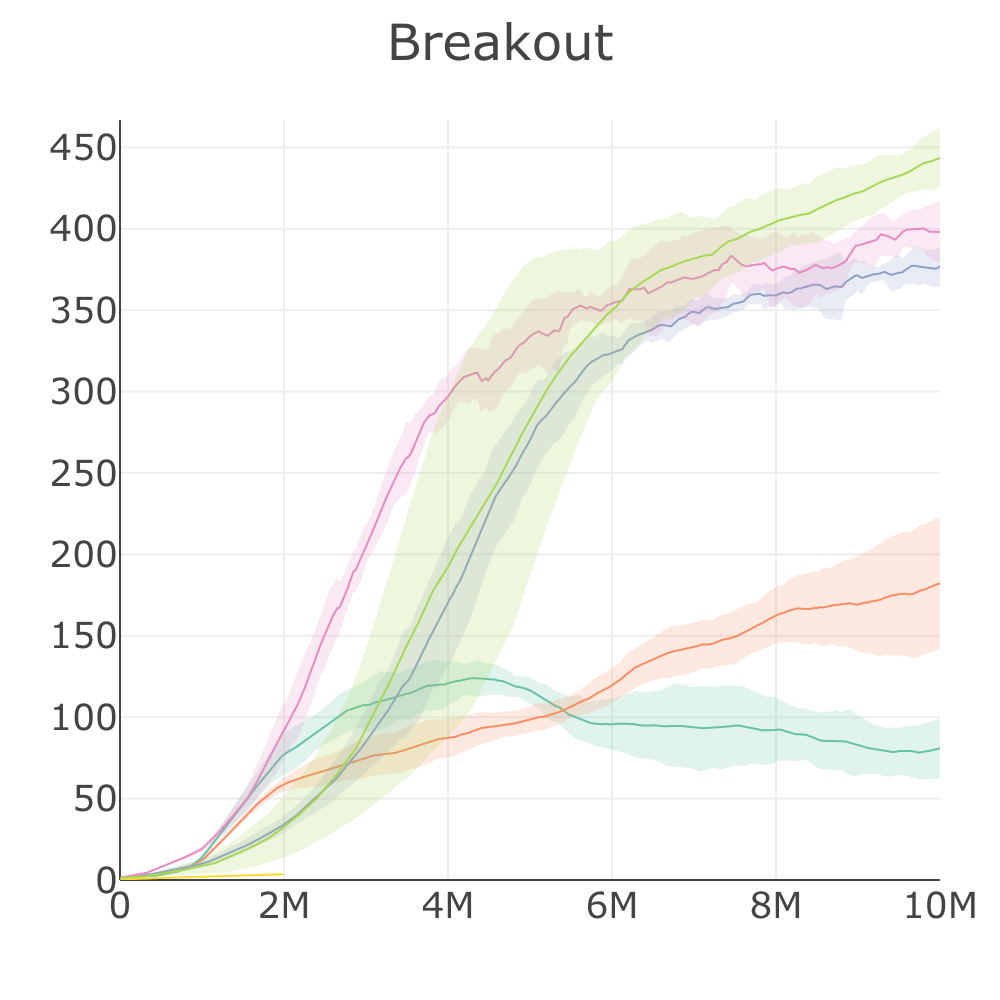} & 
\includegraphics[width=1.22in]{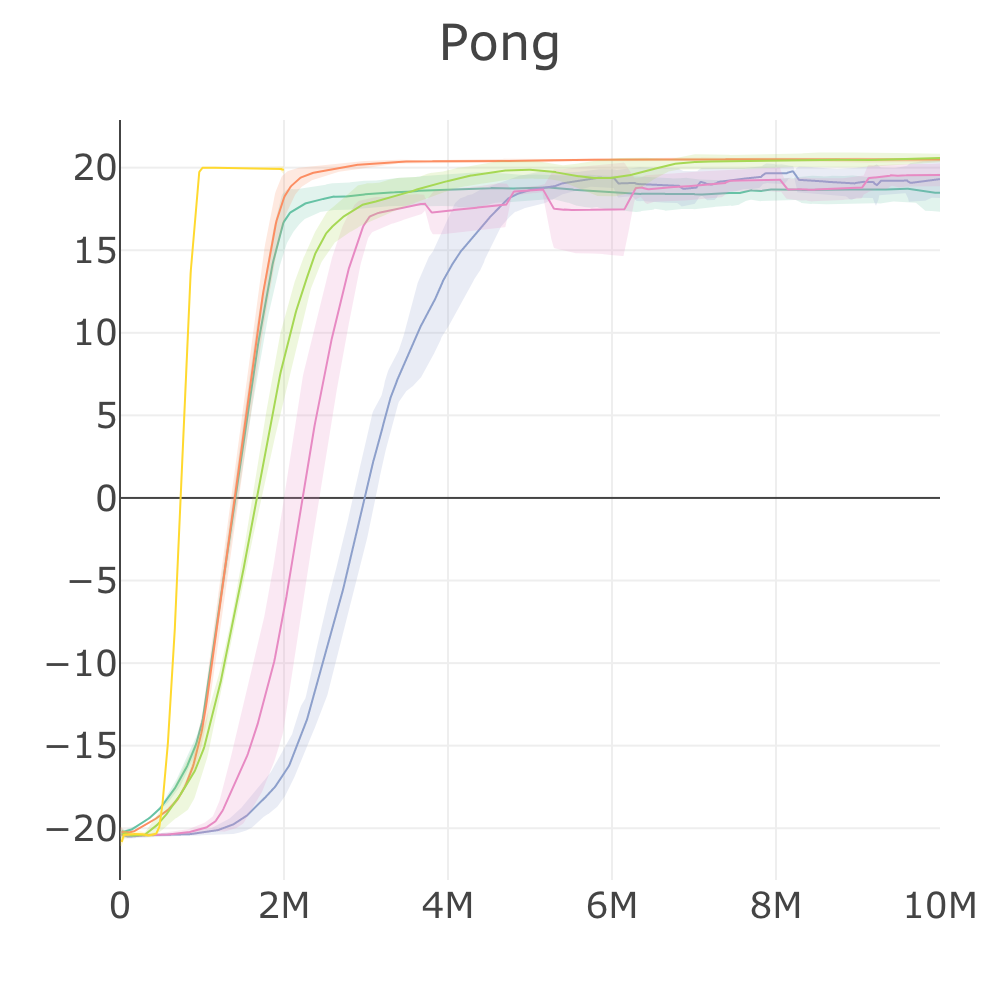} & 
\includegraphics[width=1.22in]{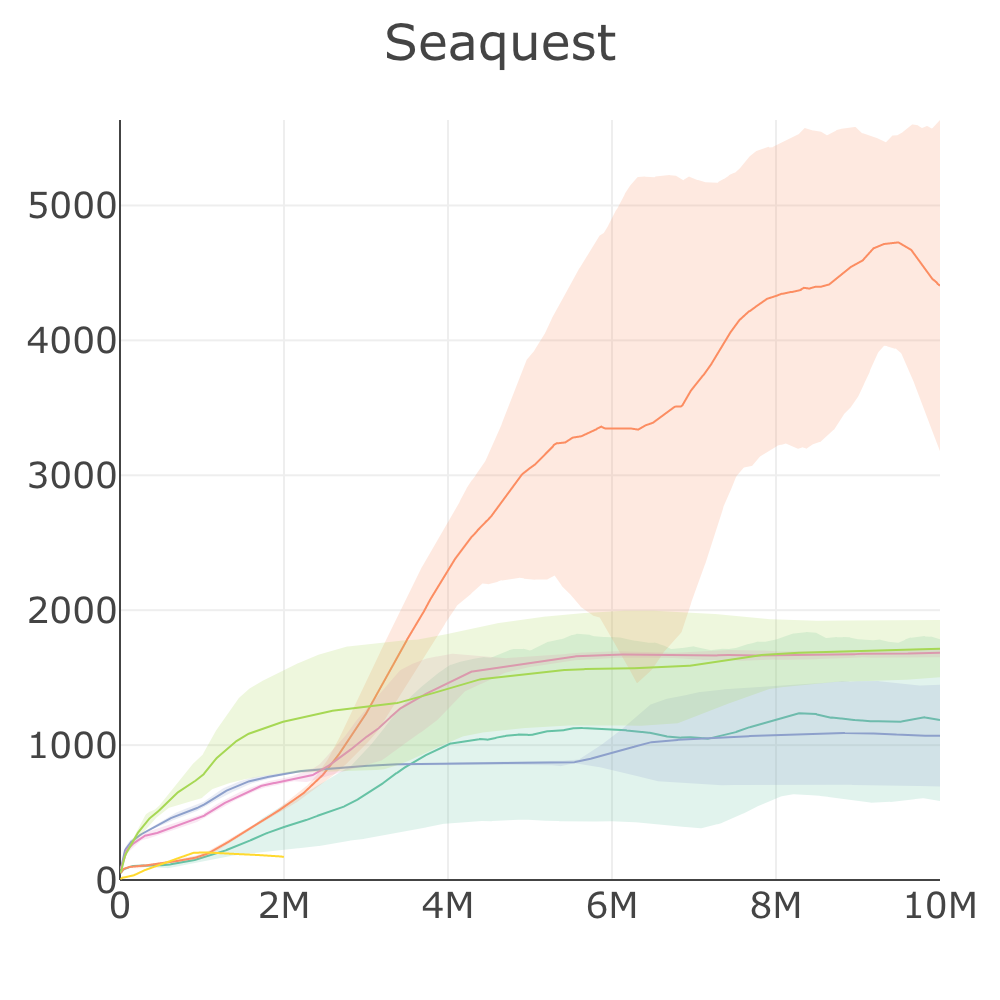} \\
\includegraphics[width=1.22in]{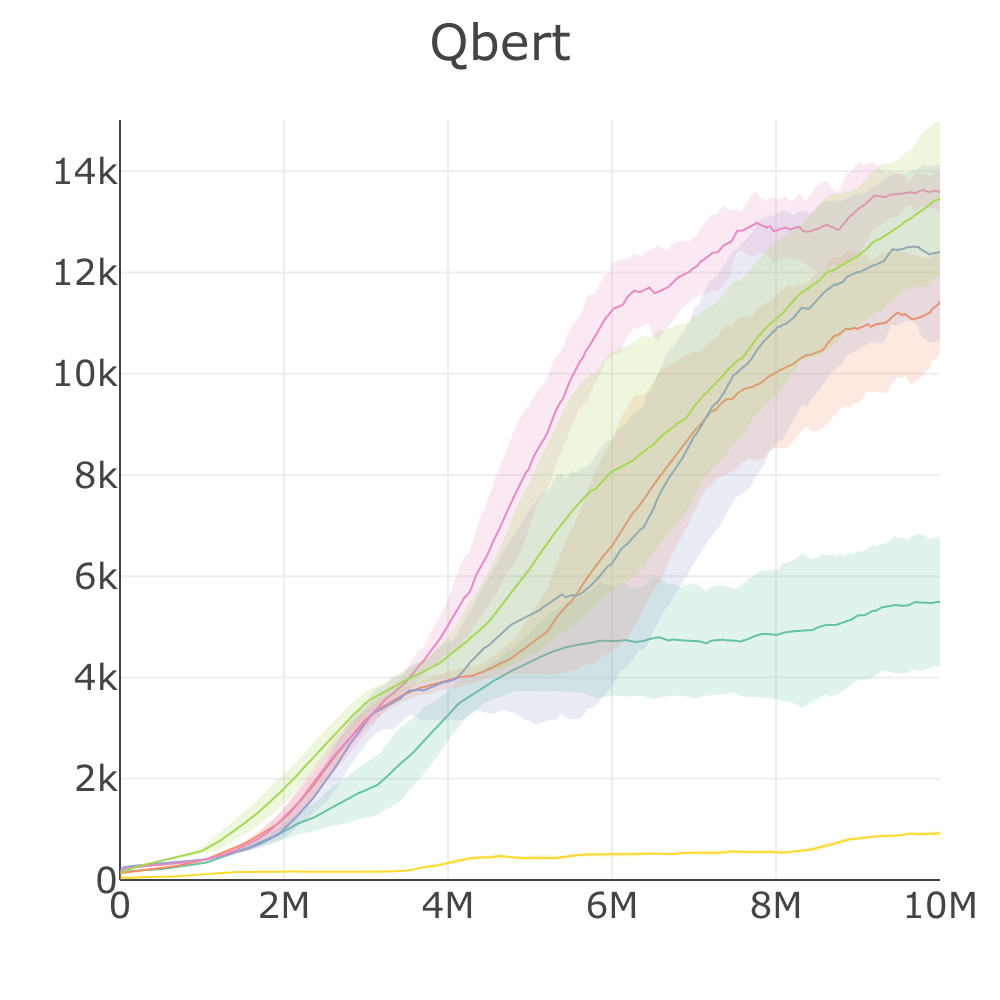} &
\includegraphics[width=1.22in]{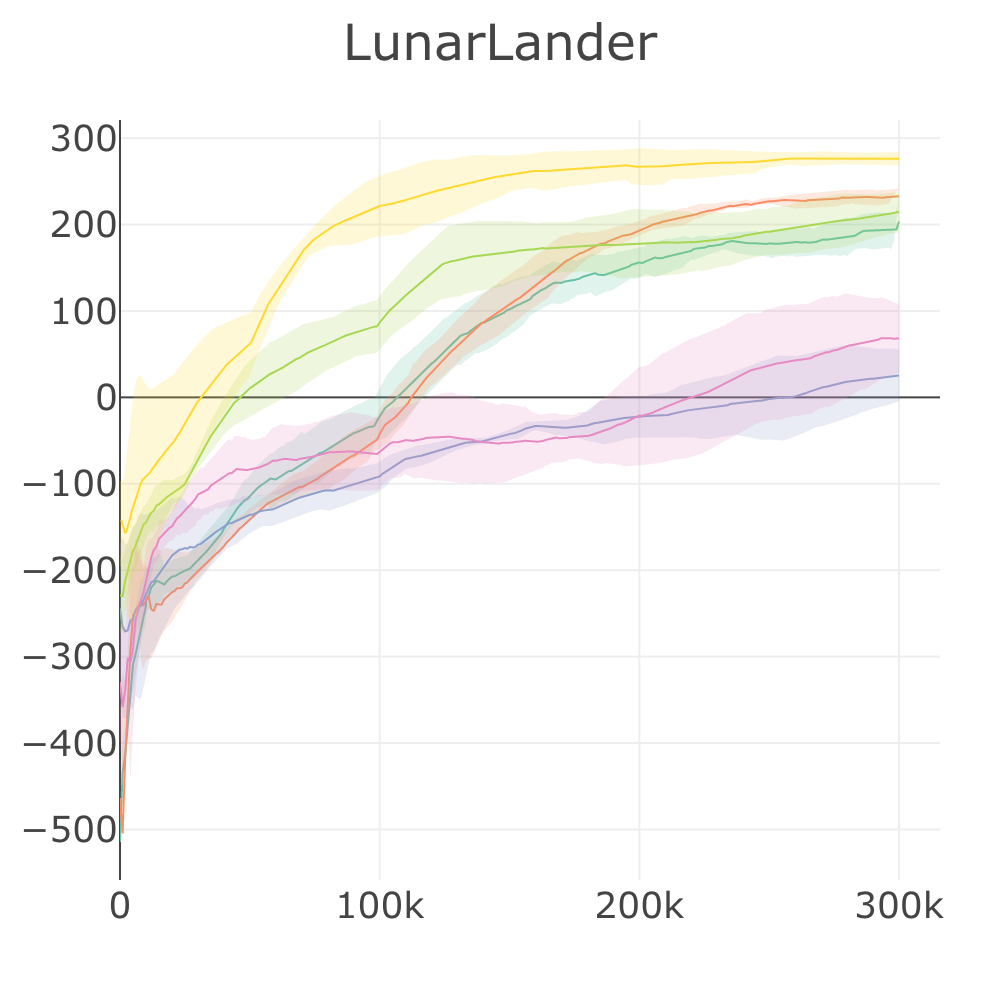} & 
\includegraphics[width=1.22in]{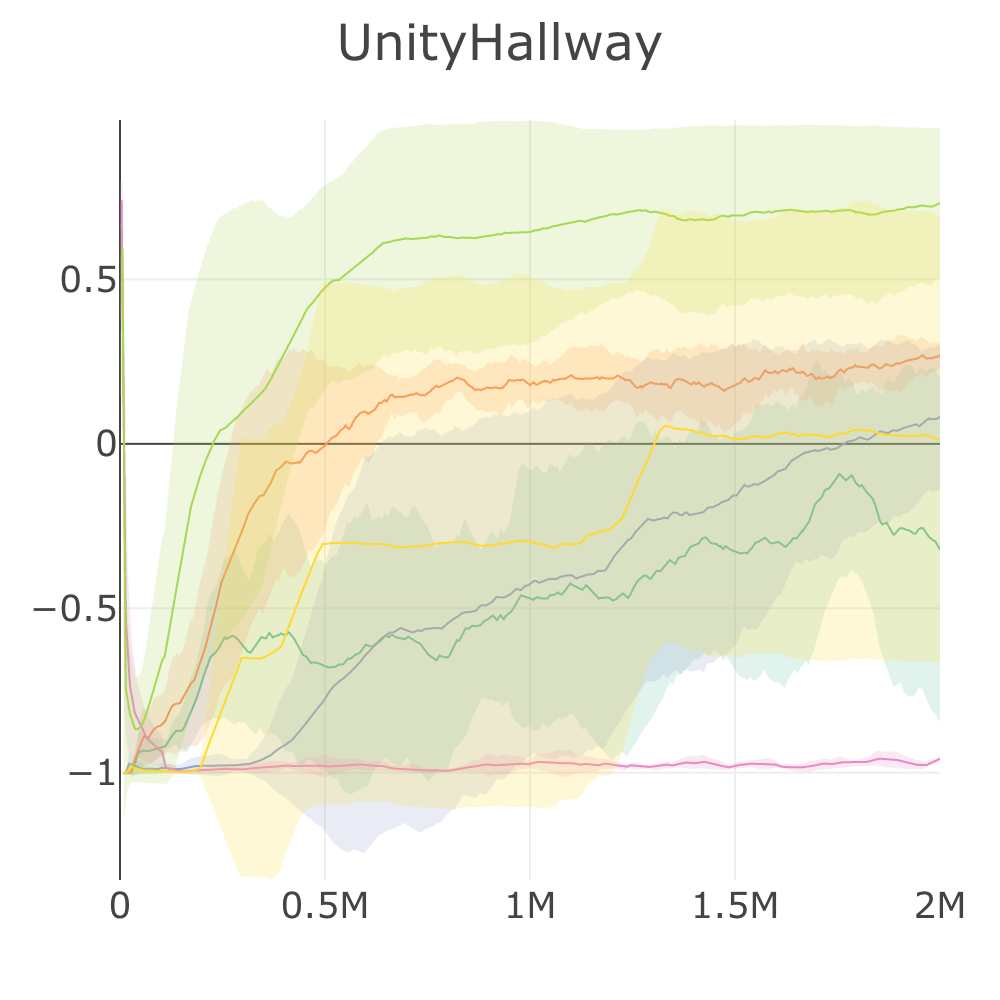} &
\includegraphics[width=1.22in]{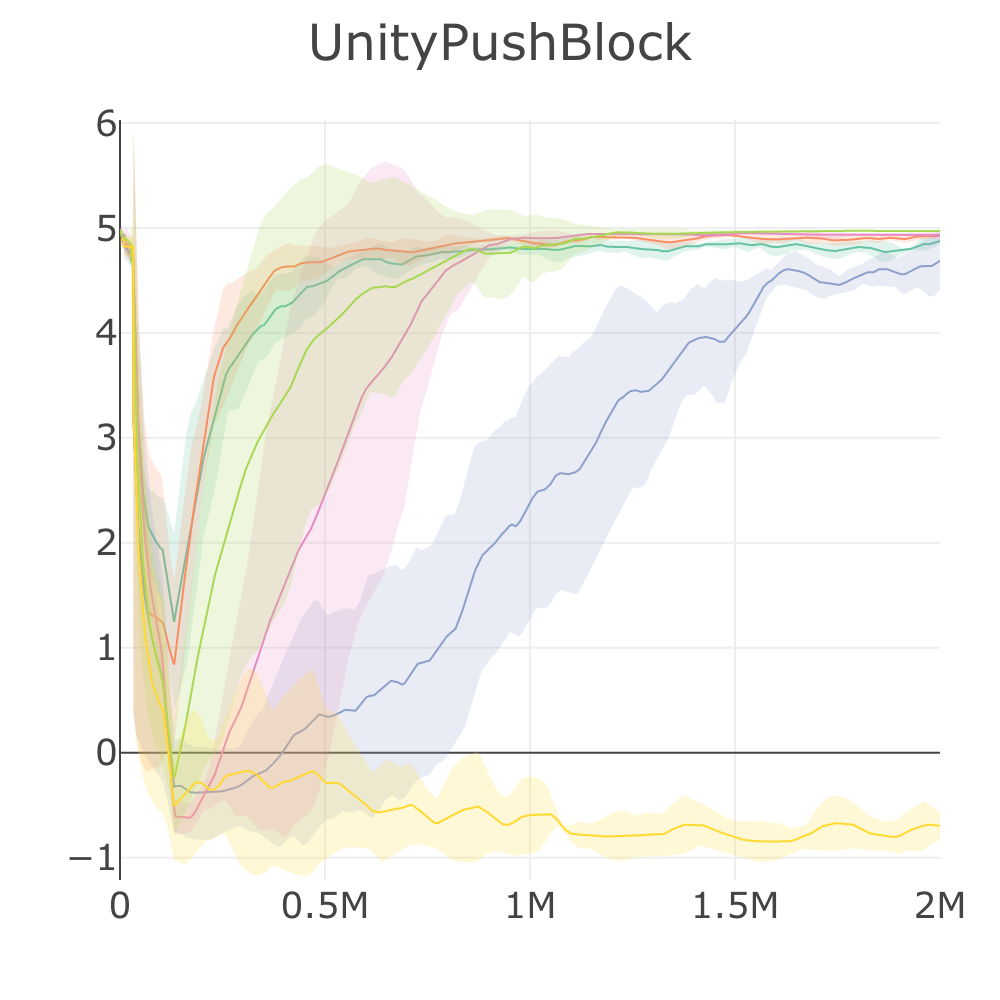}
\end{longtable}

\begin{longtable}{cccc}
\includegraphics[width=1.22in]{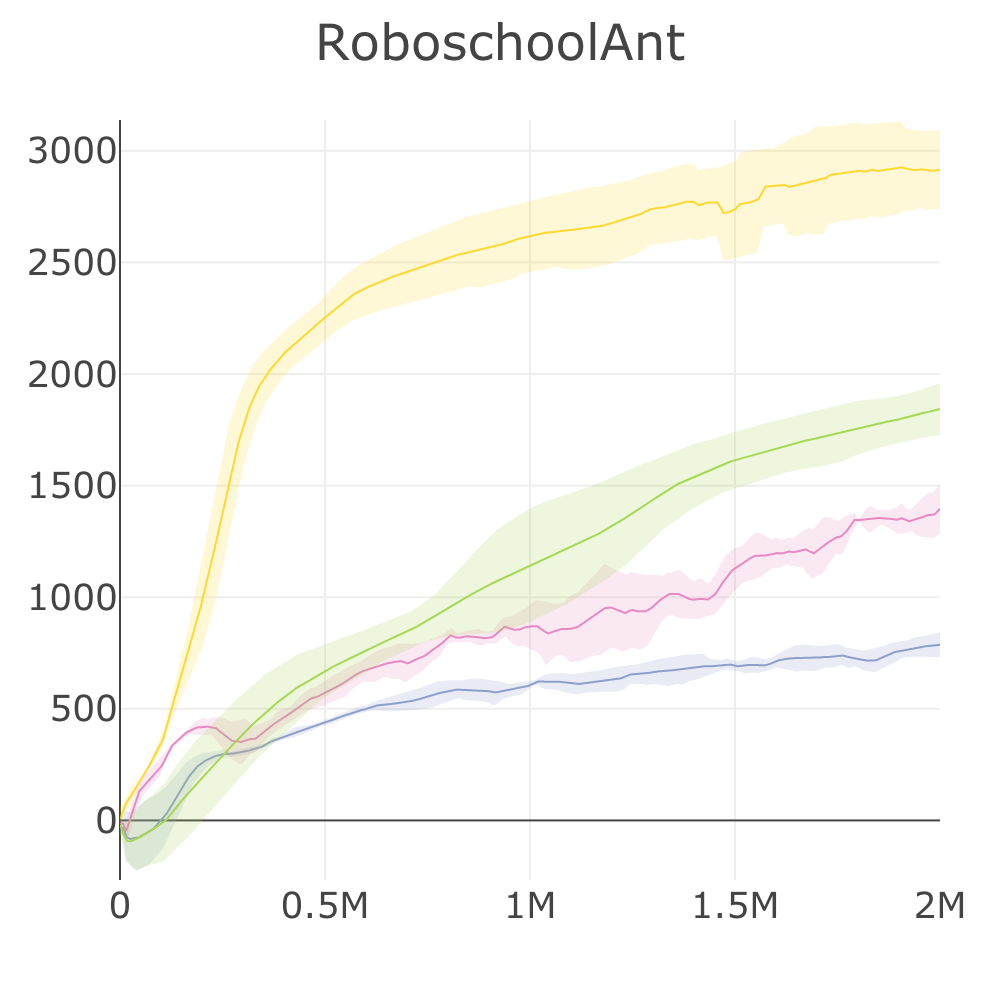} & 
\includegraphics[width=1.22in]{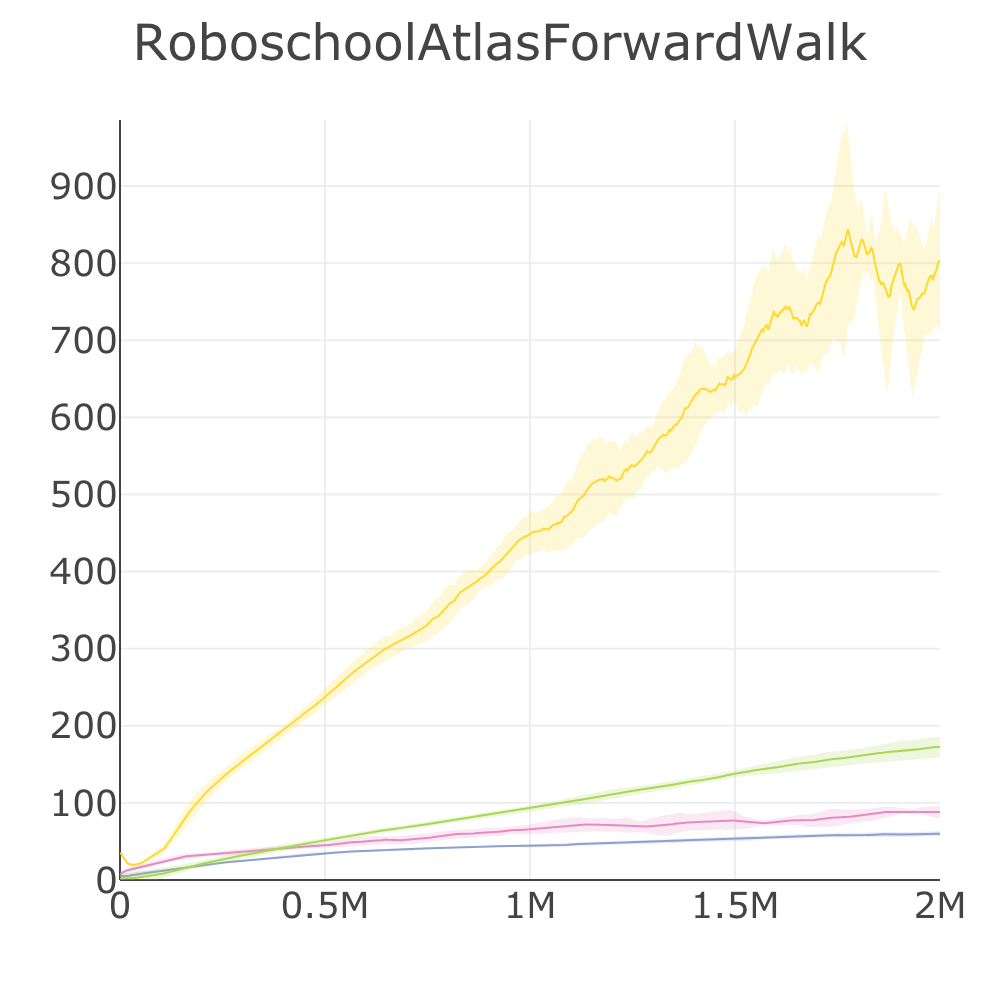} & 
\includegraphics[width=1.22in]{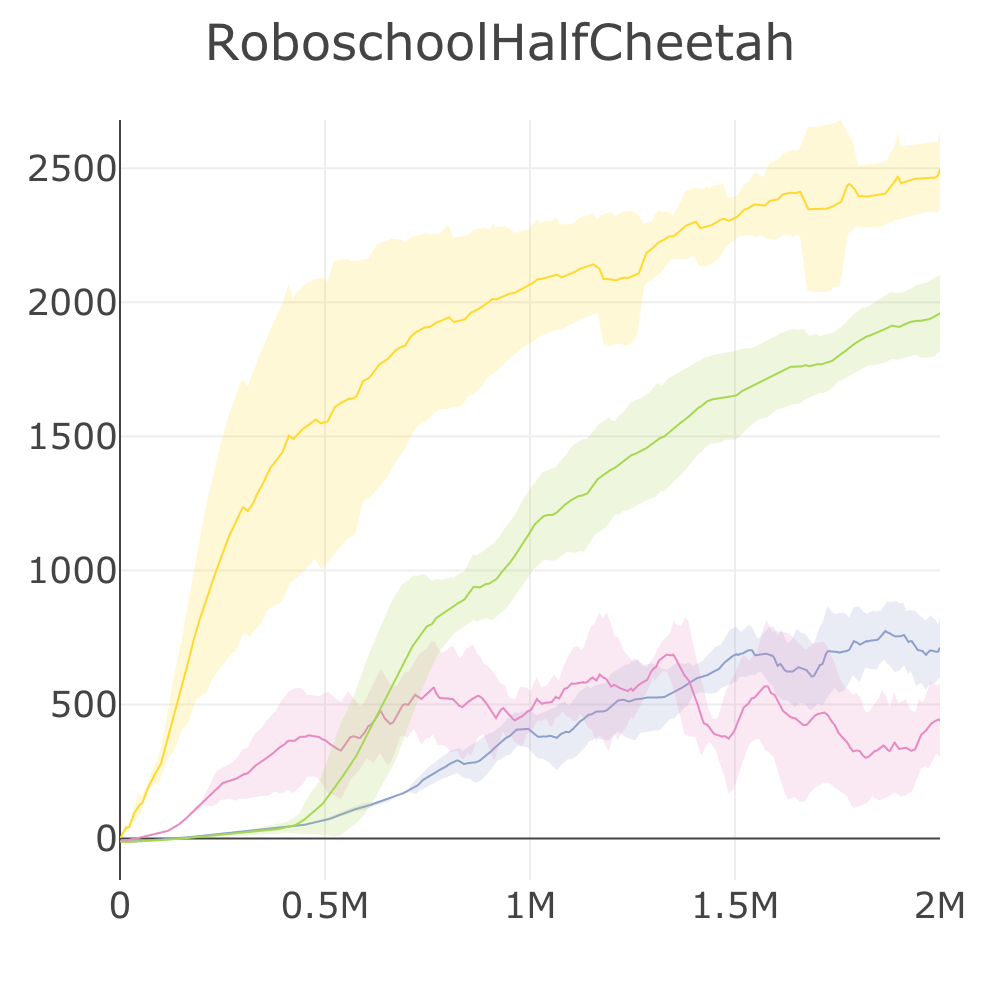} & 
\includegraphics[width=1.22in]{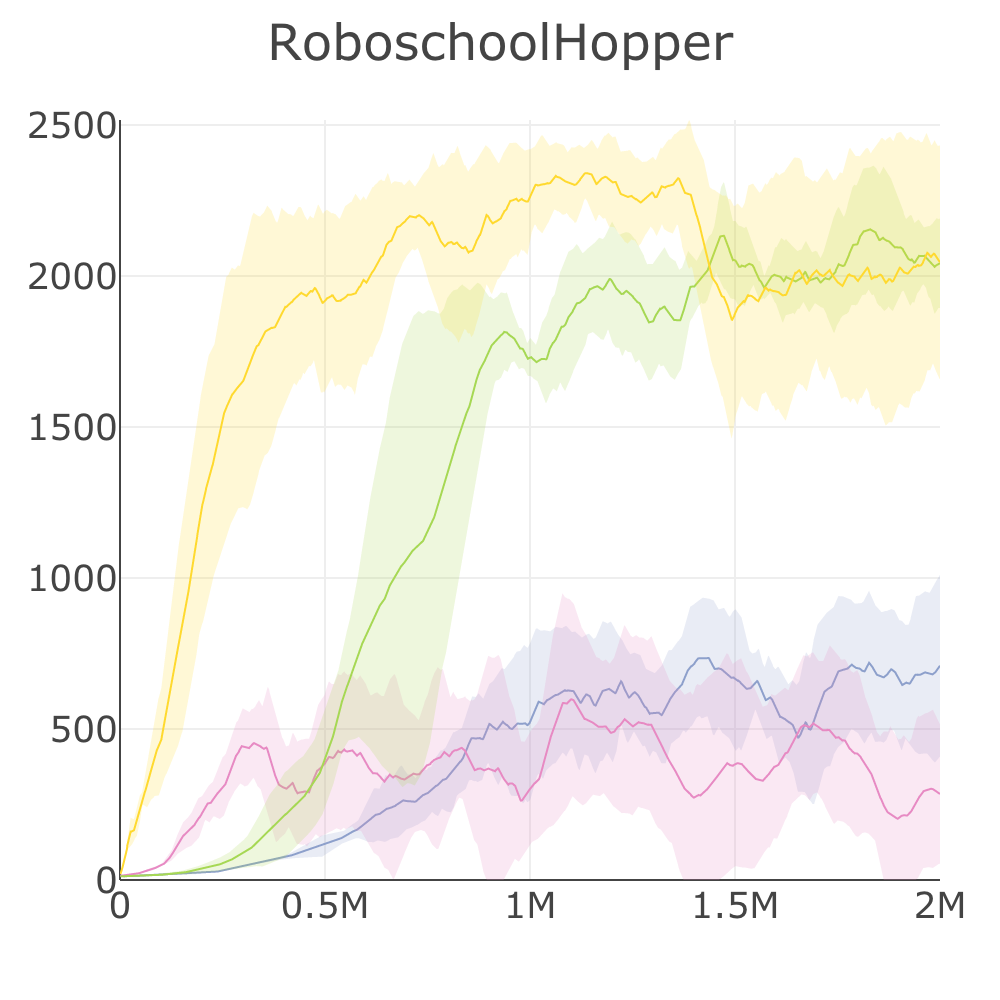}\\
\includegraphics[width=1.22in]{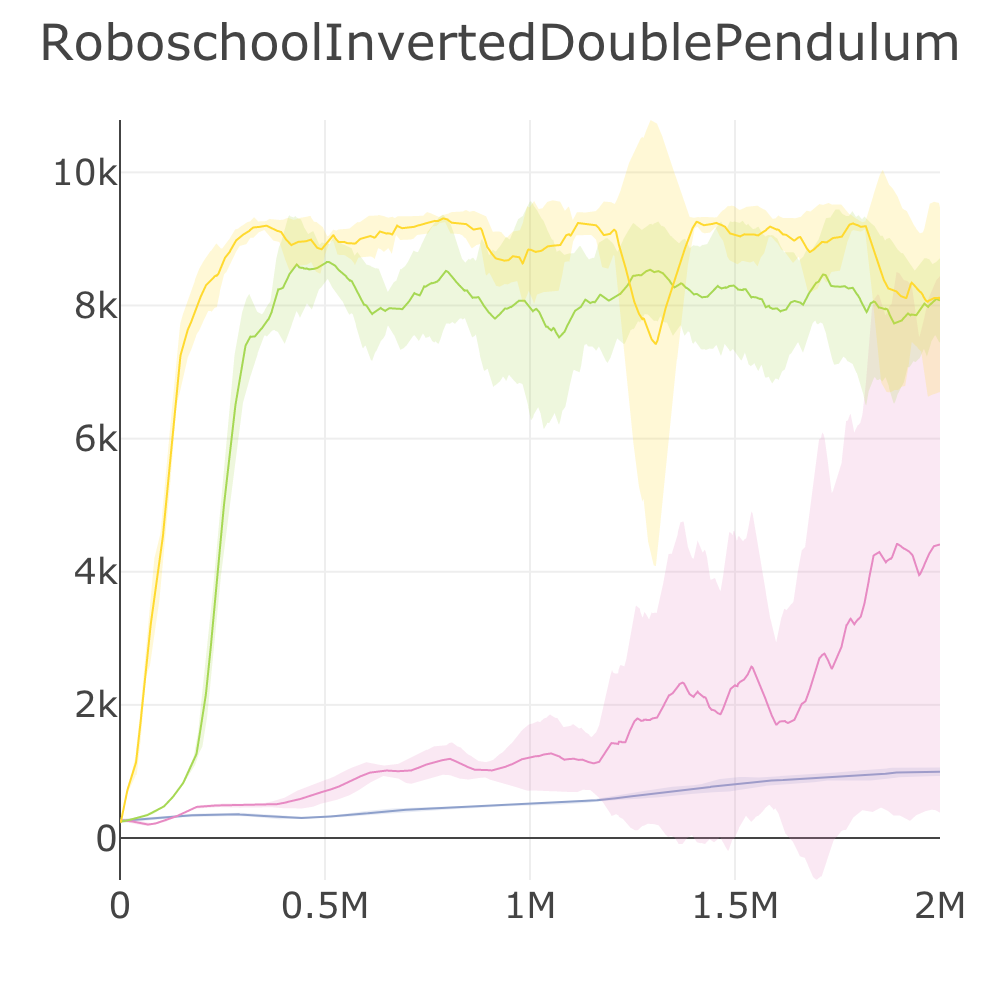} & 
\includegraphics[width=1.22in]{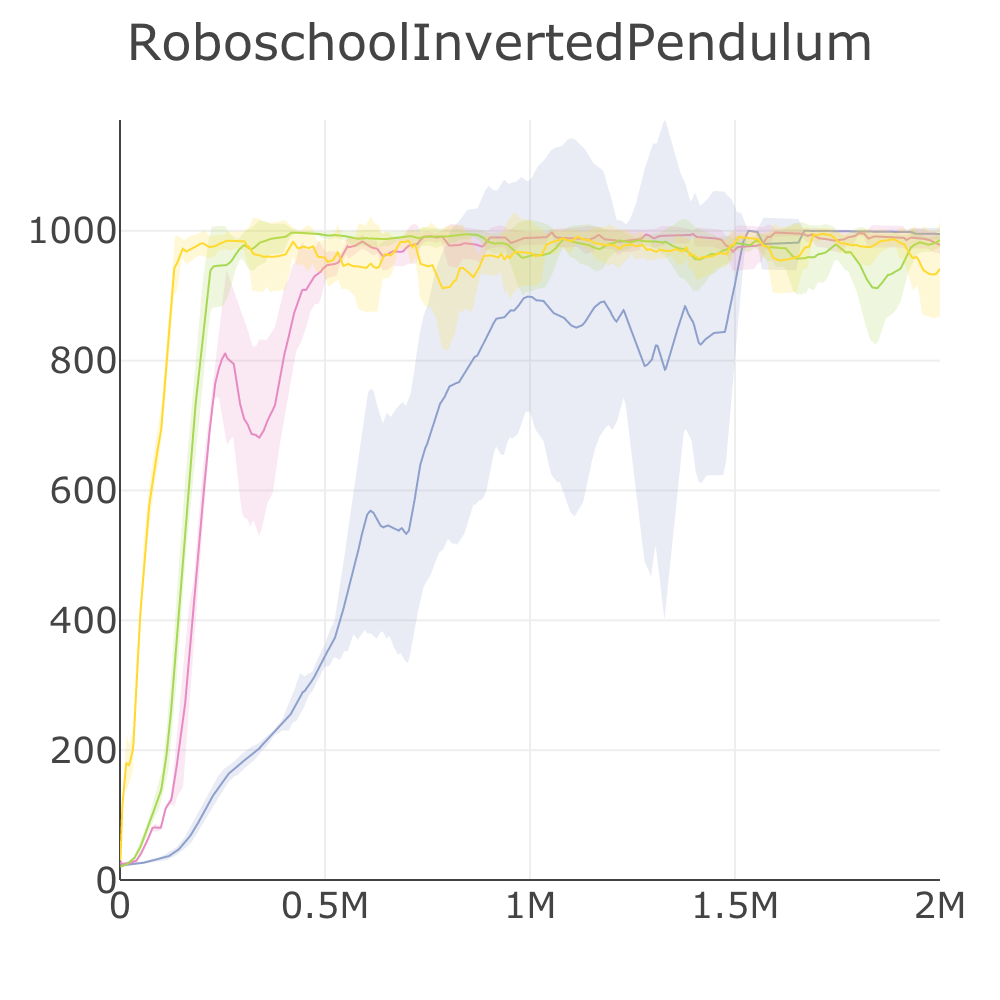} & 
\includegraphics[width=1.22in]{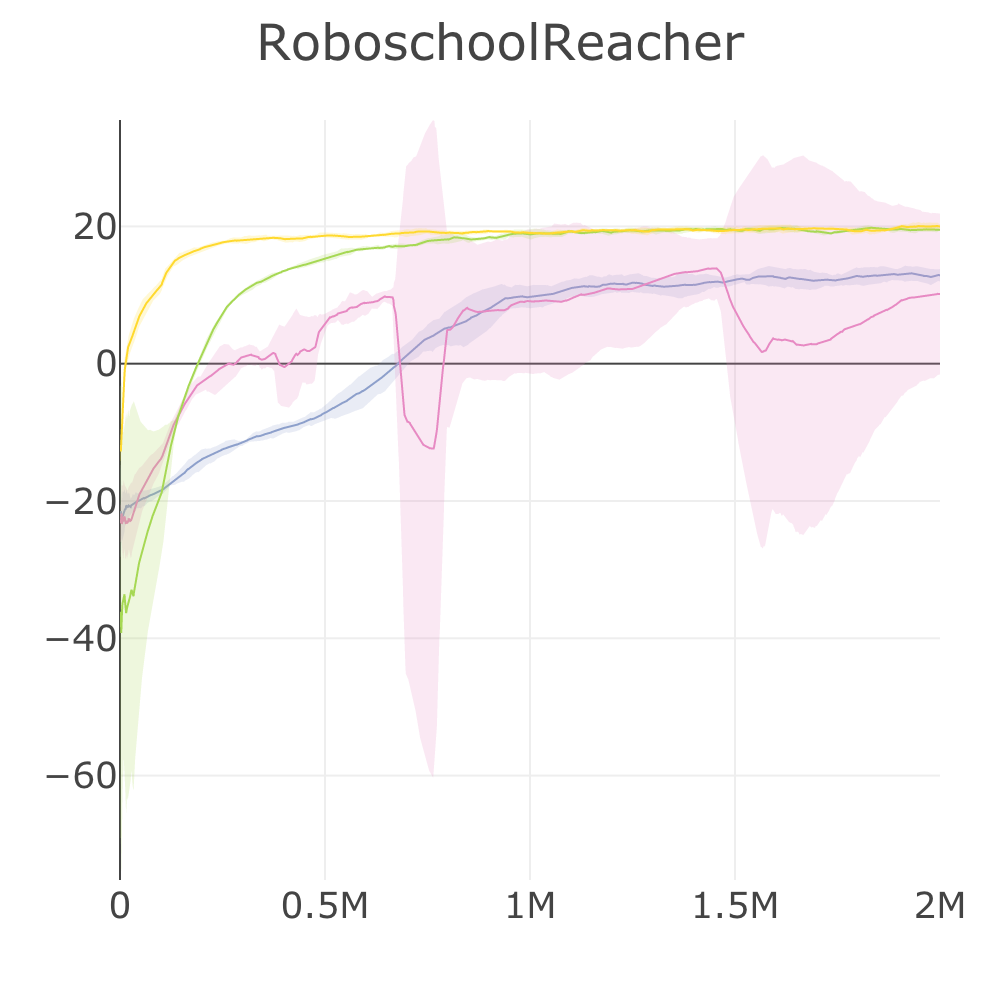} & 
\includegraphics[width=1.22in]{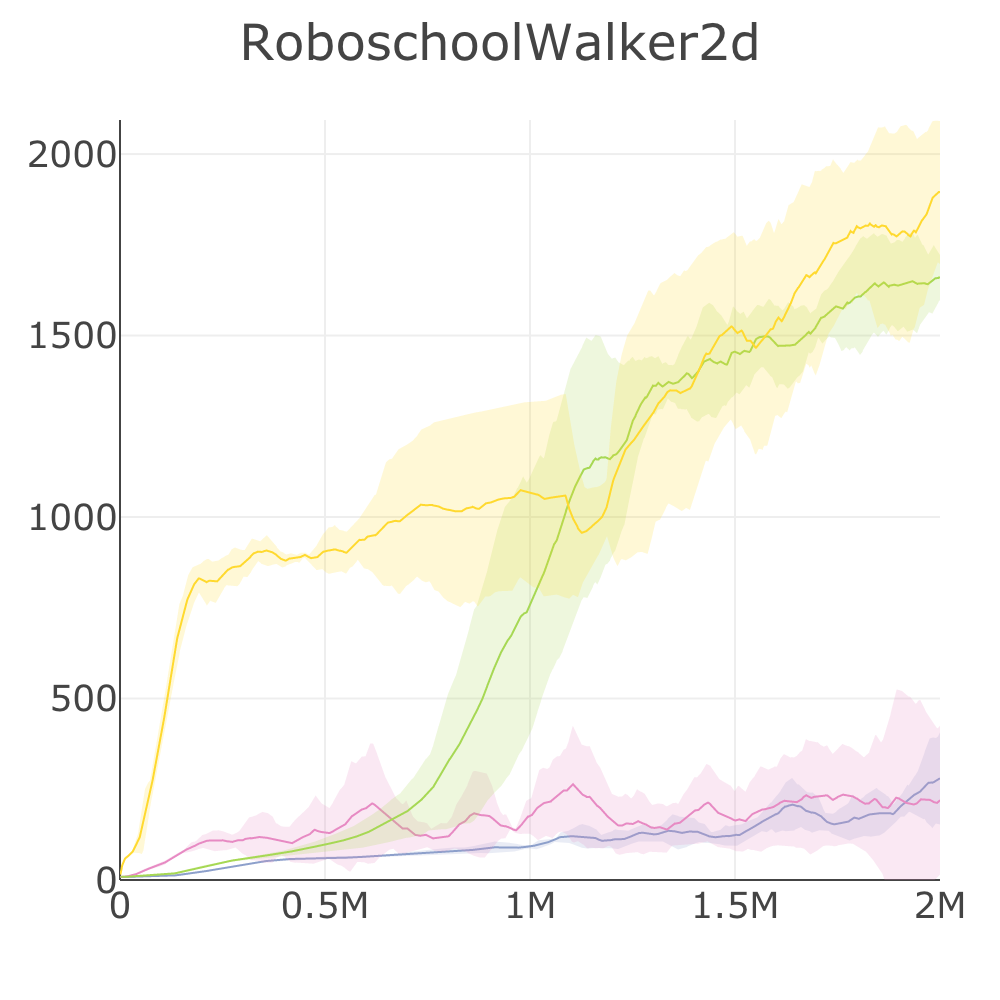}\\
\includegraphics[width=1.22in]{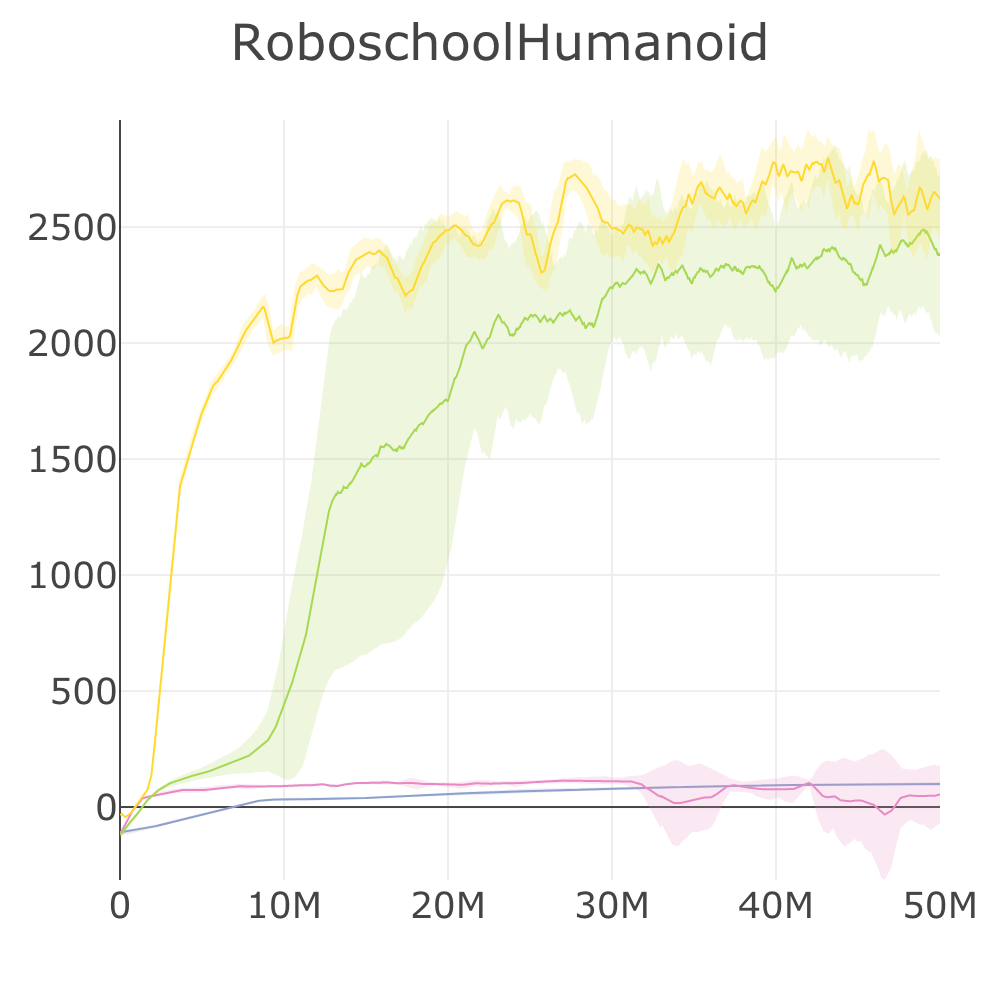} & 
\includegraphics[width=1.22in]{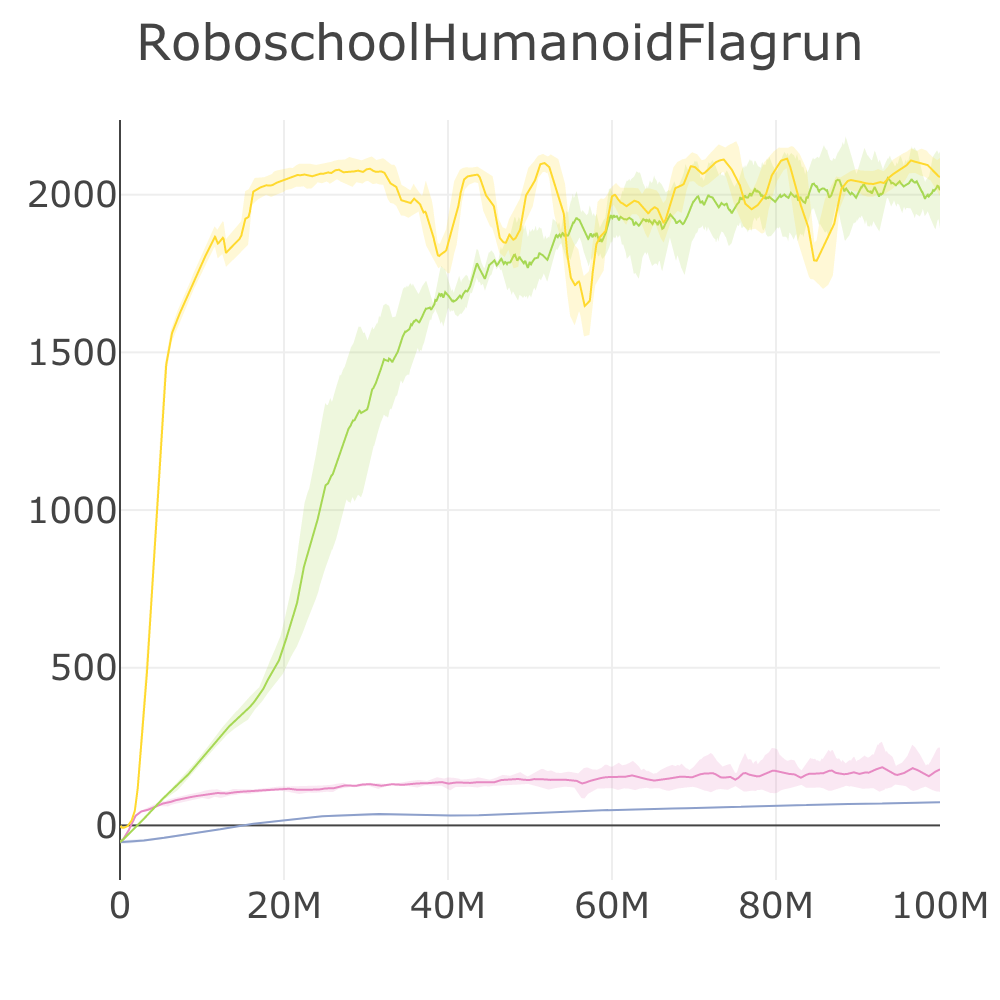} & 
\includegraphics[width=1.22in]{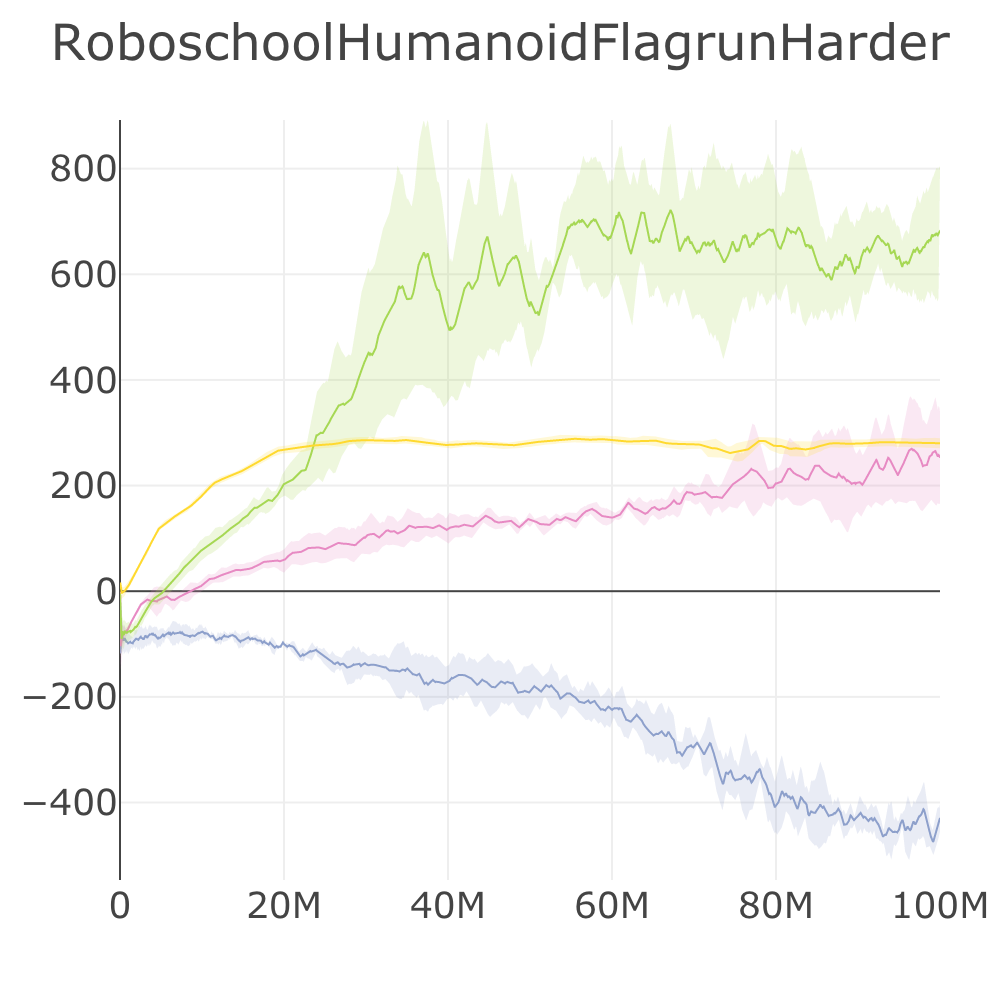} & 
\includegraphics[width=1.22in]{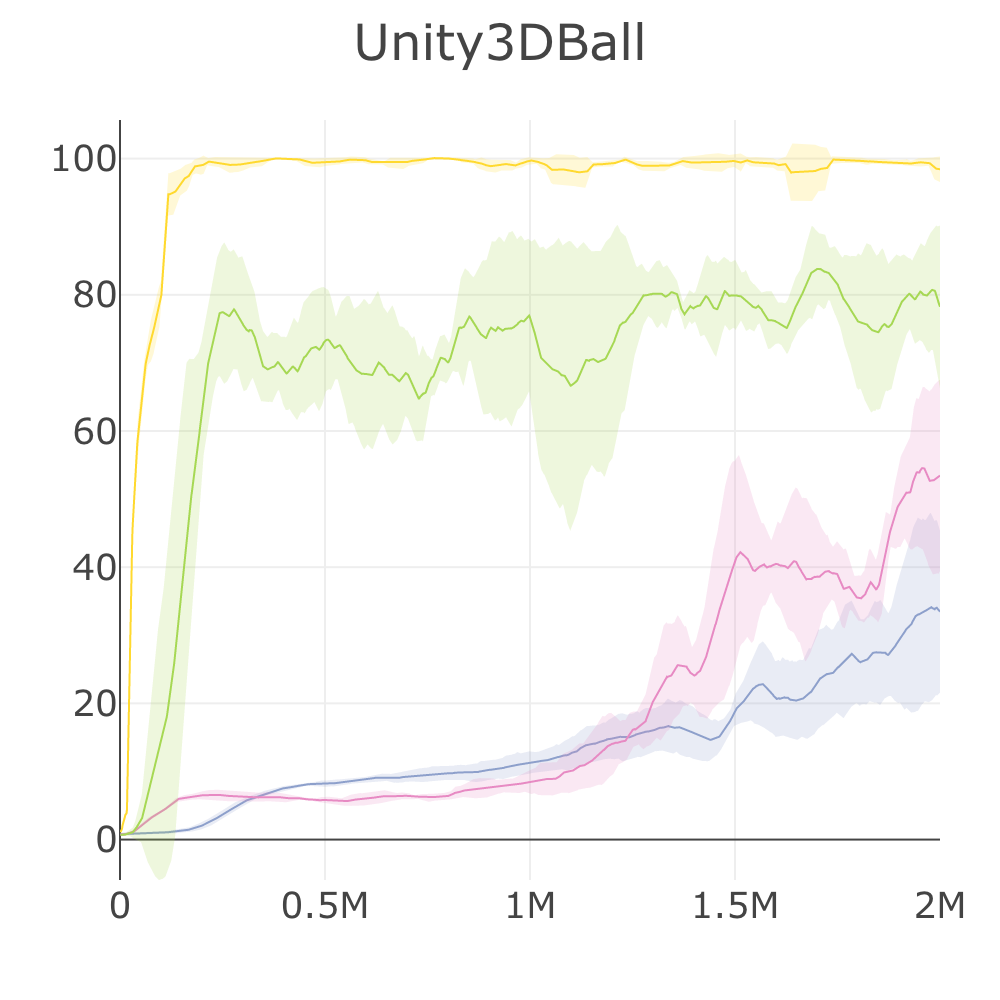}\\
\includegraphics[width=1.22in]{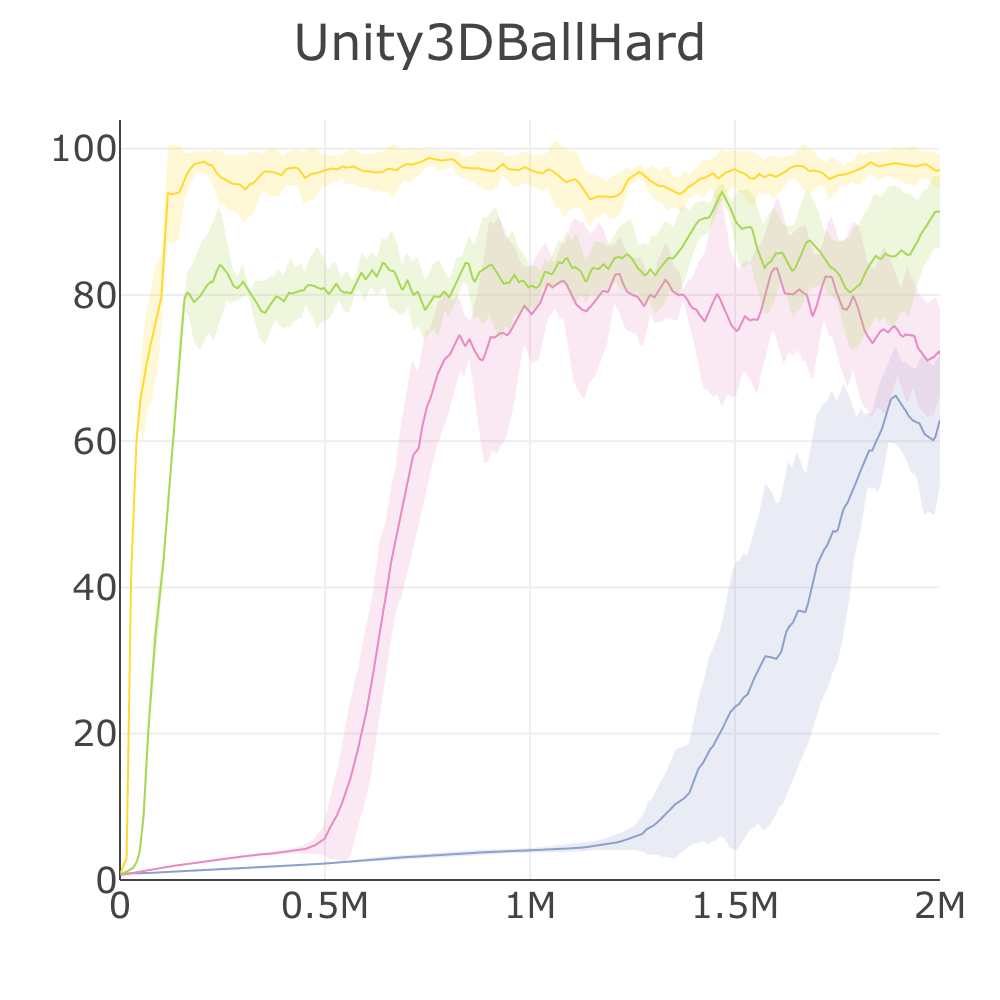} & 
\end{longtable}

\begin{longtable}{cccc}
\includegraphics[width=1.22in]{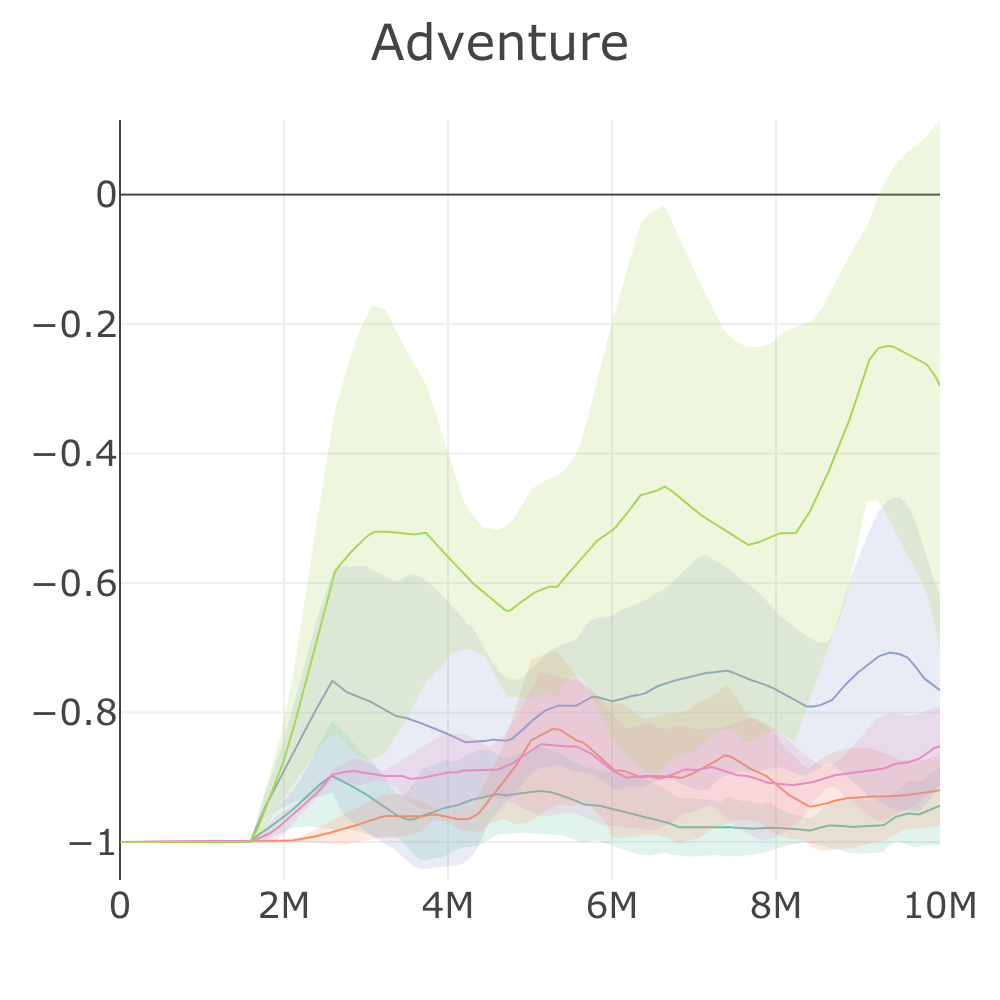} & 
\includegraphics[width=1.22in]{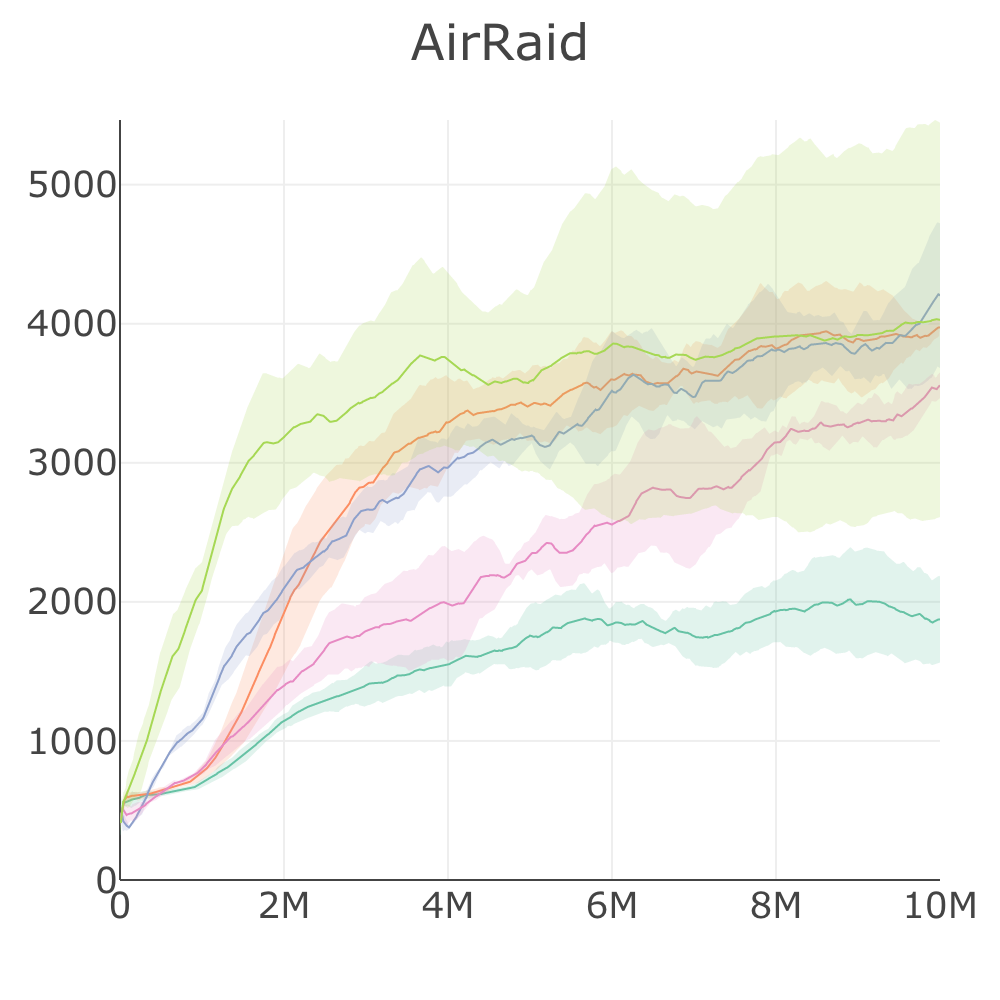} & 
\includegraphics[width=1.22in]{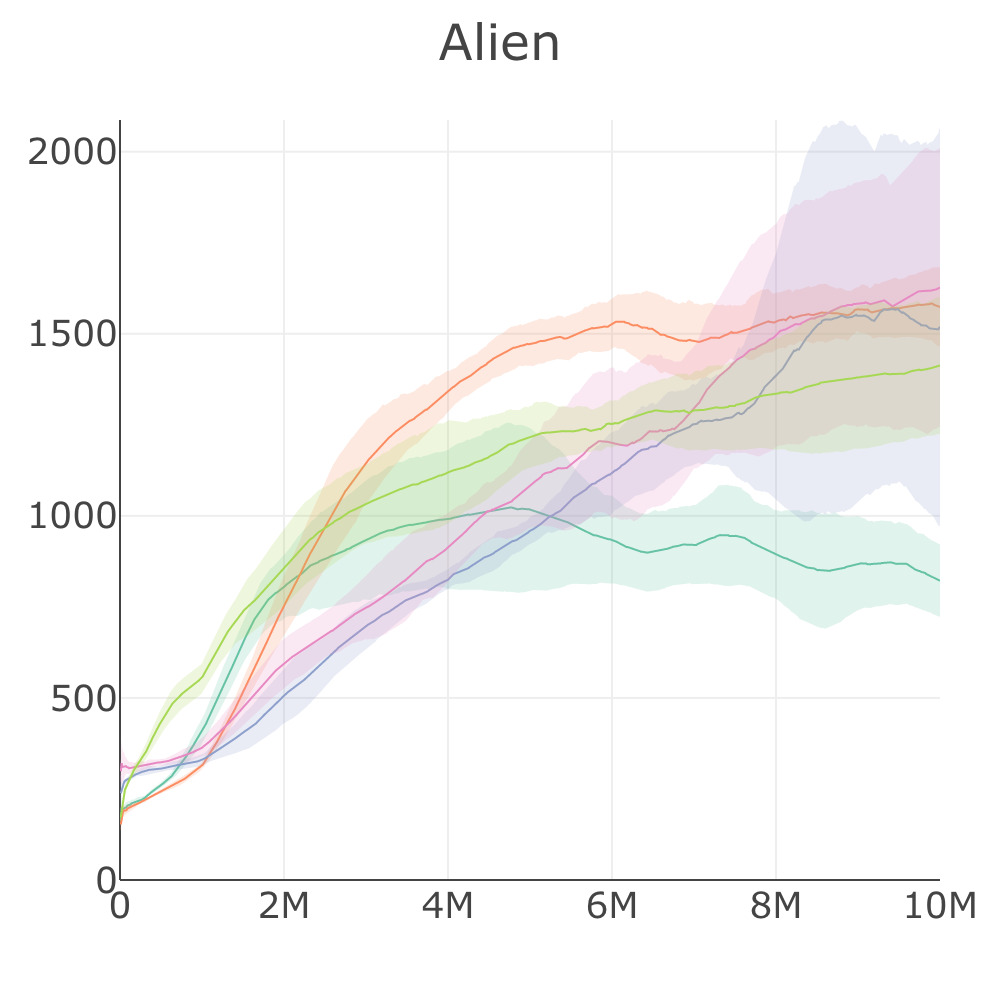} & 
\includegraphics[width=1.22in]{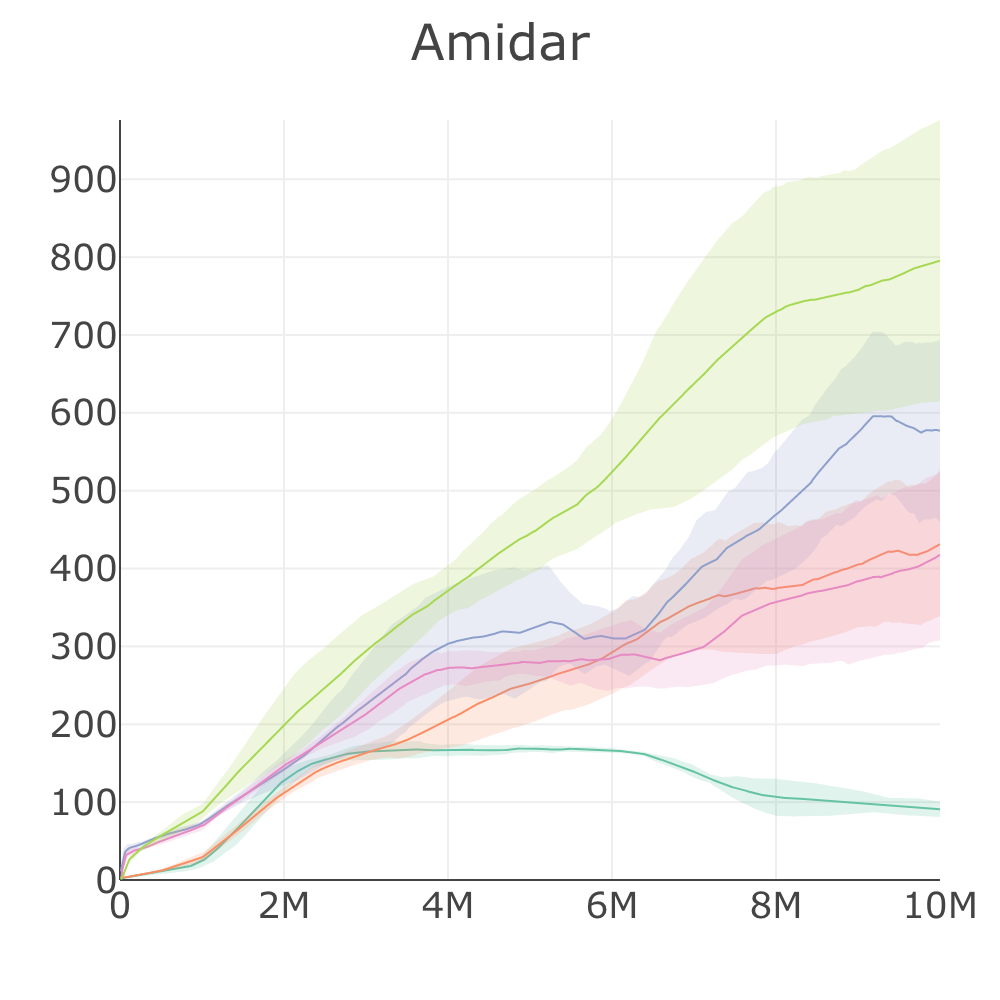}\\
\includegraphics[width=1.22in]{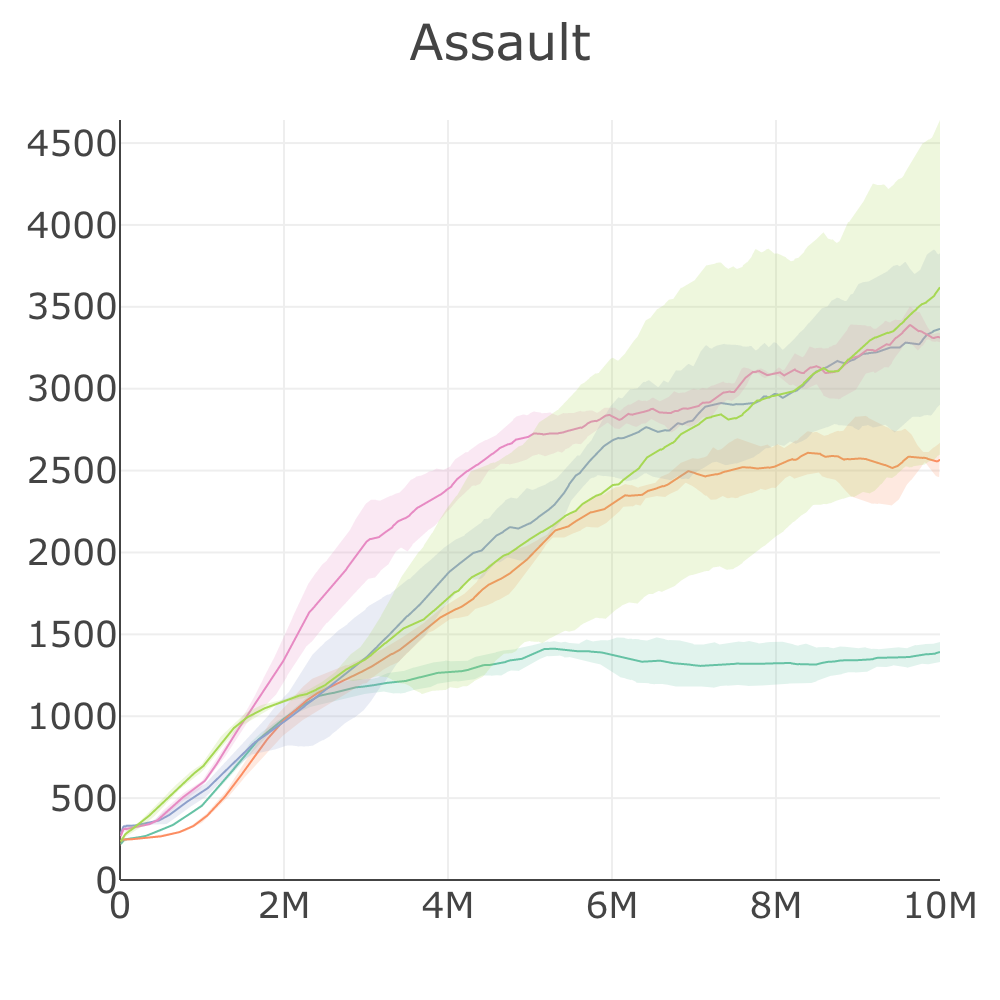} & 
\includegraphics[width=1.22in]{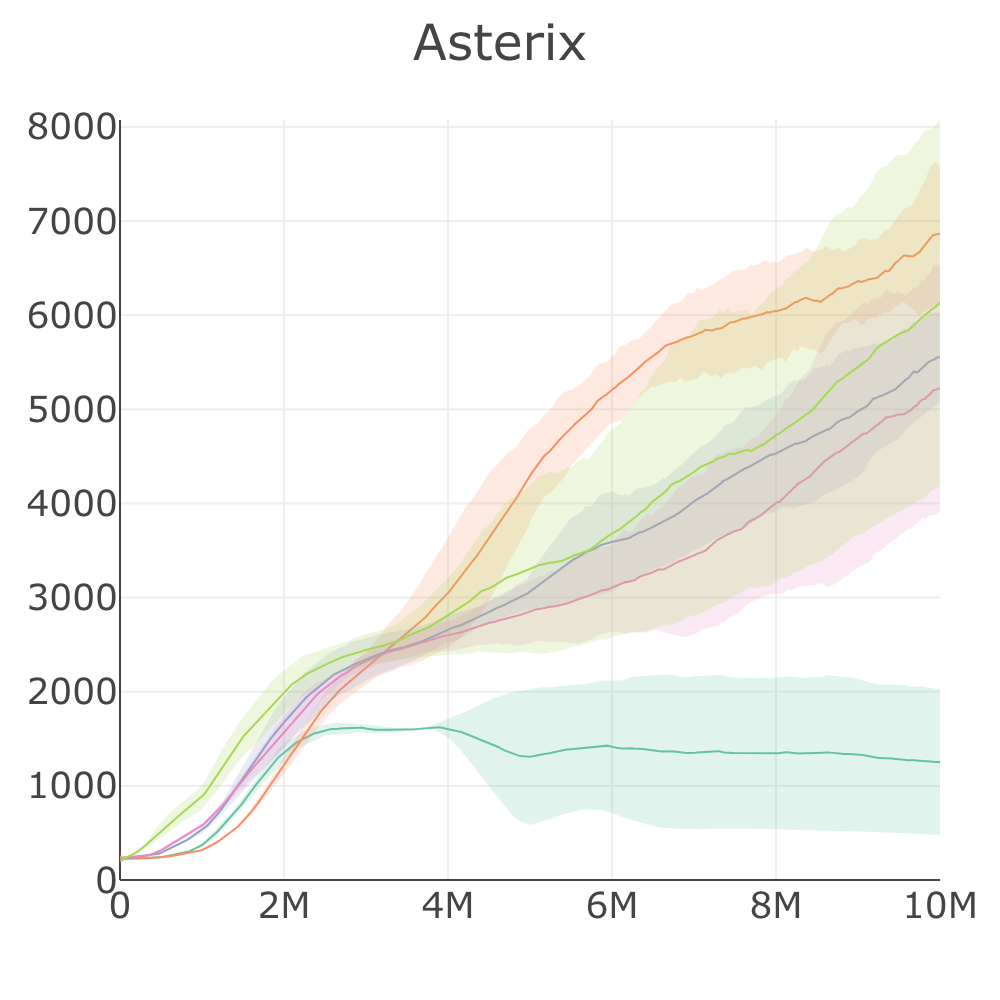} & 
\includegraphics[width=1.22in]{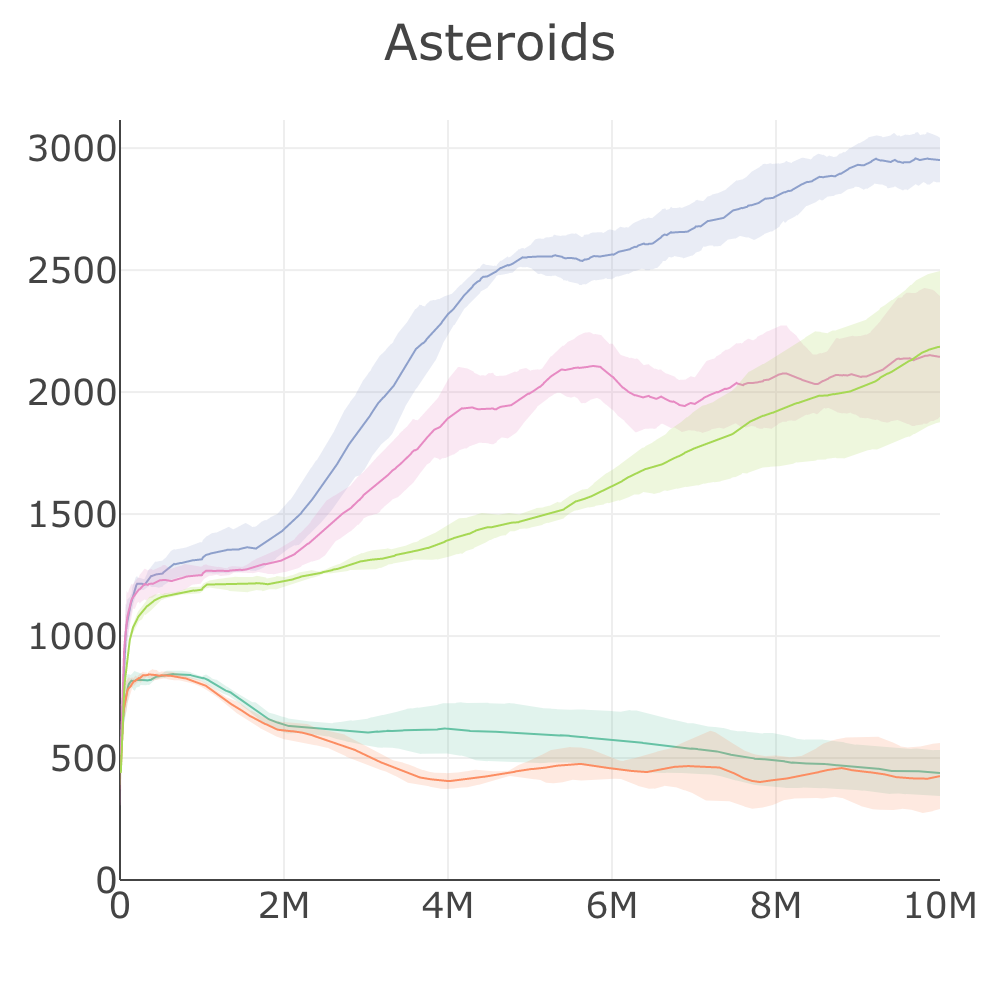} & 
\includegraphics[width=1.22in]{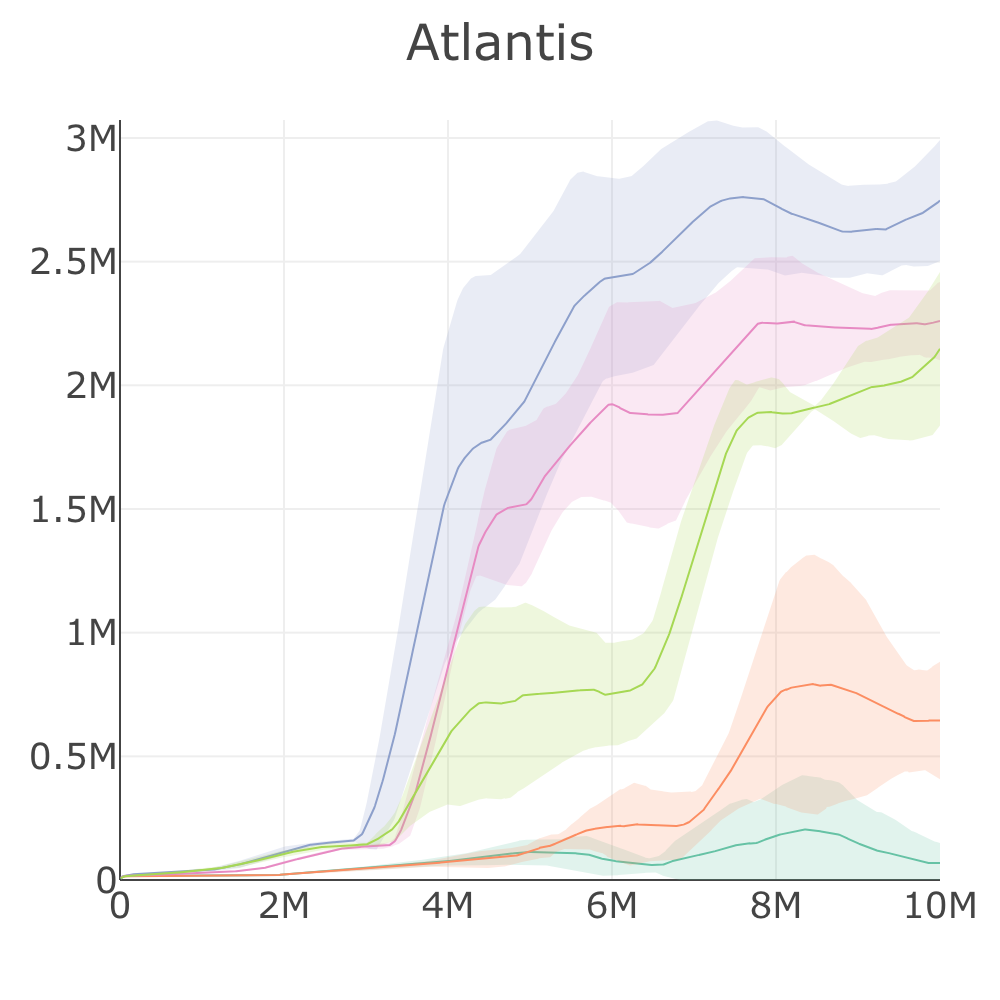}\\
\includegraphics[width=1.22in]{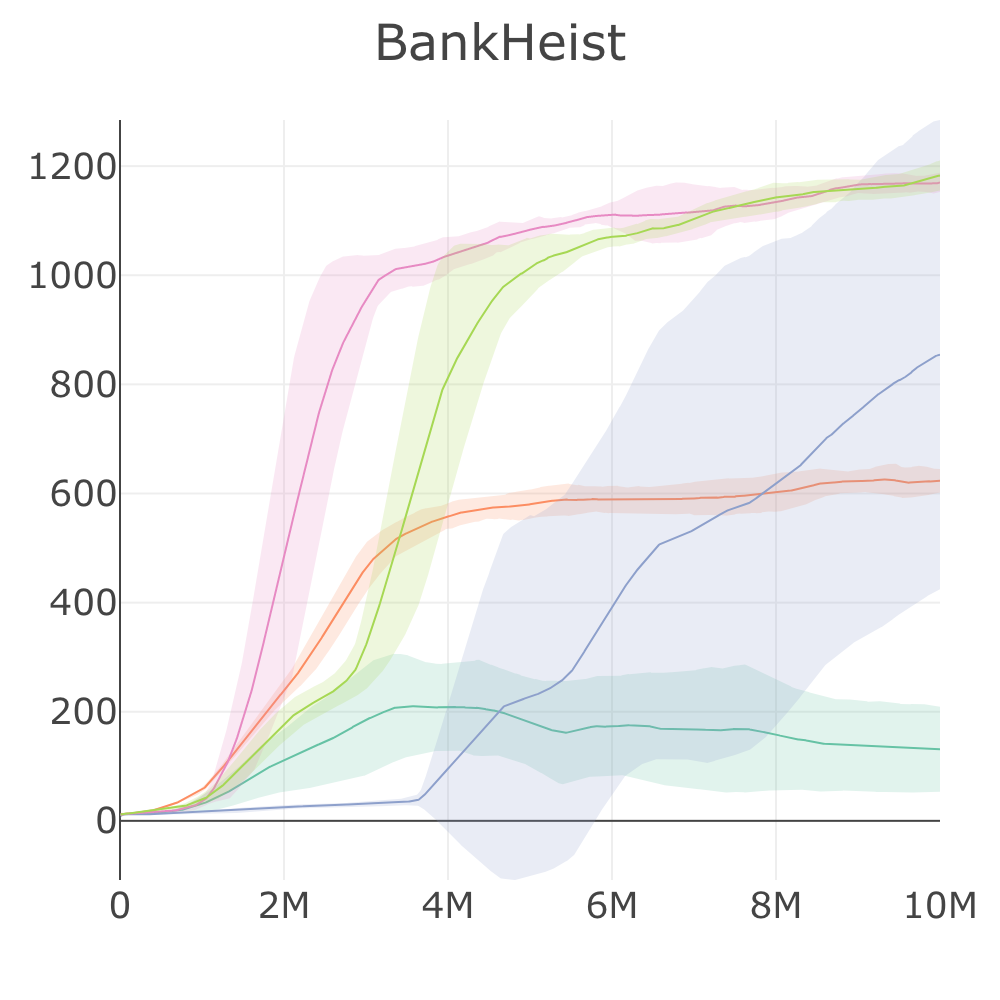} & 
\includegraphics[width=1.22in]{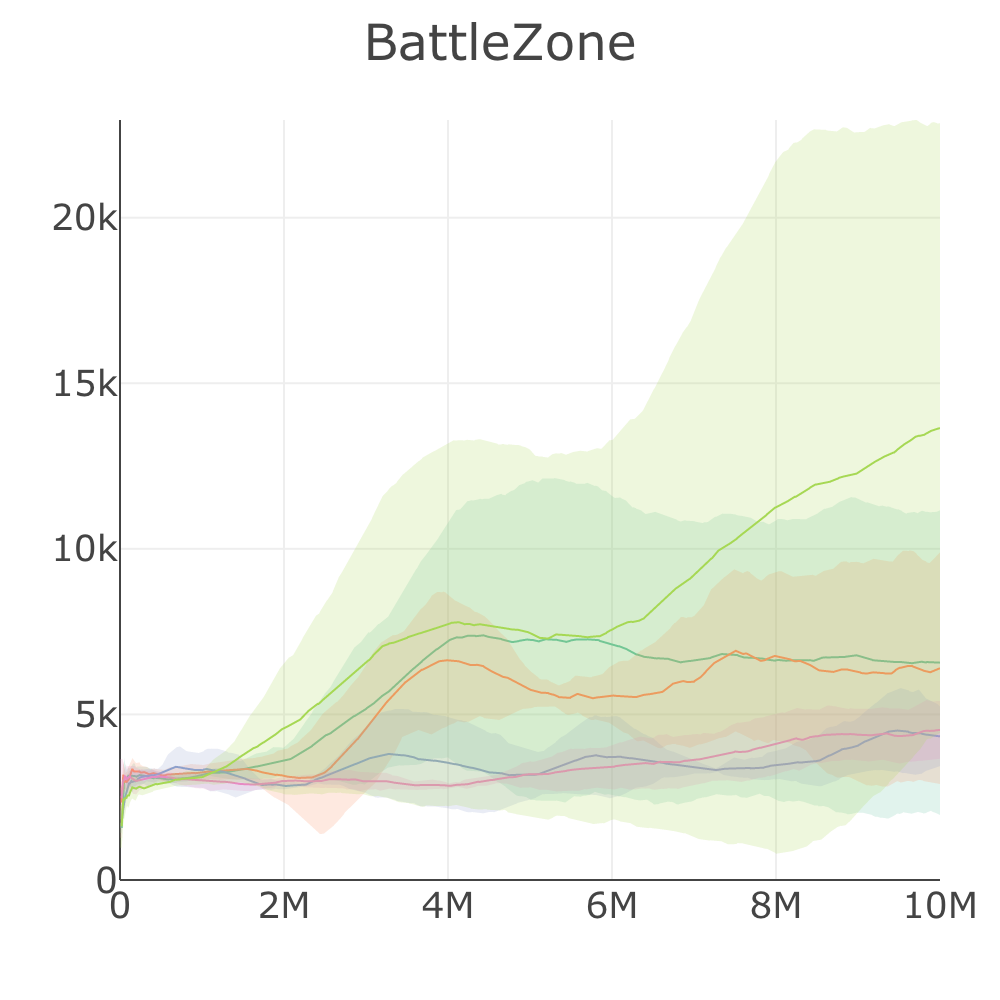} & 
\includegraphics[width=1.22in]{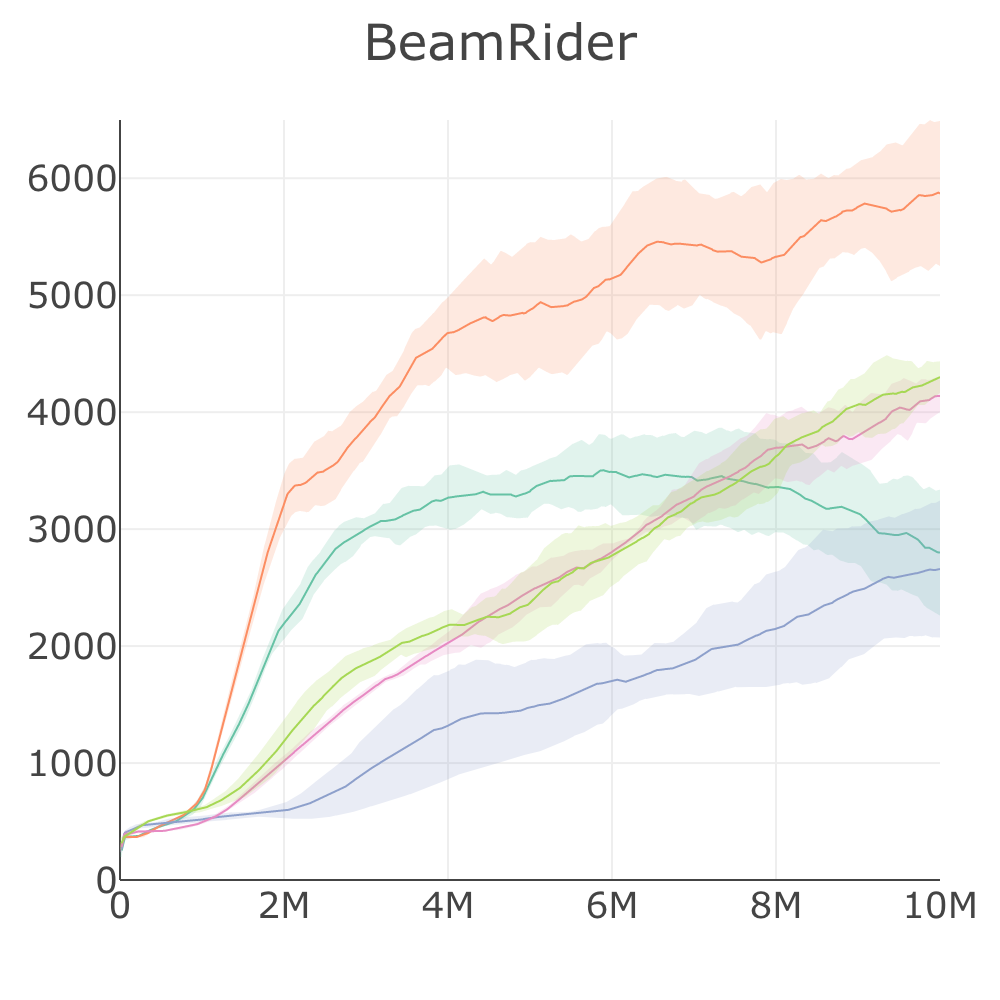} & 
\includegraphics[width=1.22in]{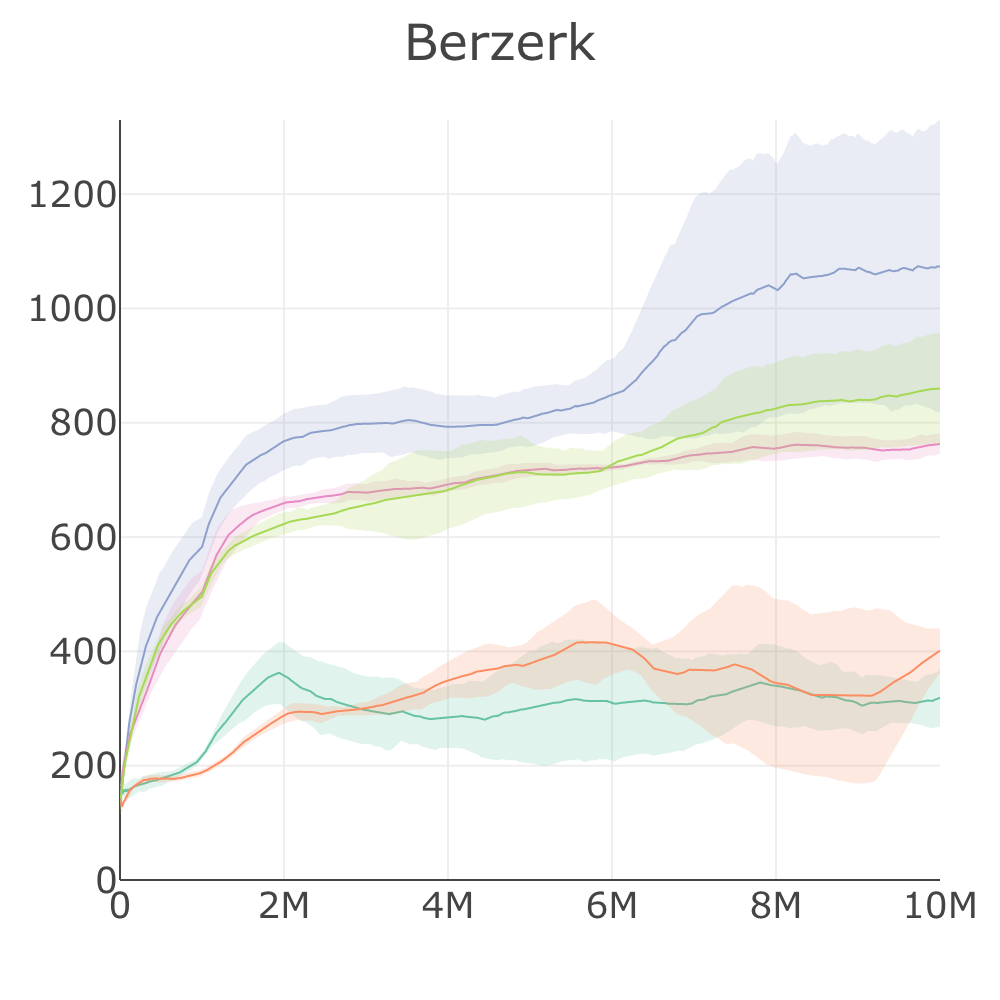}\\
\includegraphics[width=1.22in]{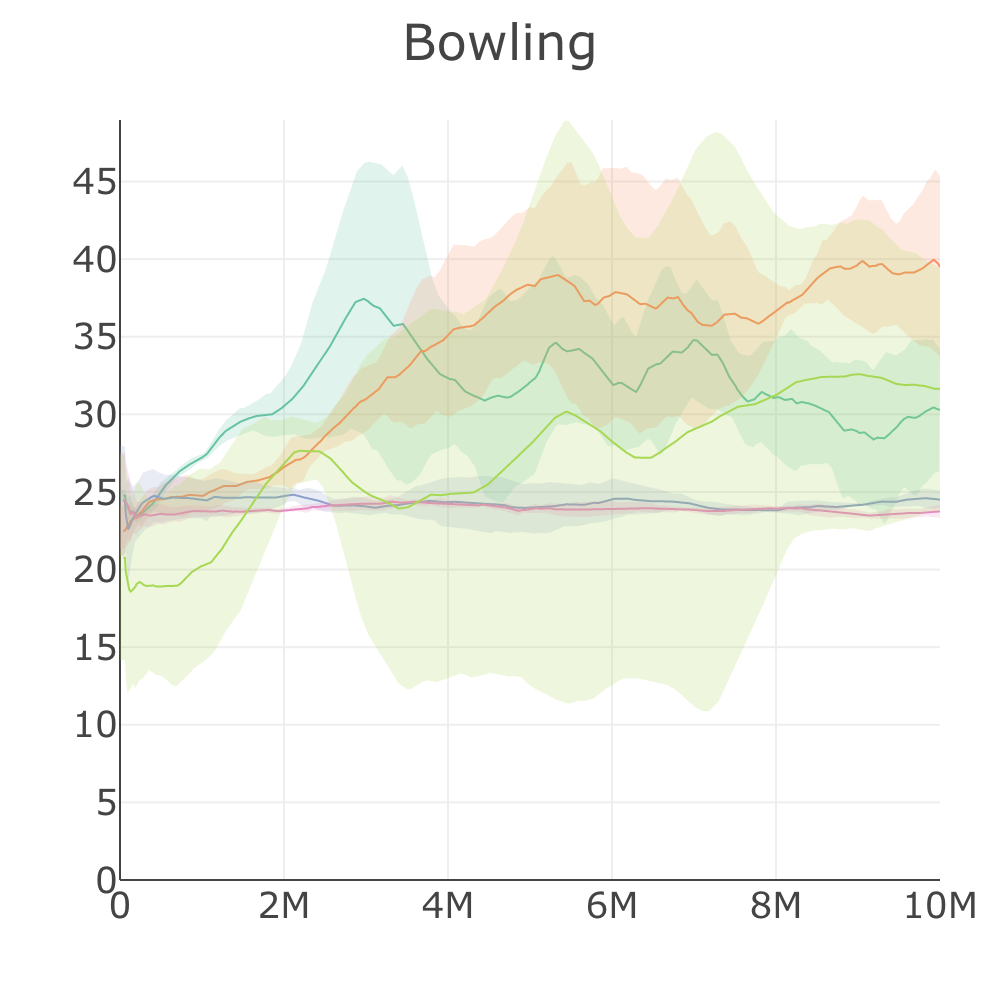} & 
\includegraphics[width=1.22in]{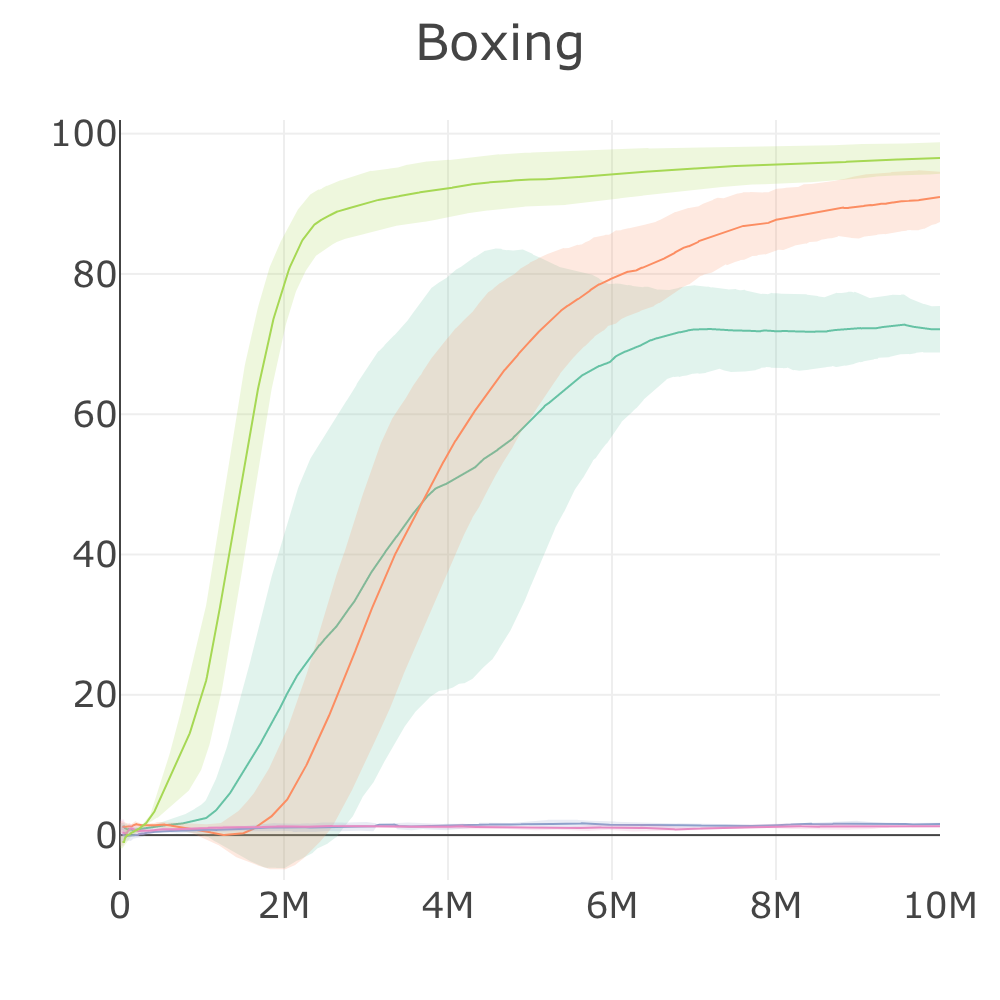} & 
\includegraphics[width=1.22in]{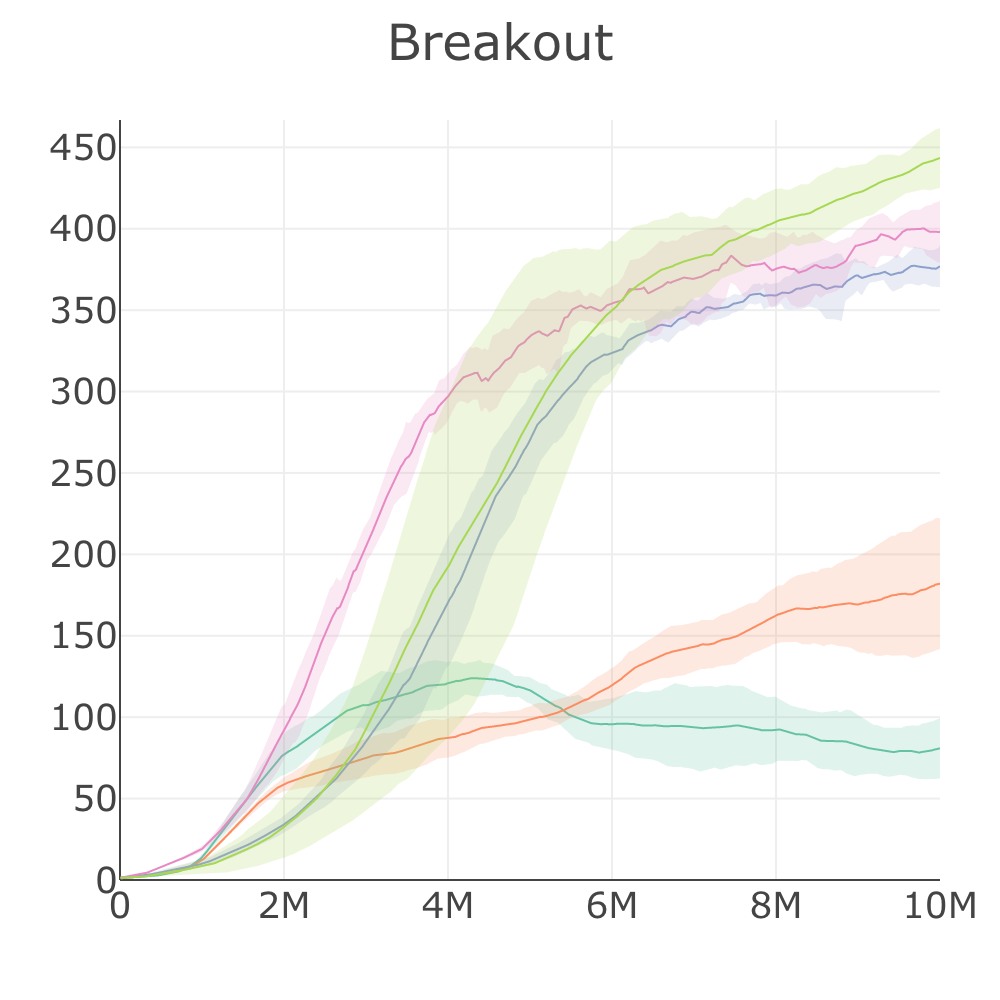} & 
\includegraphics[width=1.22in]{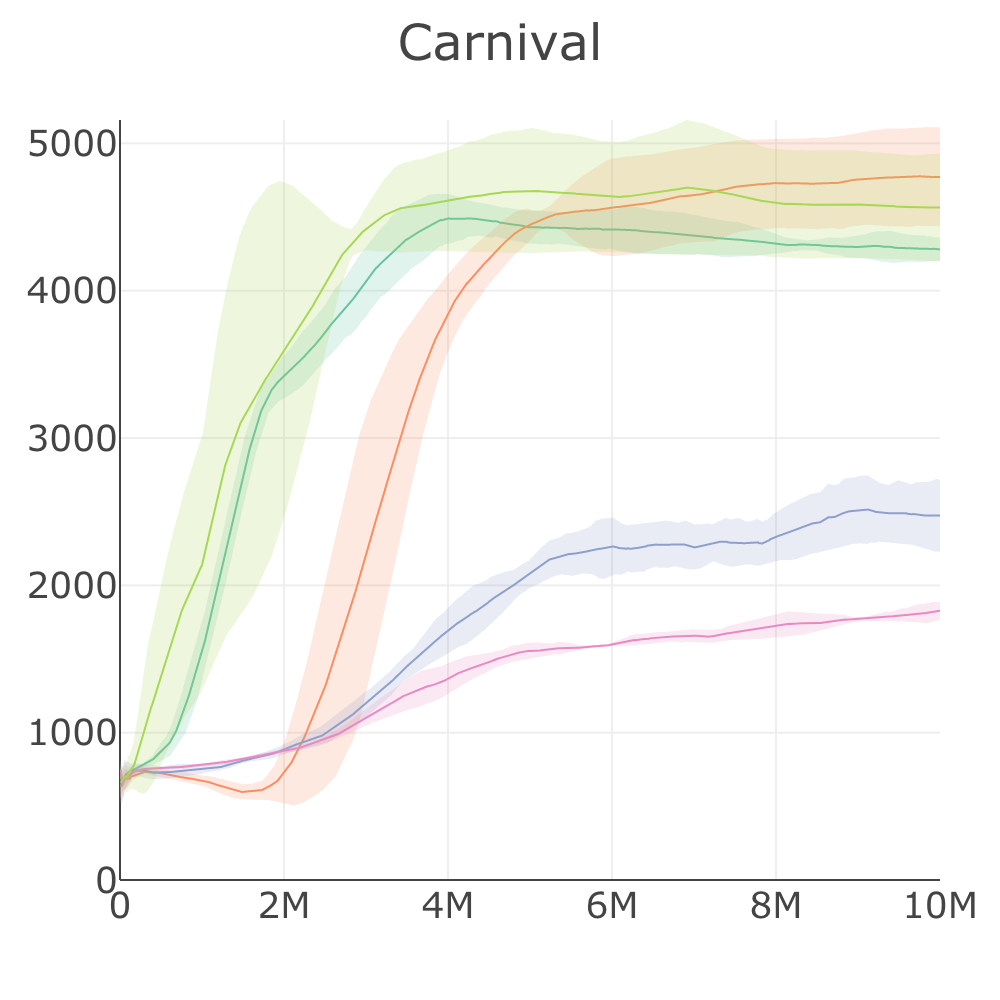}\\
\includegraphics[width=1.22in]{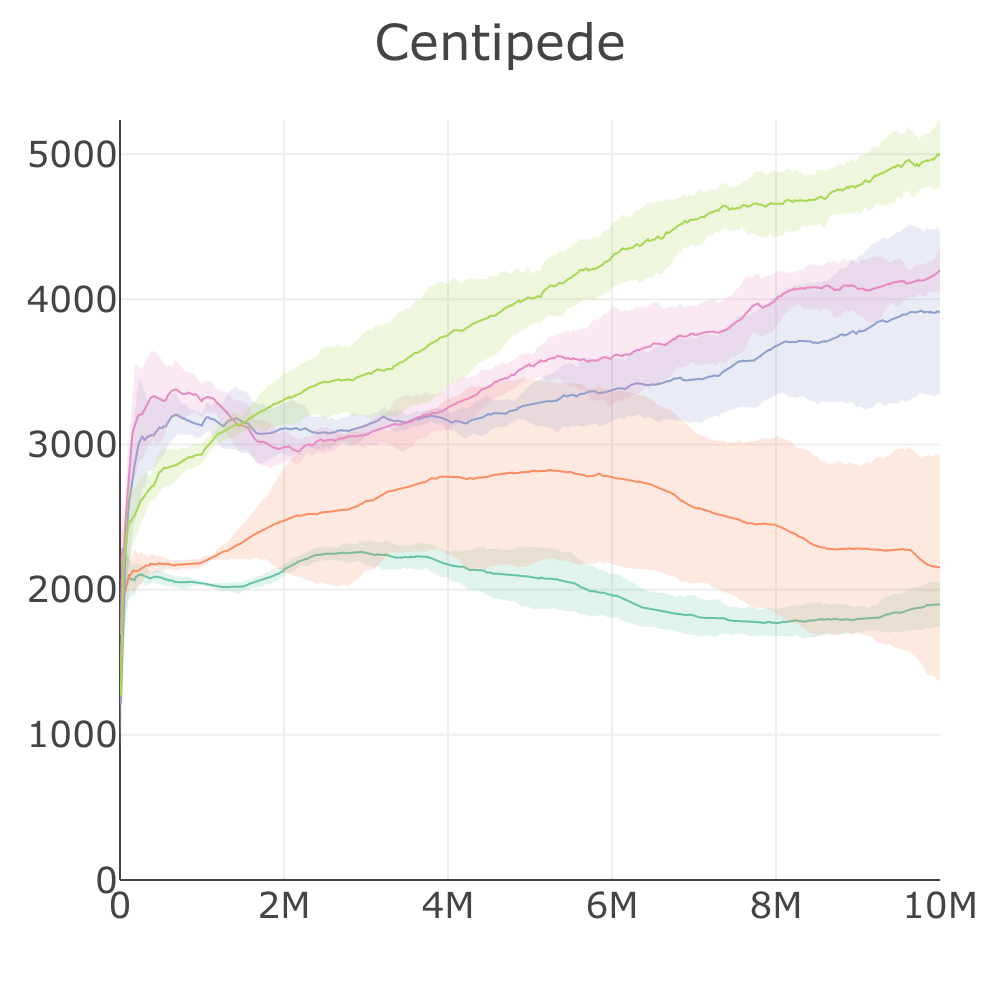} & 
\includegraphics[width=1.22in]{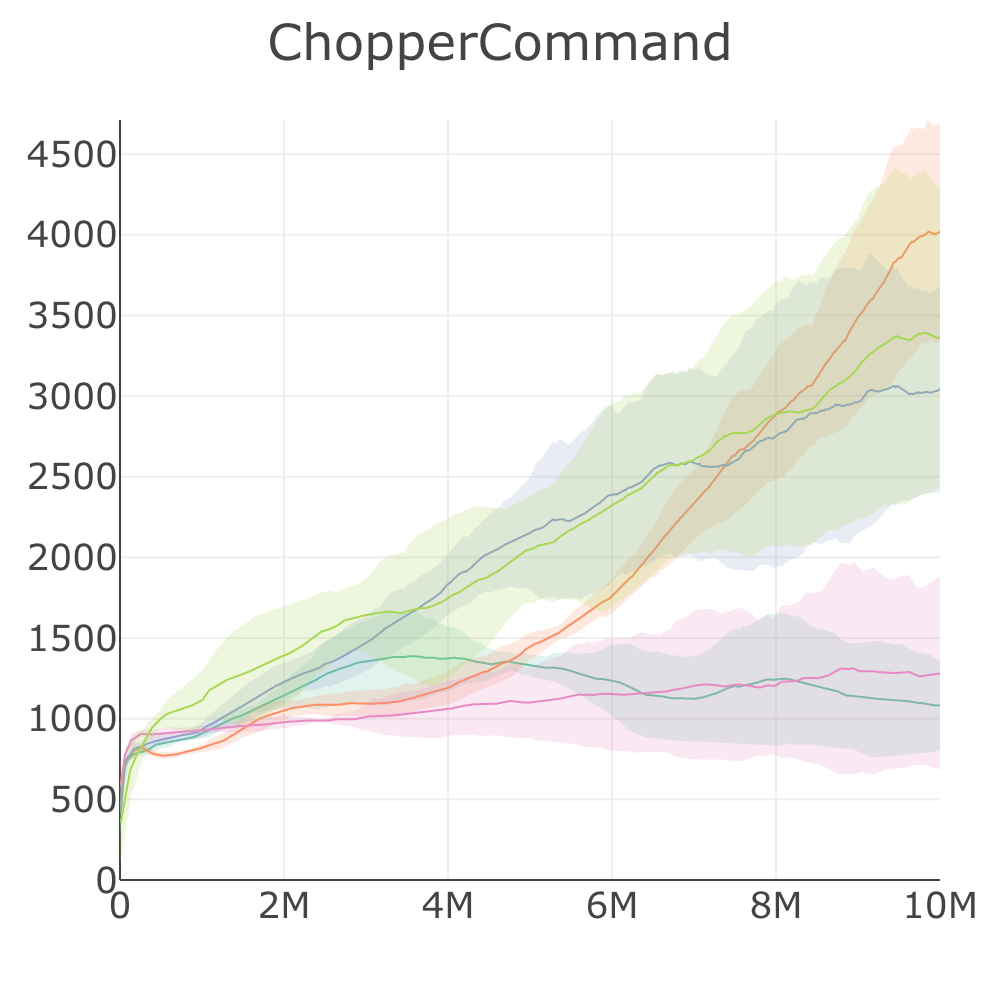} & 
\includegraphics[width=1.22in]{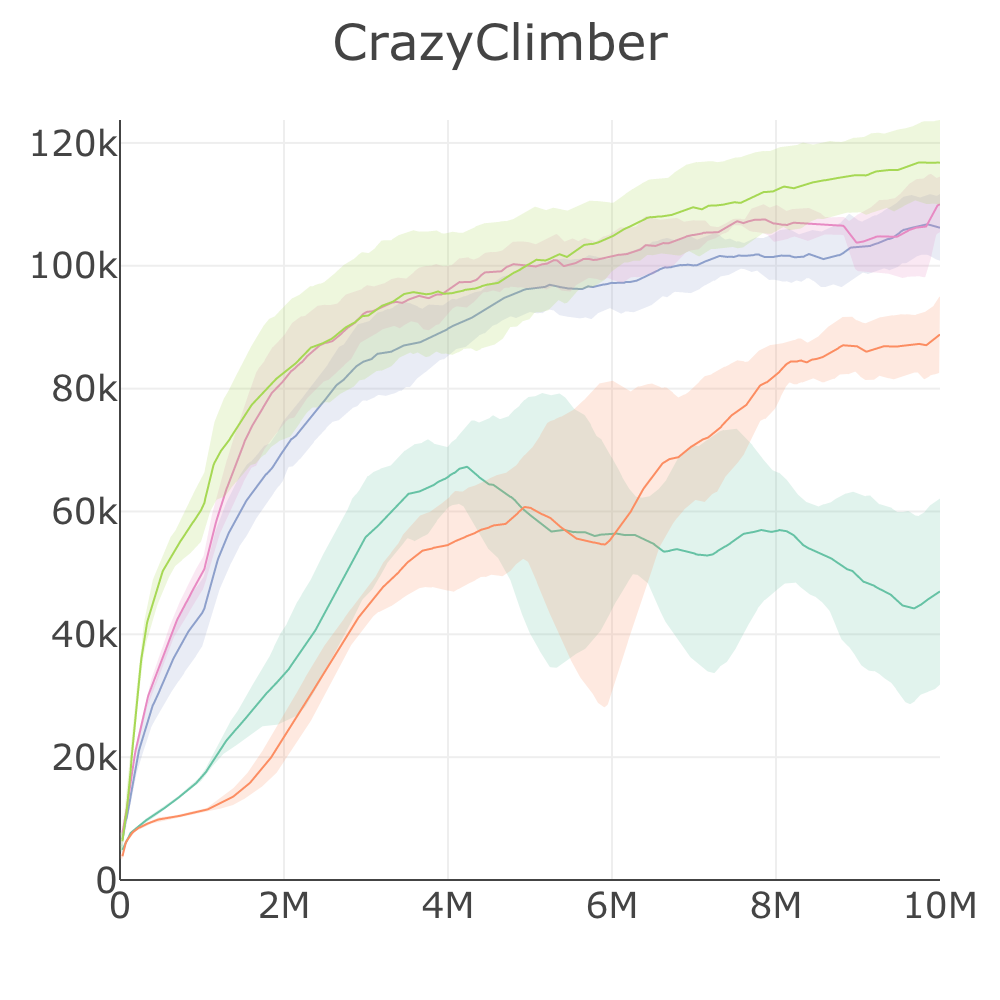} & 
\includegraphics[width=1.22in]{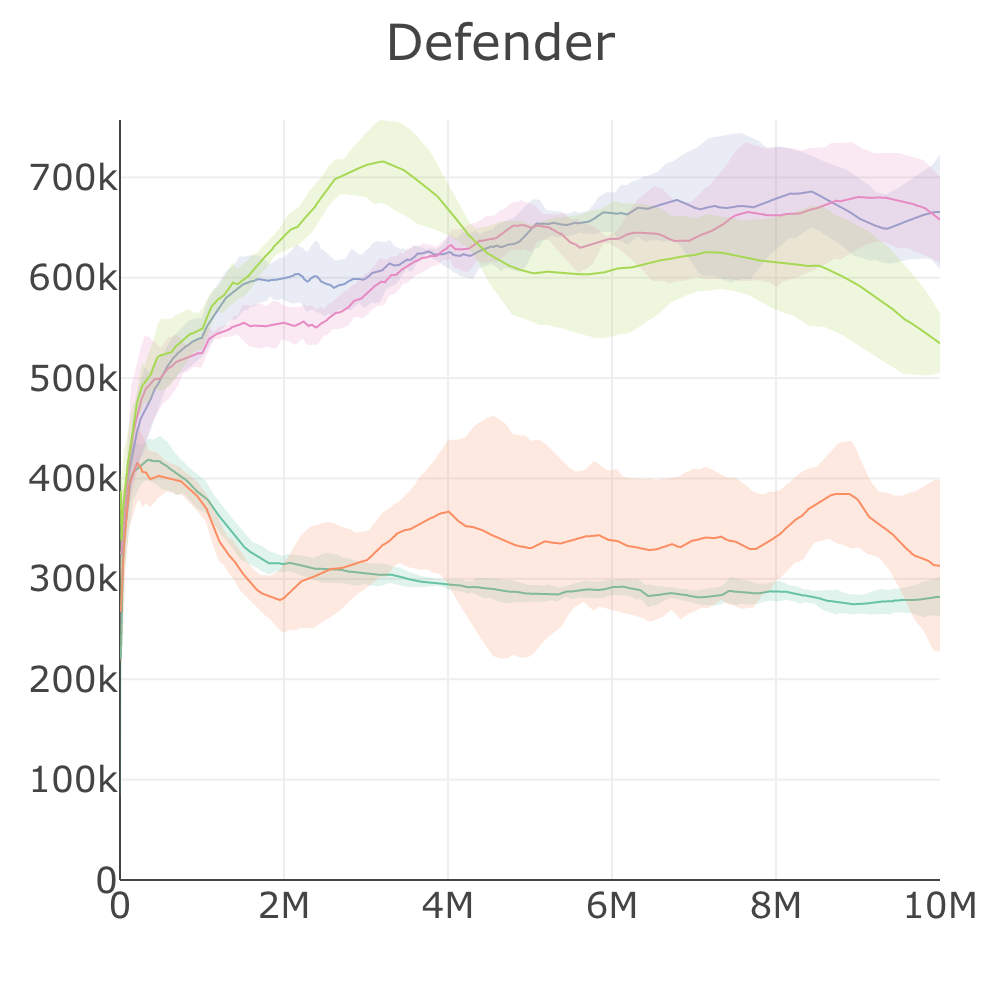}\\
\includegraphics[width=1.22in]{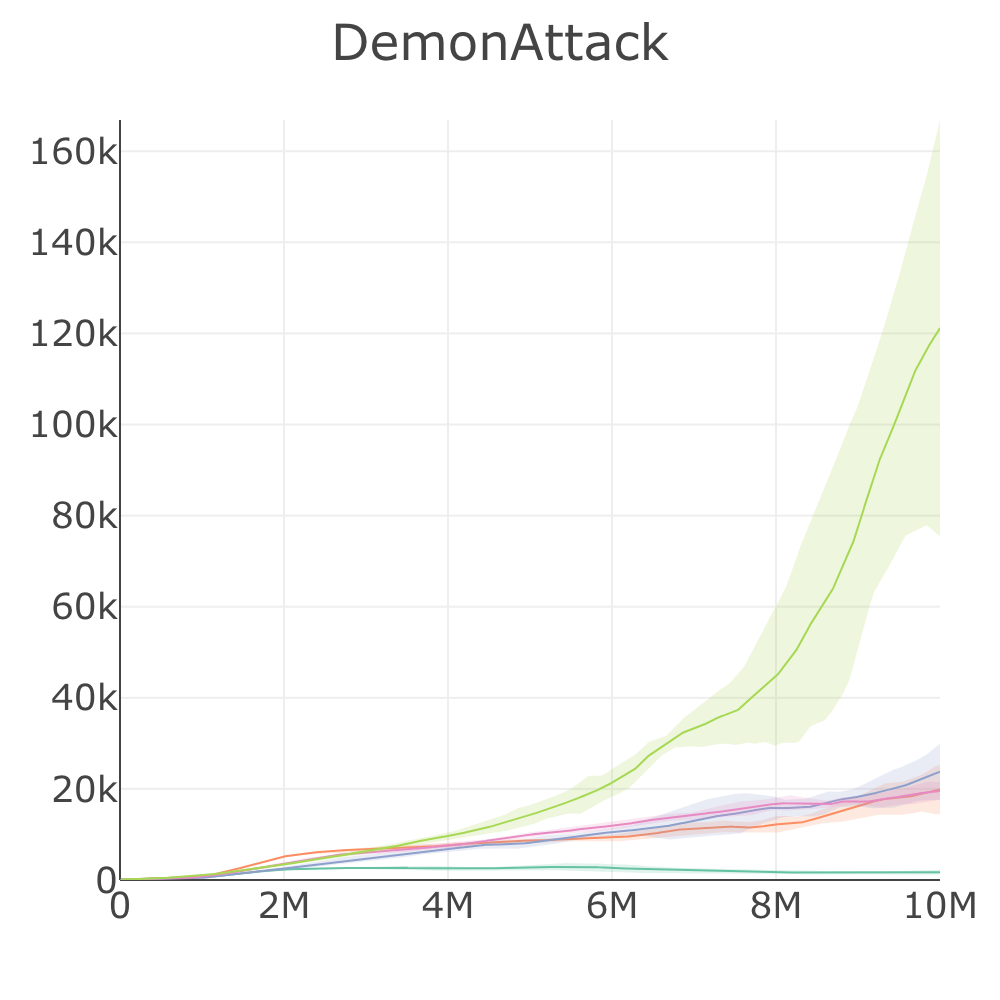} & 
\includegraphics[width=1.22in]{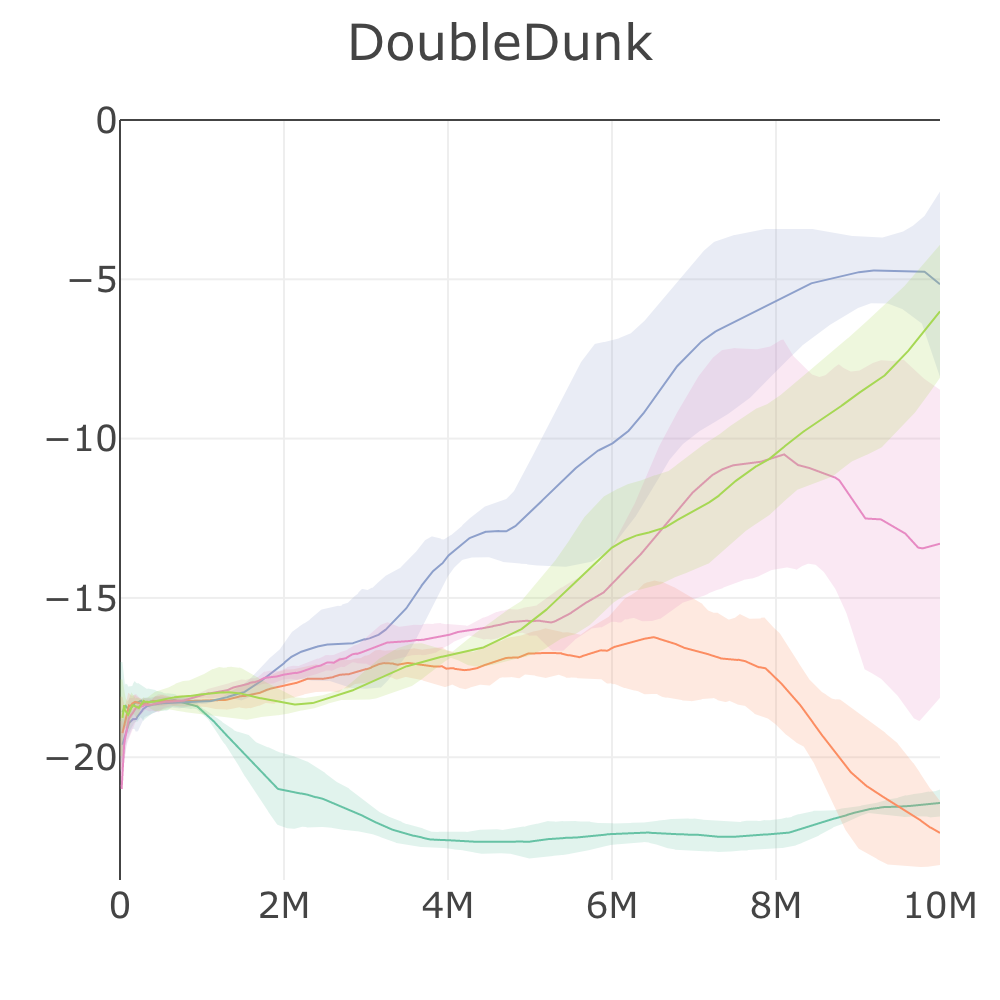} & 
\includegraphics[width=1.22in]{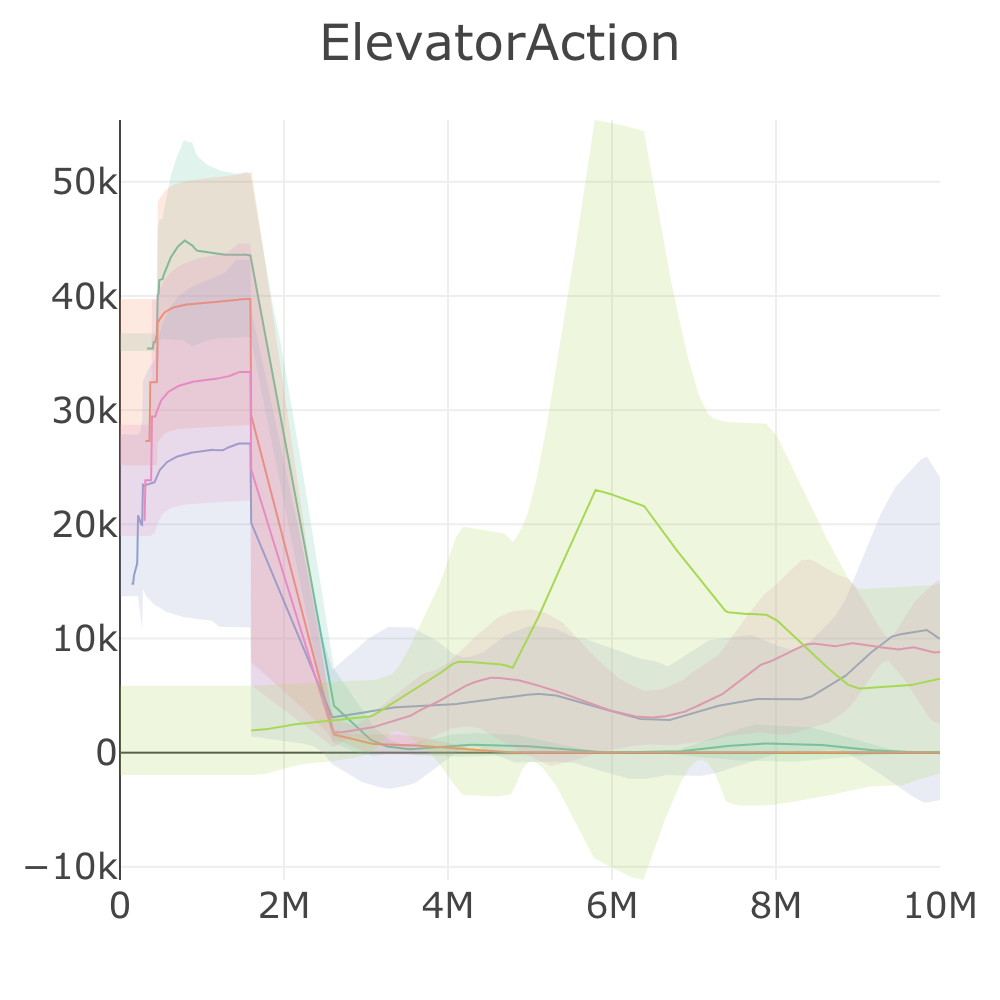} & 
\includegraphics[width=1.22in]{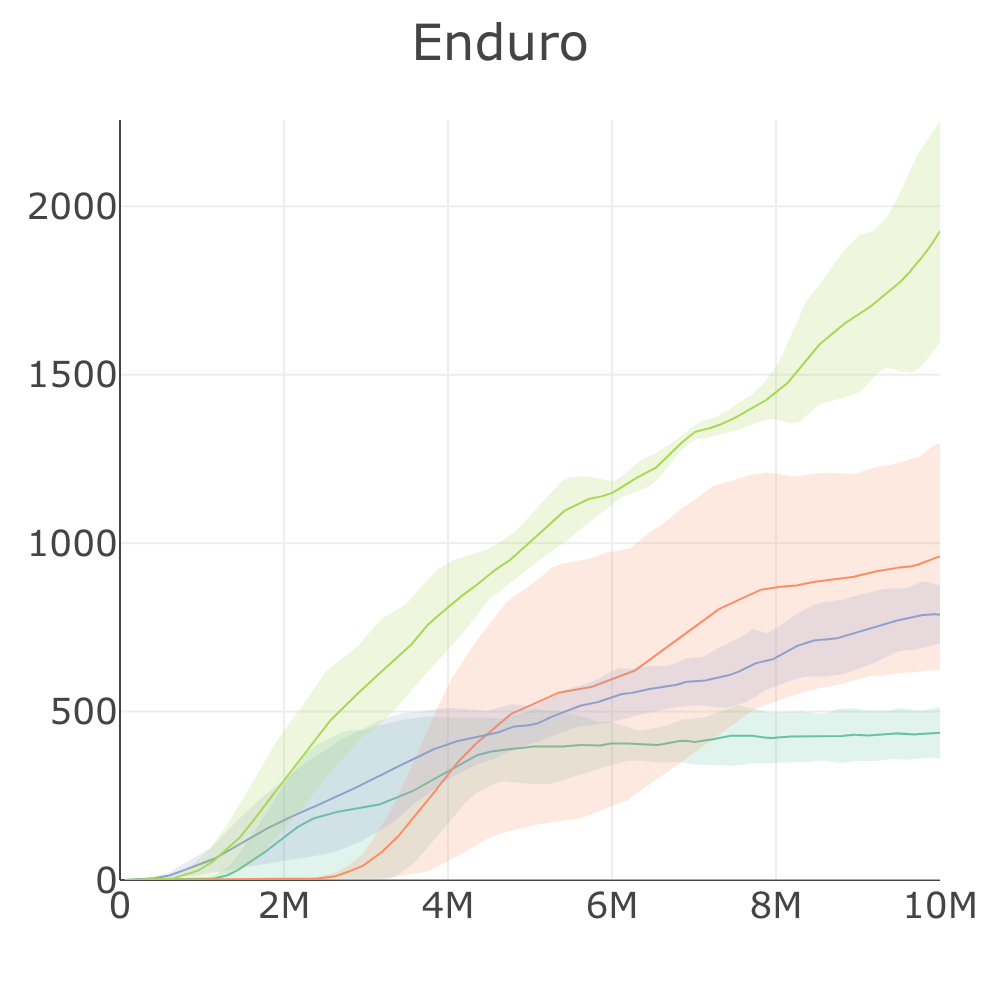}\\
\includegraphics[width=1.22in]{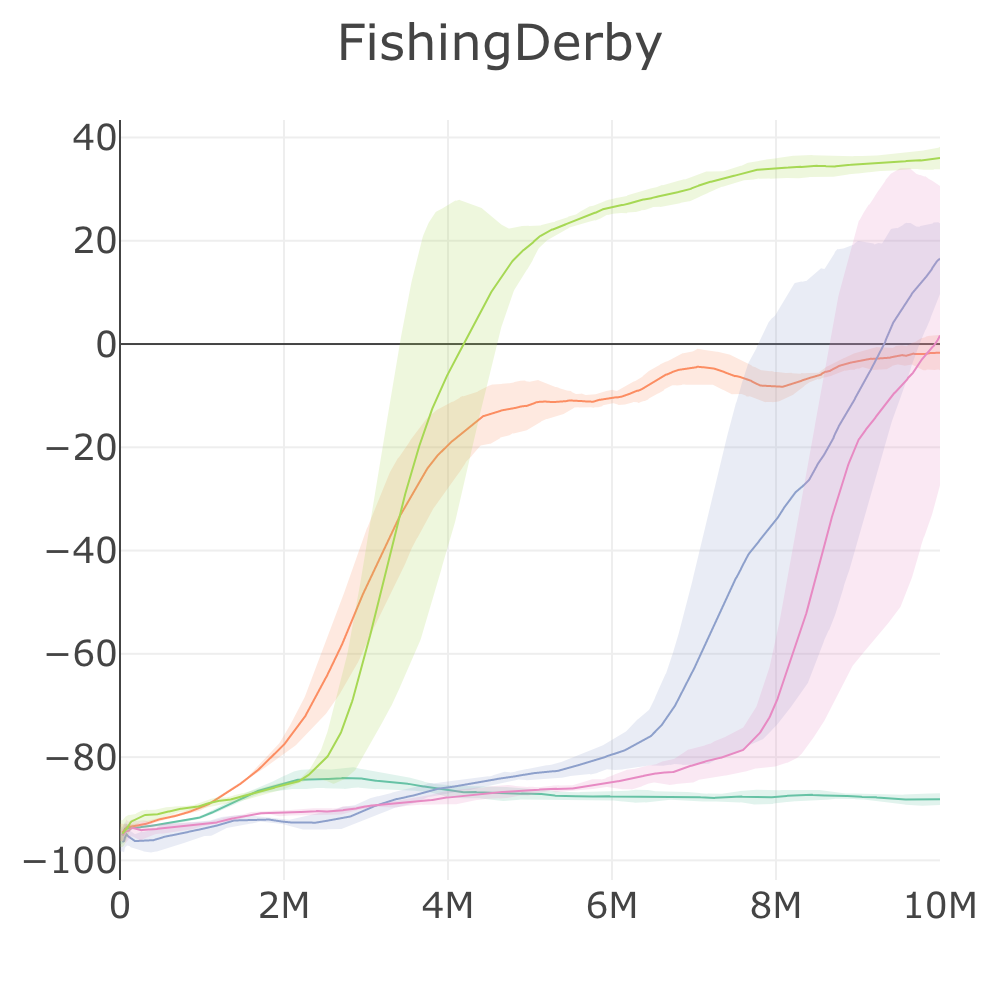} & 
\includegraphics[width=1.22in]{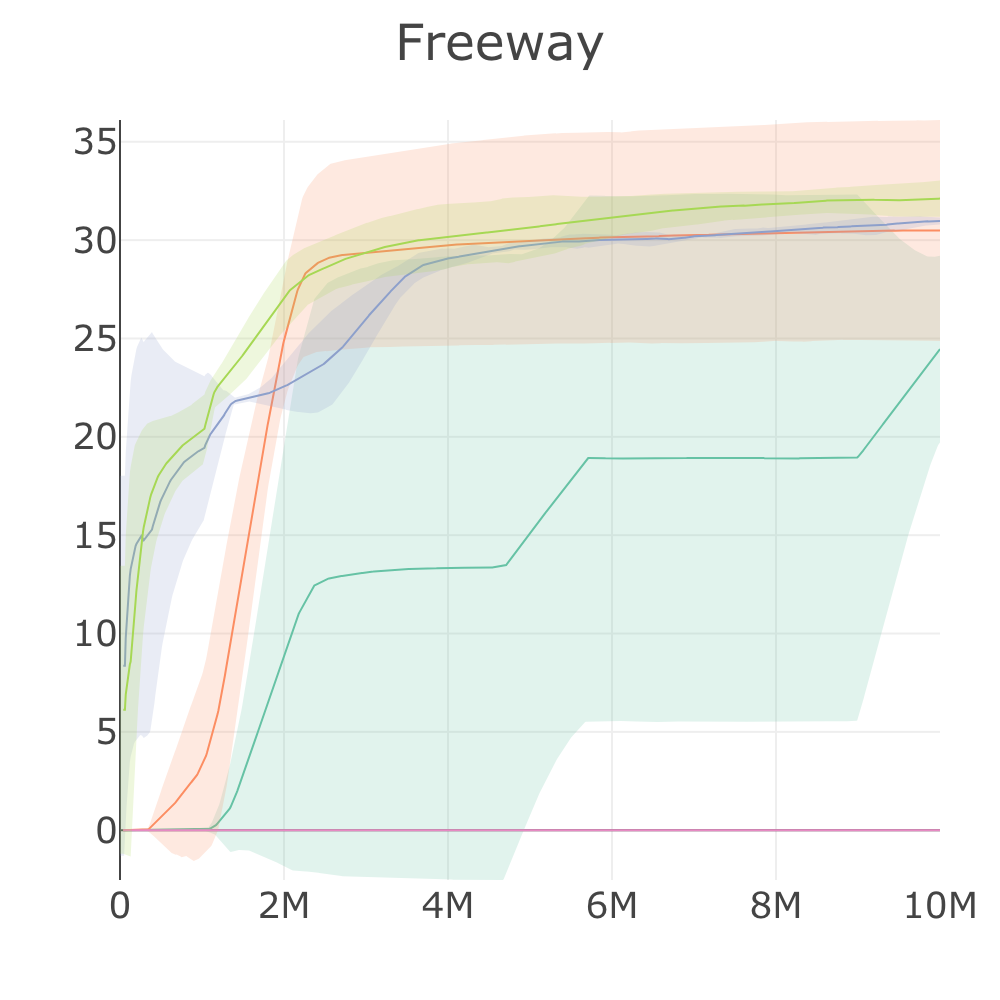} & 
\includegraphics[width=1.22in]{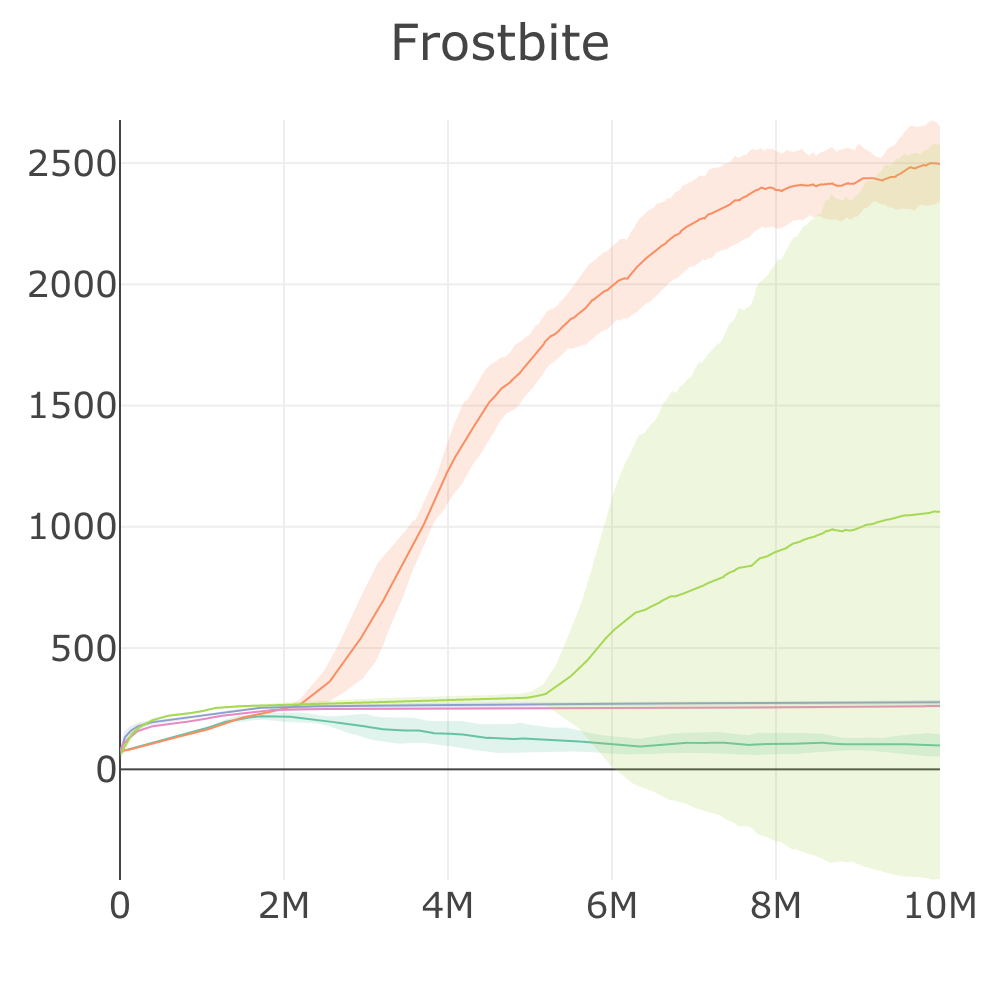} & 
\includegraphics[width=1.22in]{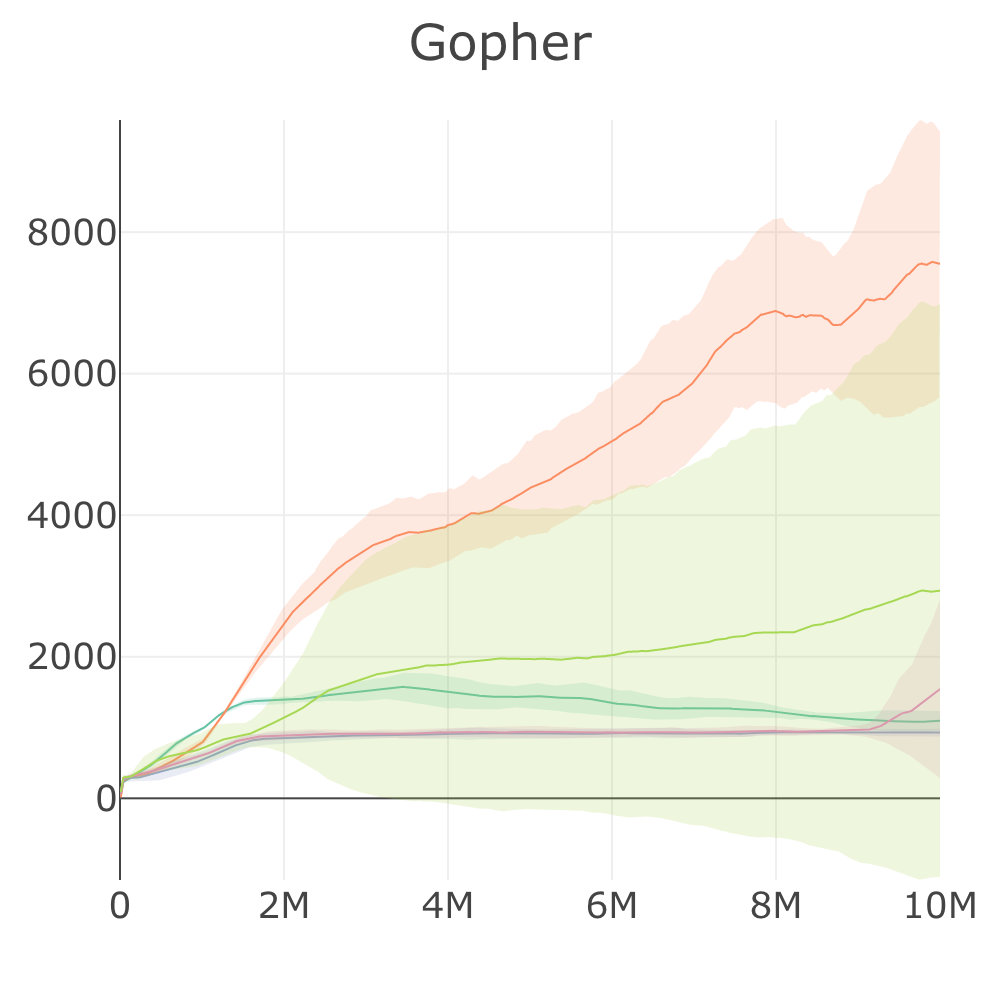}\\
\includegraphics[width=1.22in]{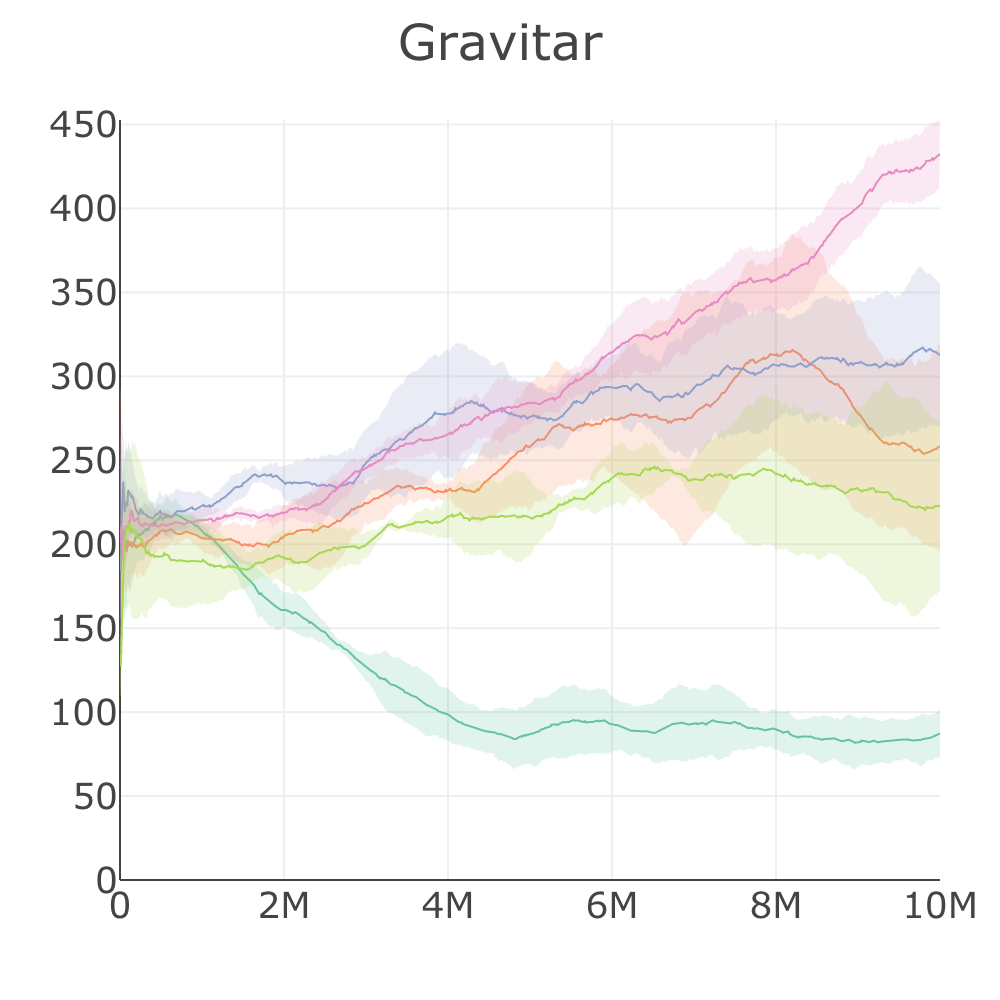} & 
\includegraphics[width=1.22in]{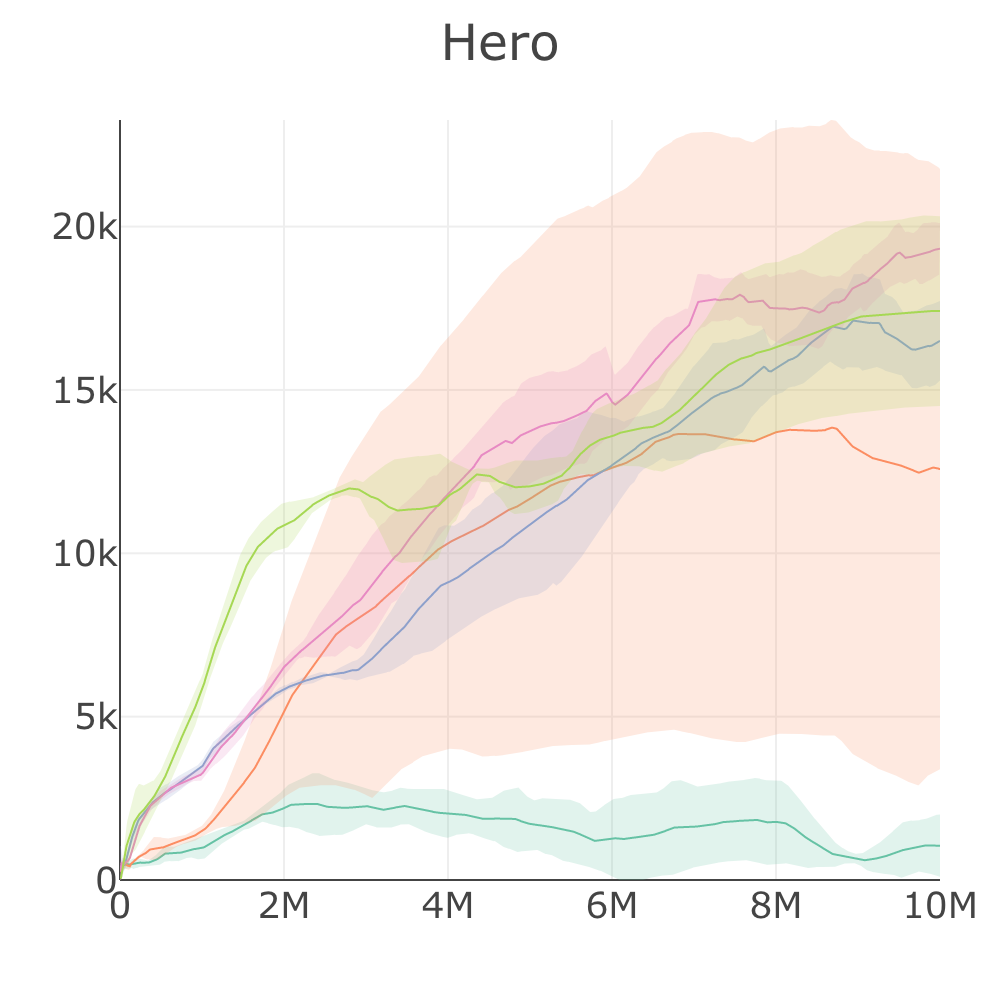} & 
\includegraphics[width=1.22in]{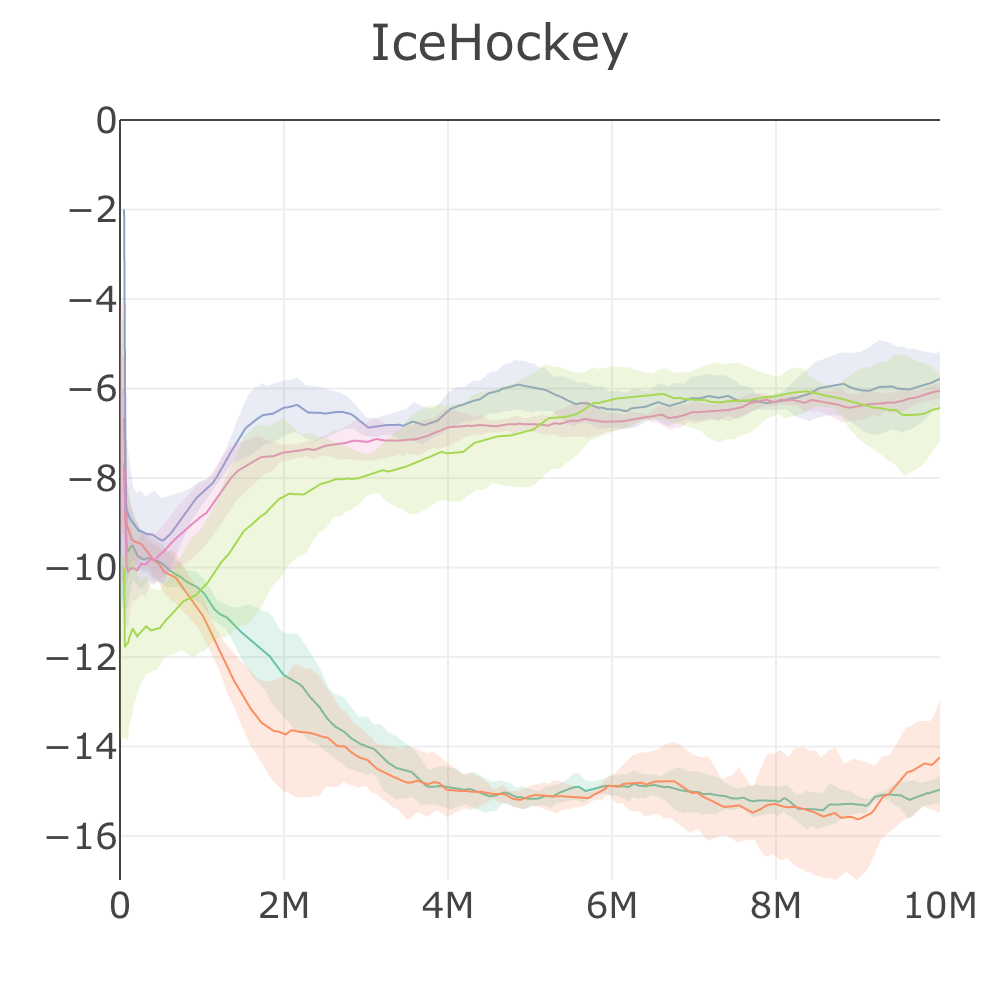} & 
\includegraphics[width=1.22in]{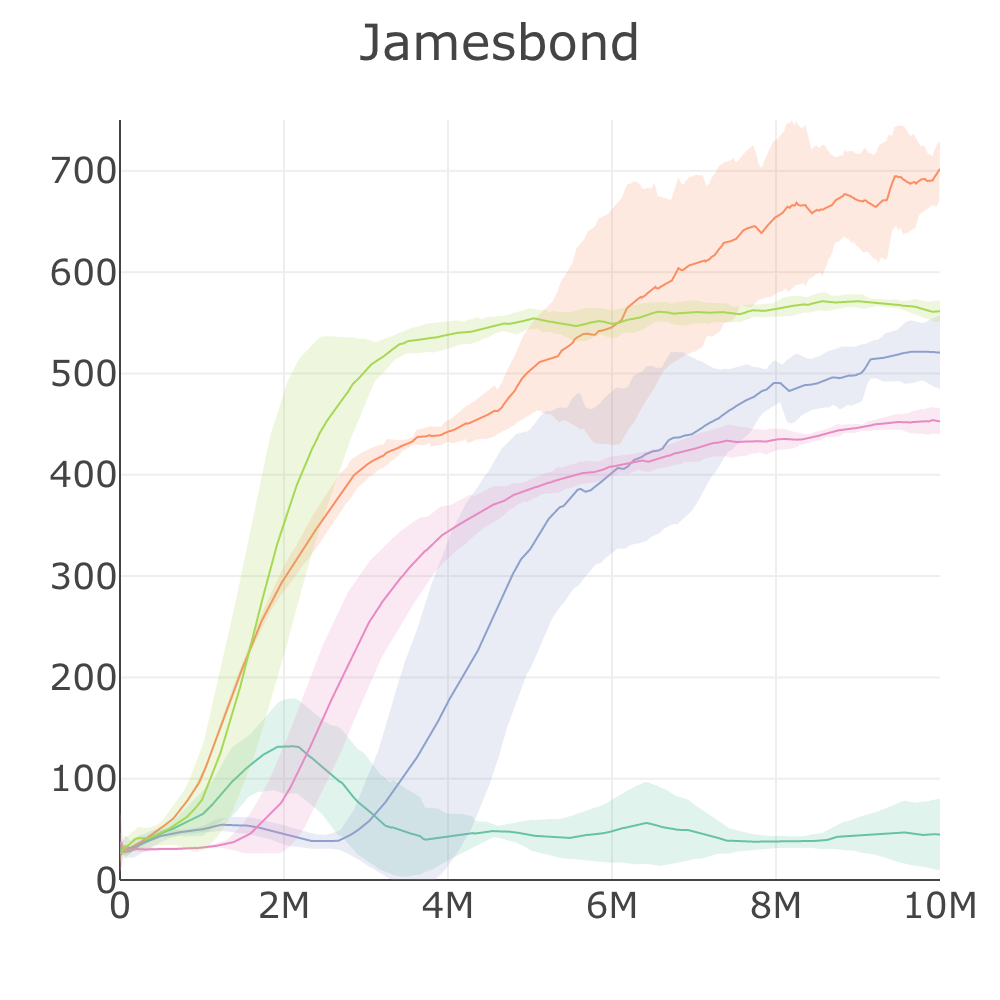}\\
\includegraphics[width=1.22in]{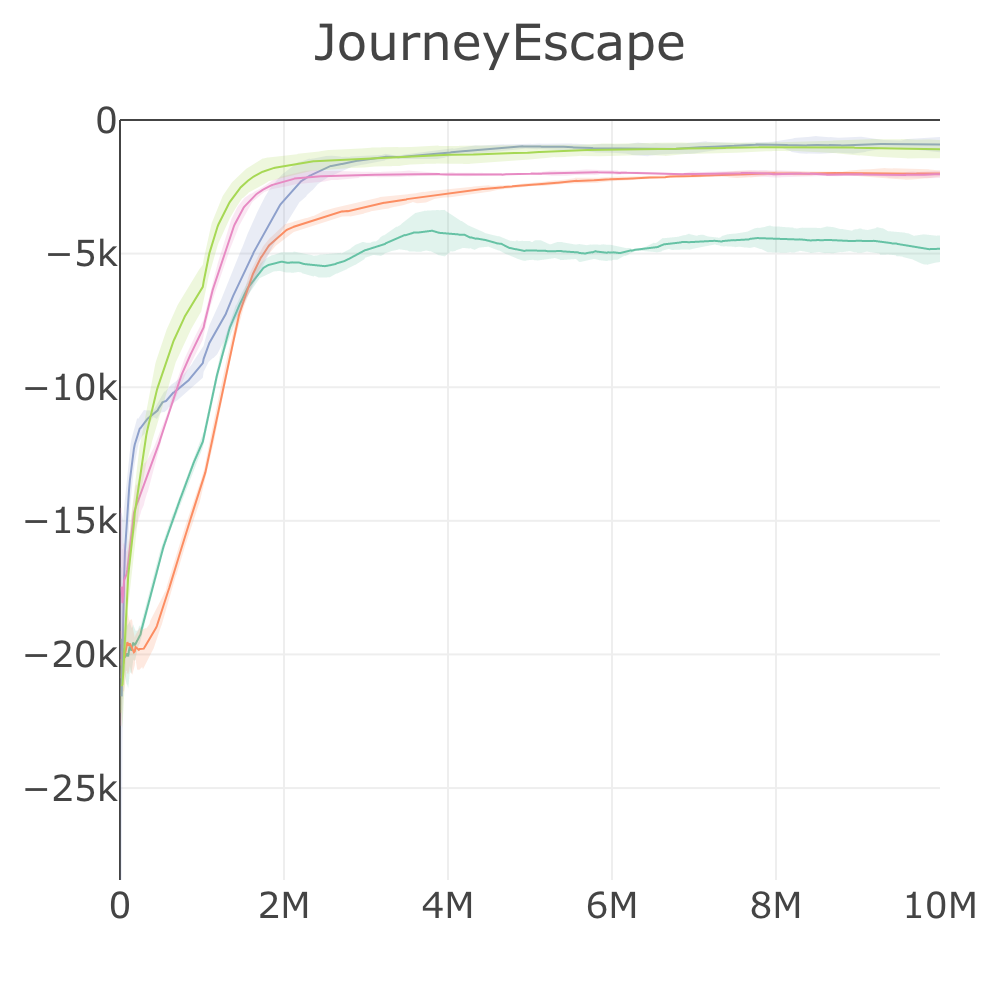} & 
\includegraphics[width=1.22in]{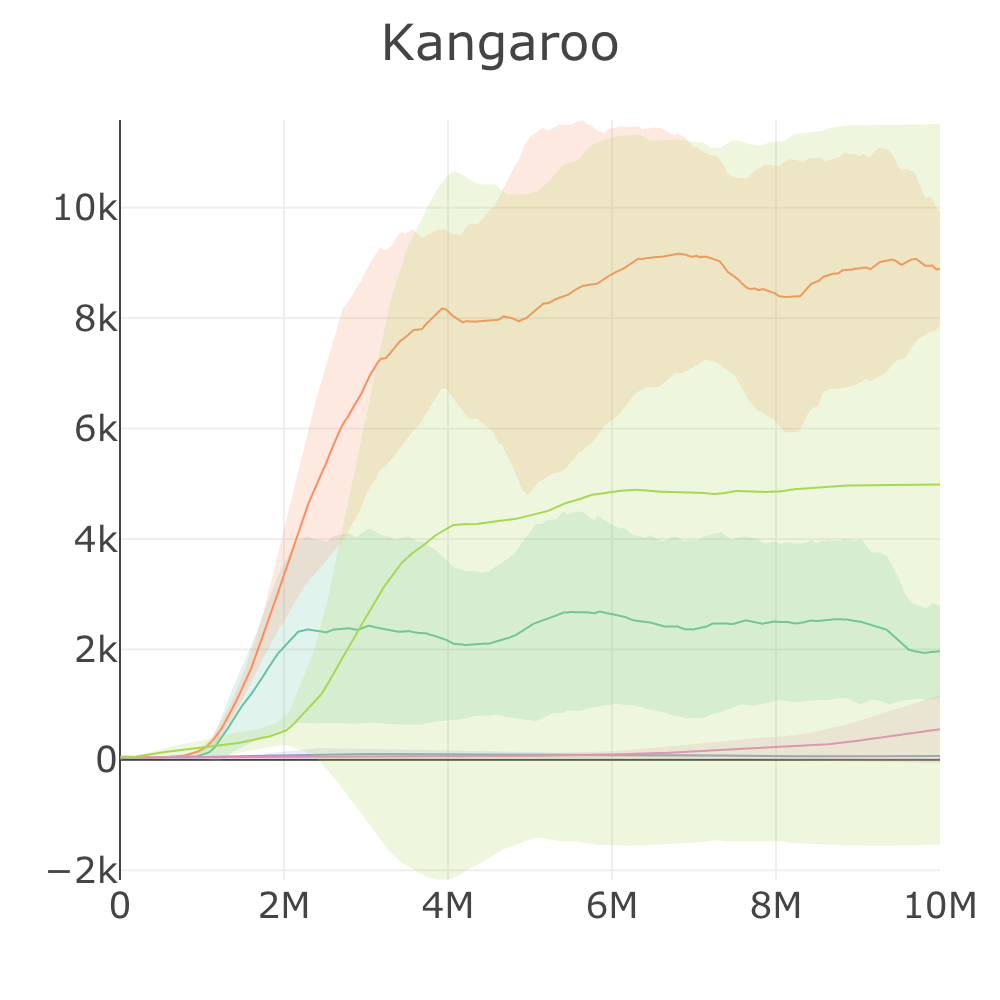} & 
\includegraphics[width=1.22in]{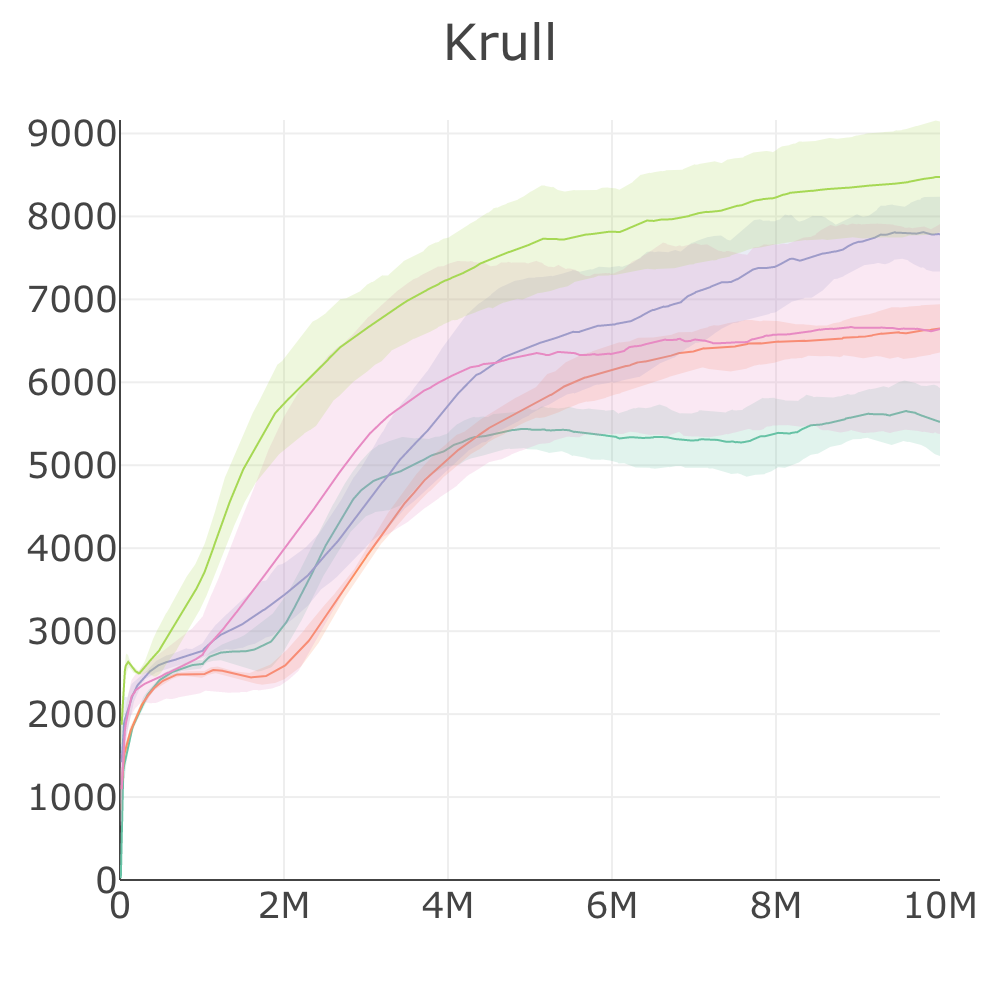} & 
\includegraphics[width=1.22in]{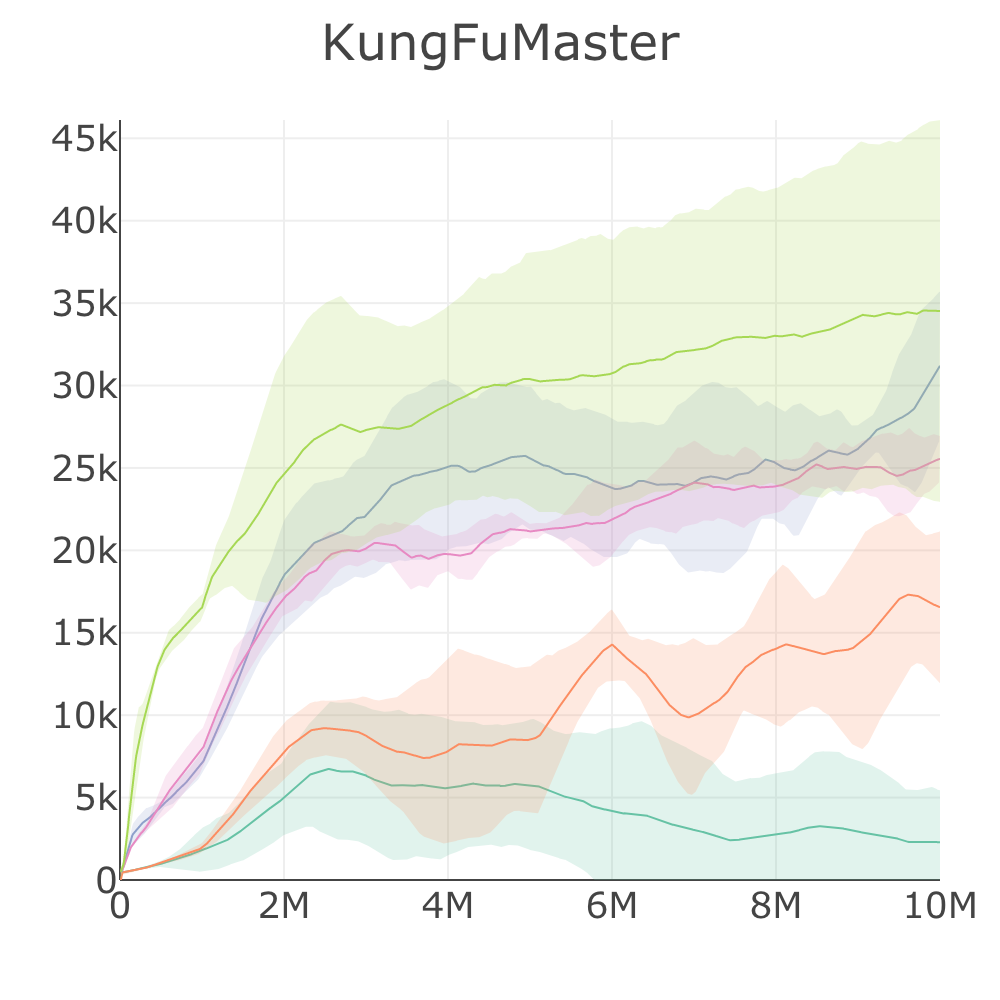}\\
\includegraphics[width=1.22in]{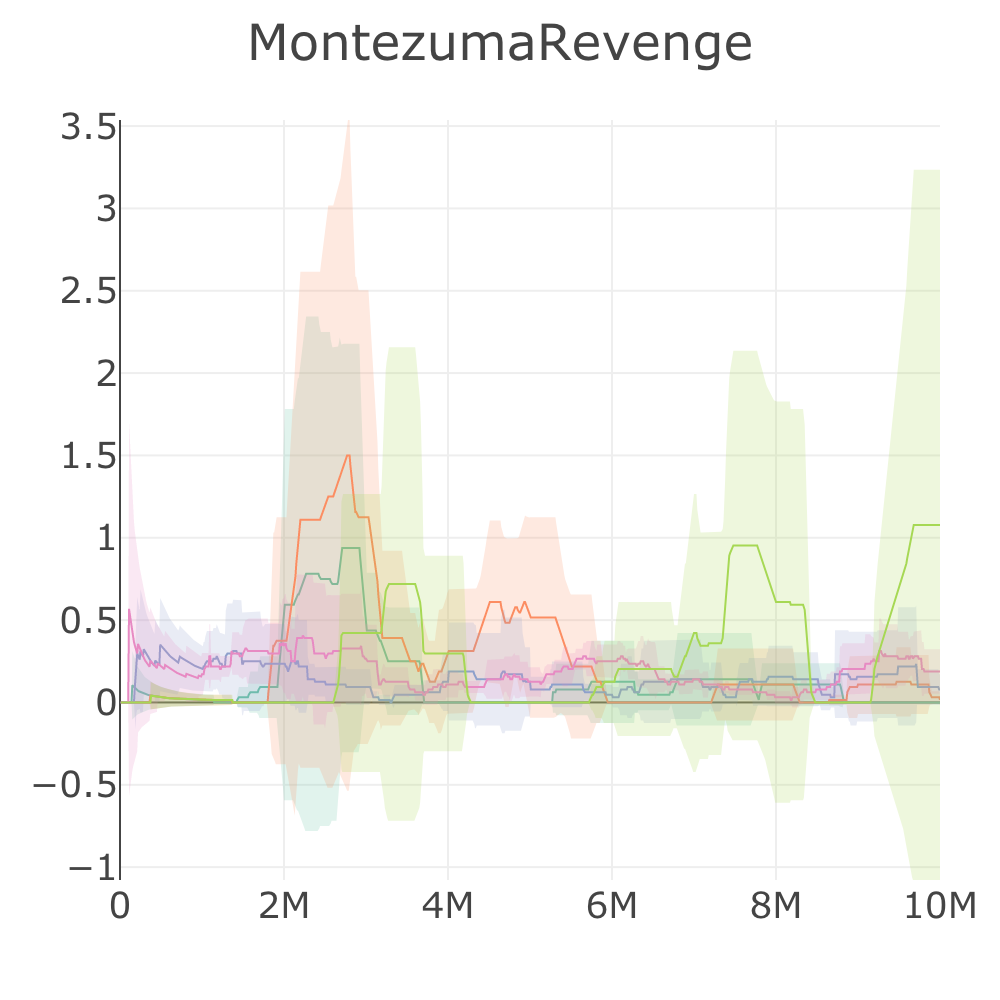} & 
\includegraphics[width=1.22in]{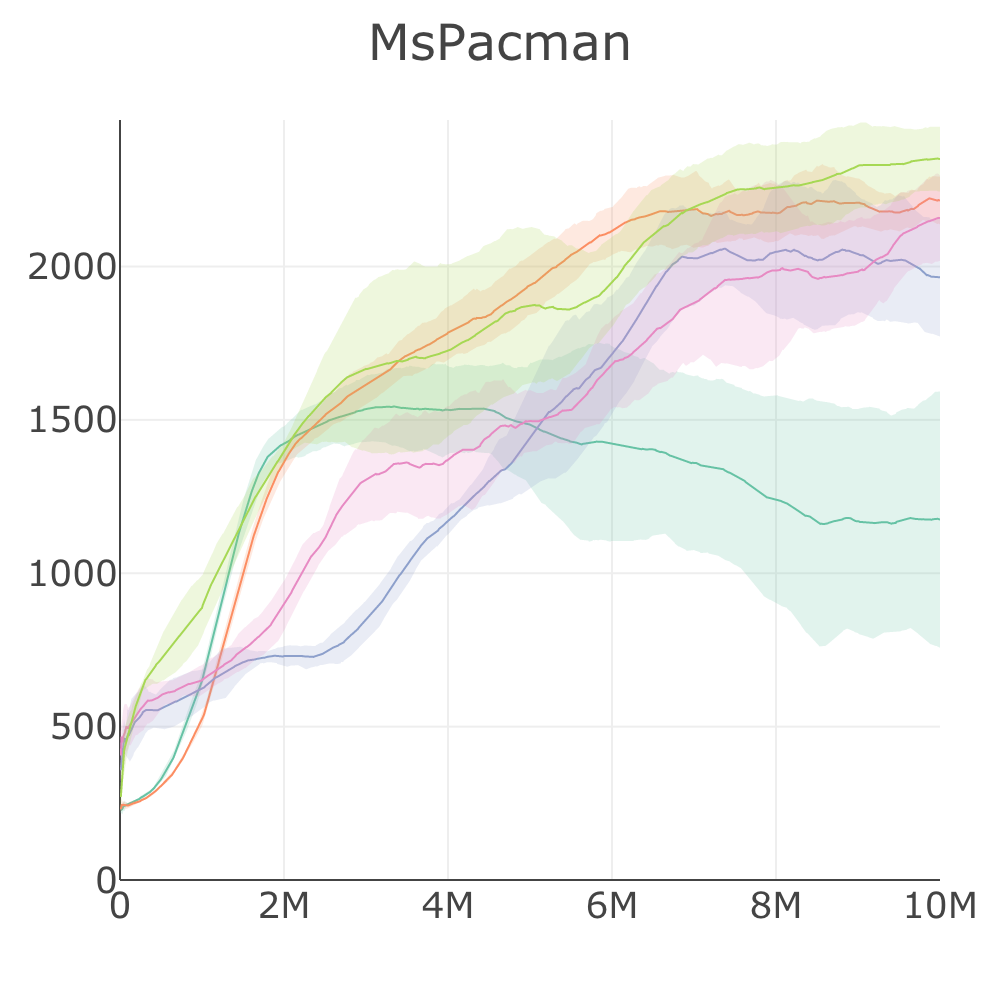} & 
\includegraphics[width=1.22in]{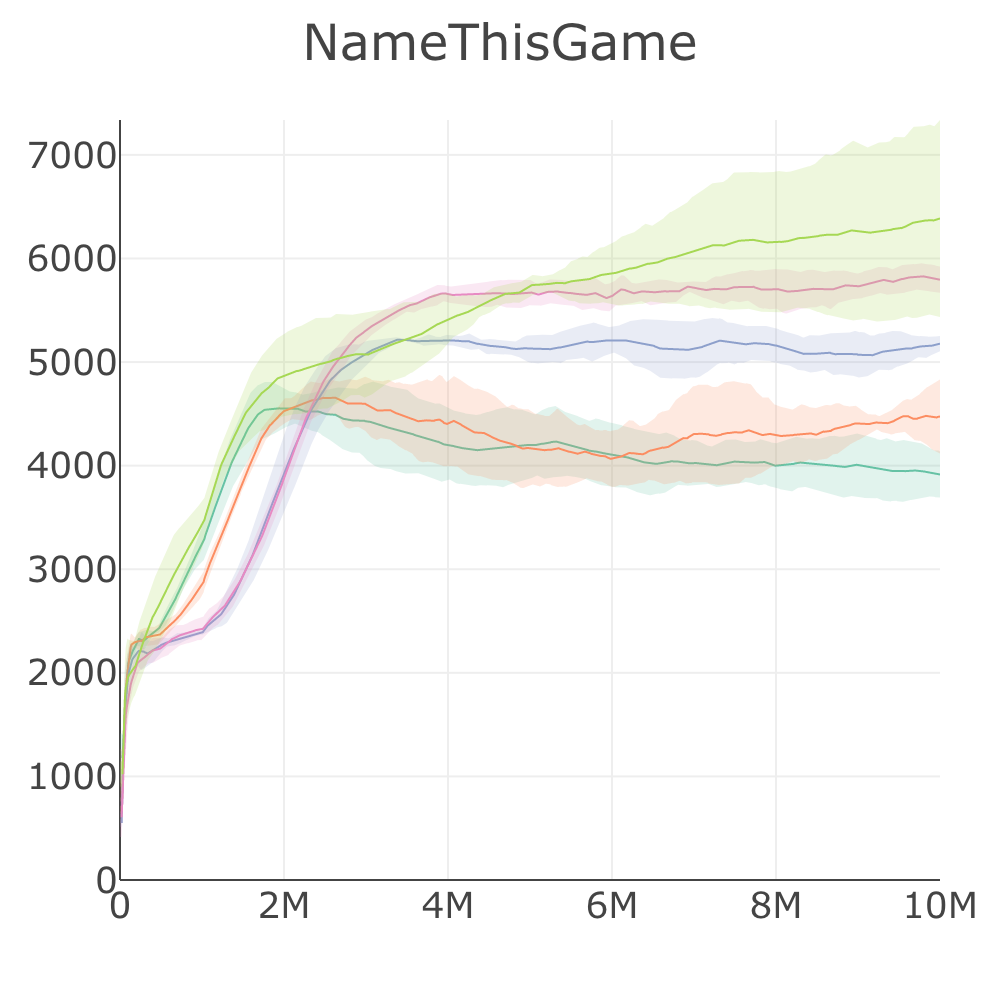} & 
\includegraphics[width=1.22in]{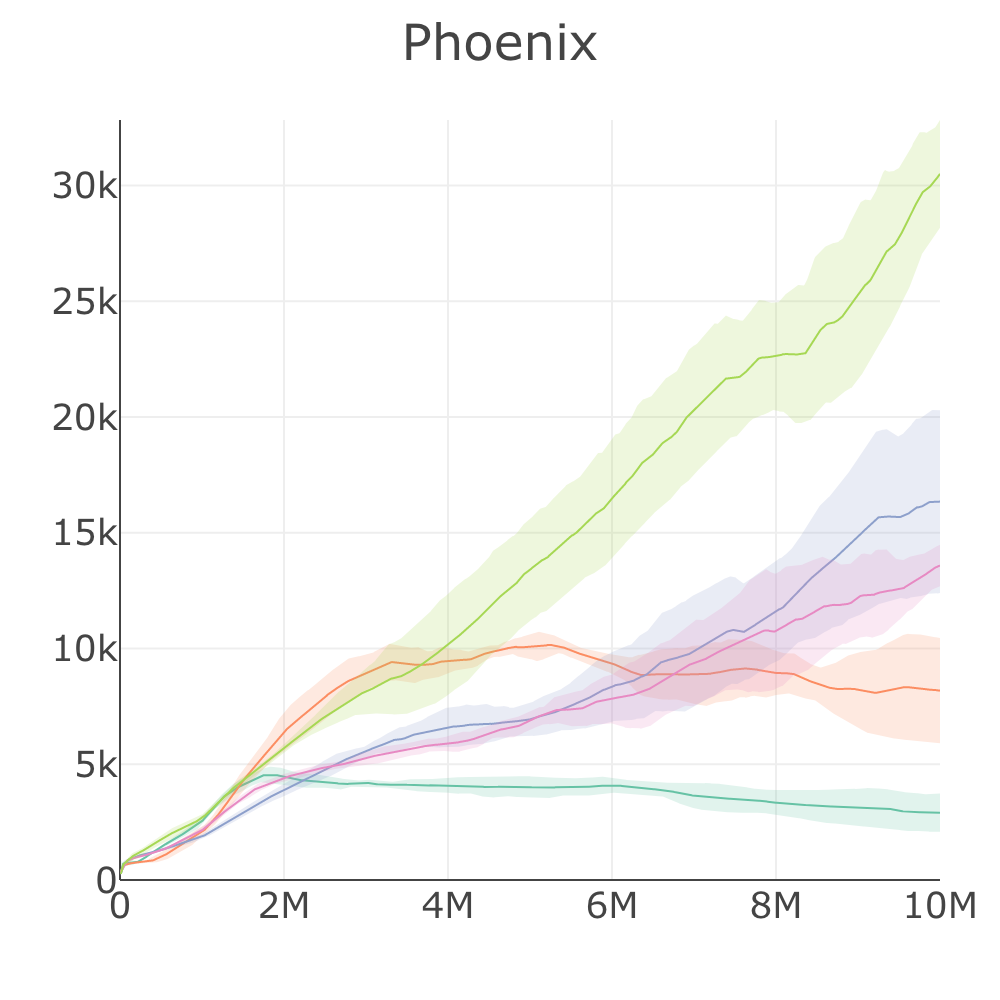}\\
\includegraphics[width=1.22in]{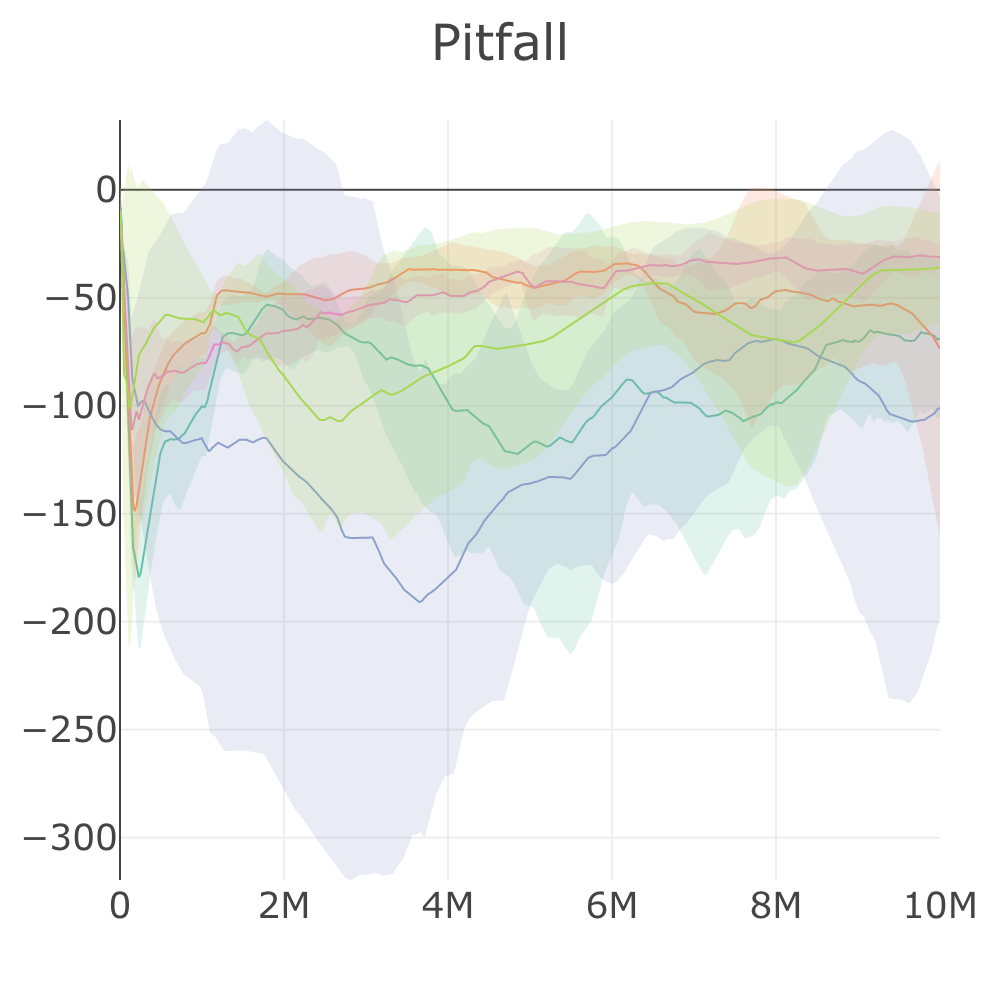} & 
\includegraphics[width=1.22in]{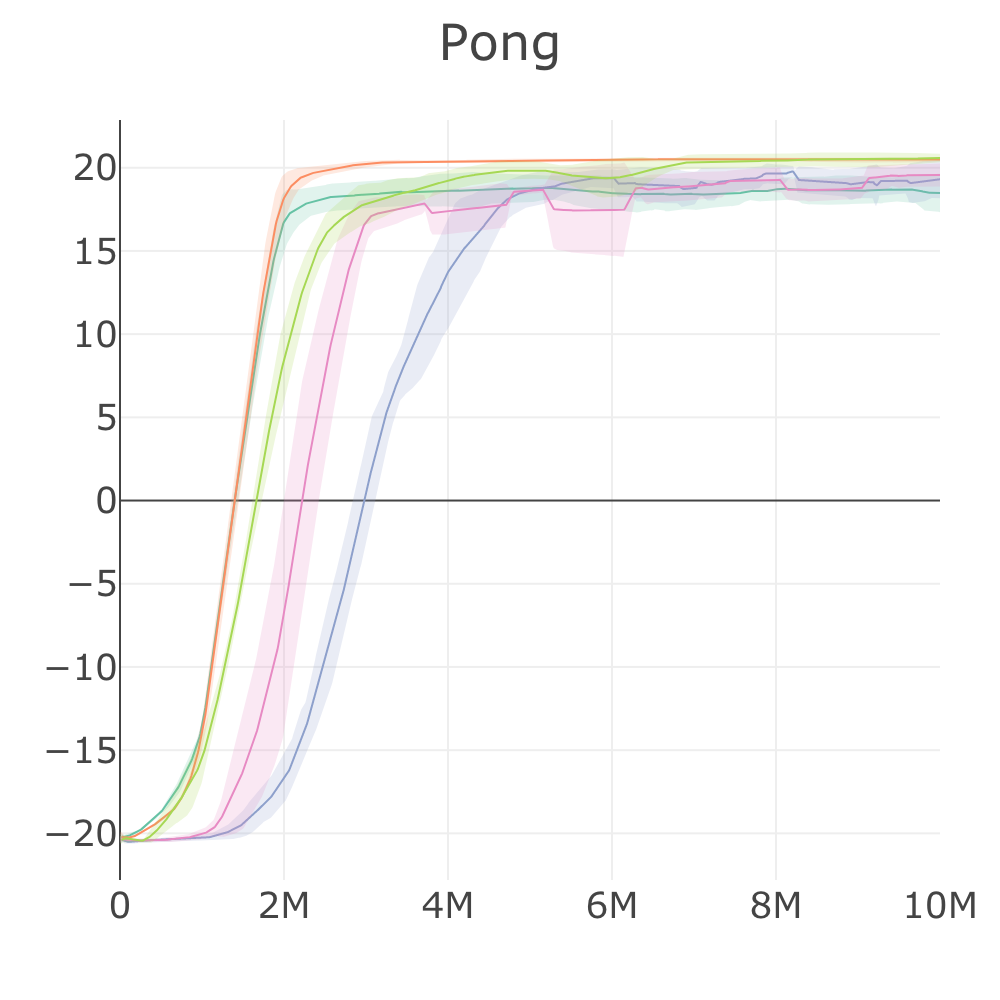} & 
\includegraphics[width=1.22in]{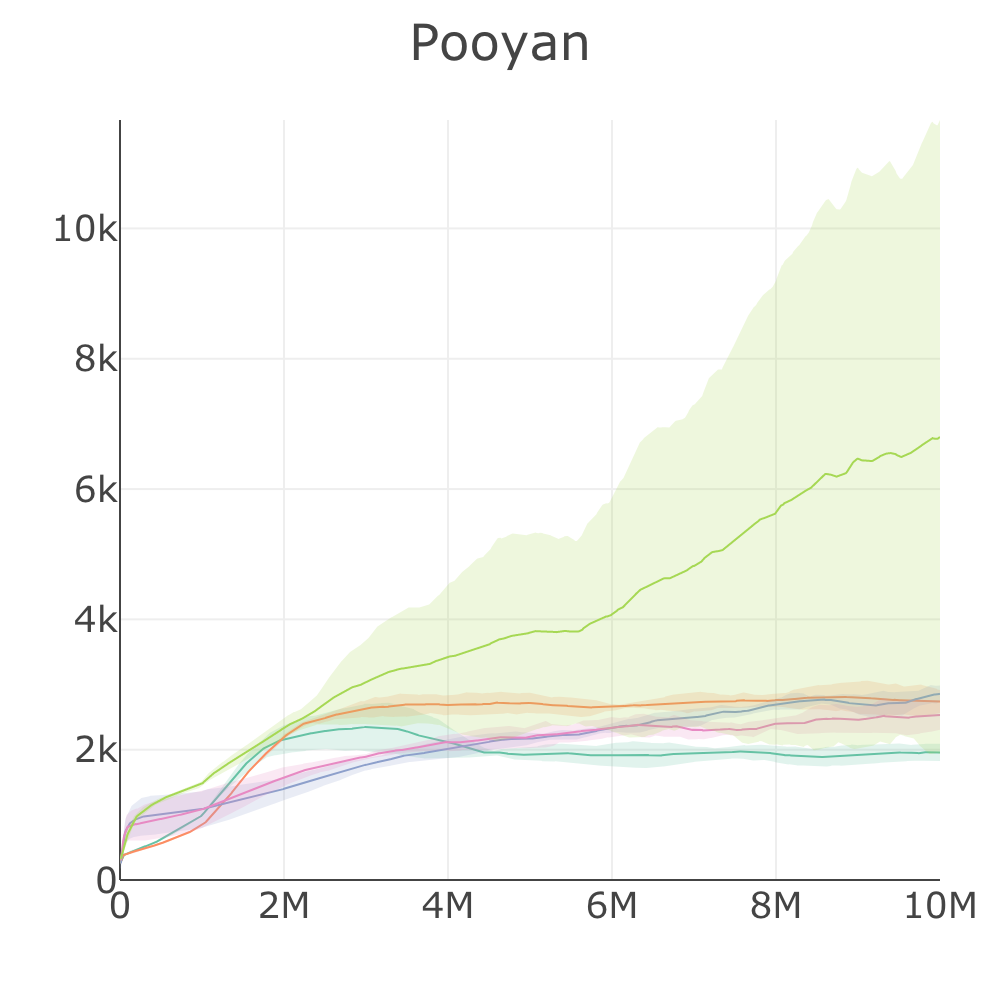} & 
\includegraphics[width=1.22in]{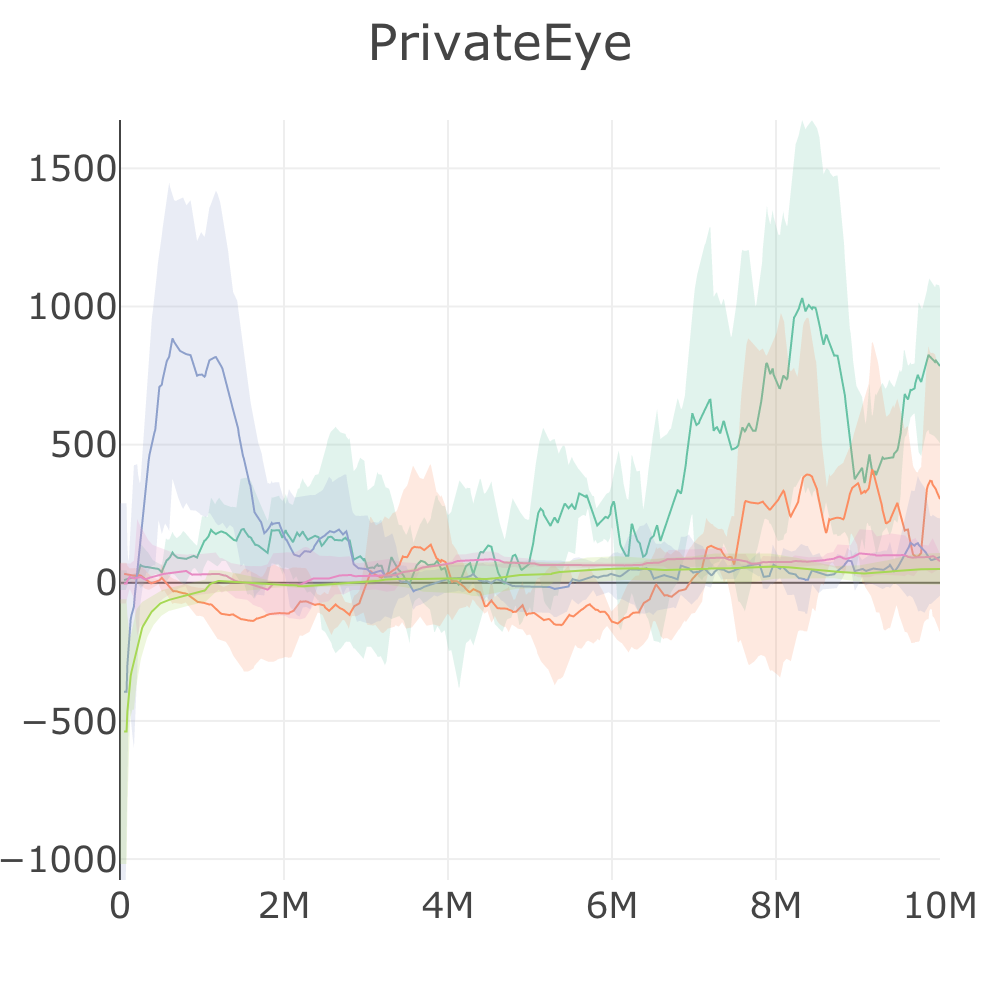}\\
\includegraphics[width=1.22in]{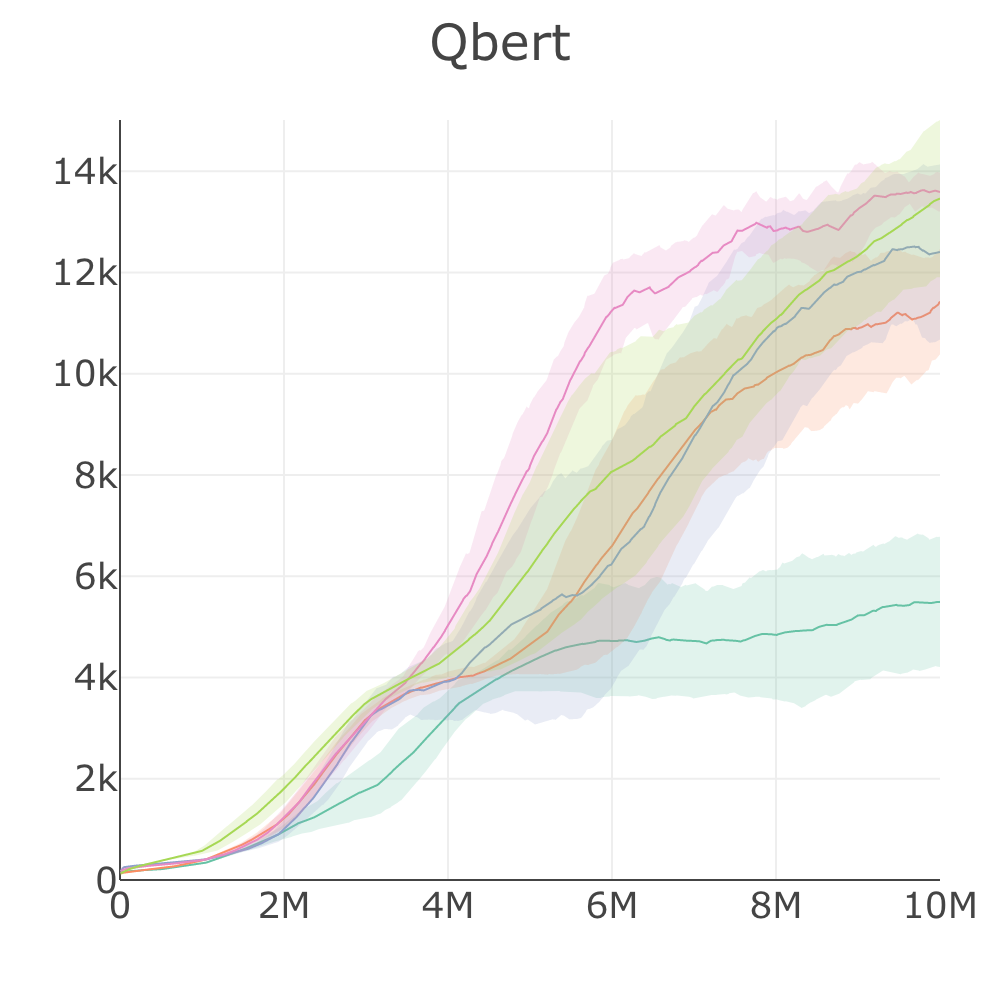} & 
\includegraphics[width=1.22in]{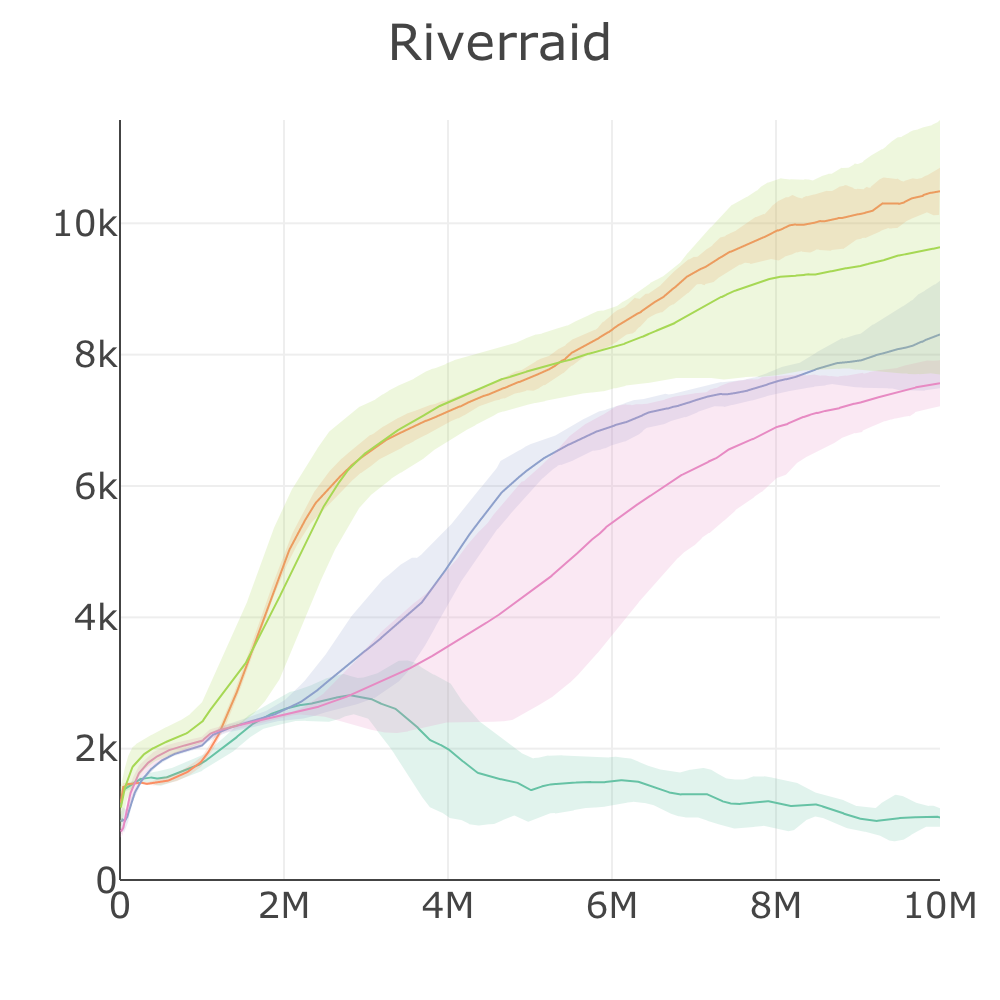} & 
\includegraphics[width=1.22in]{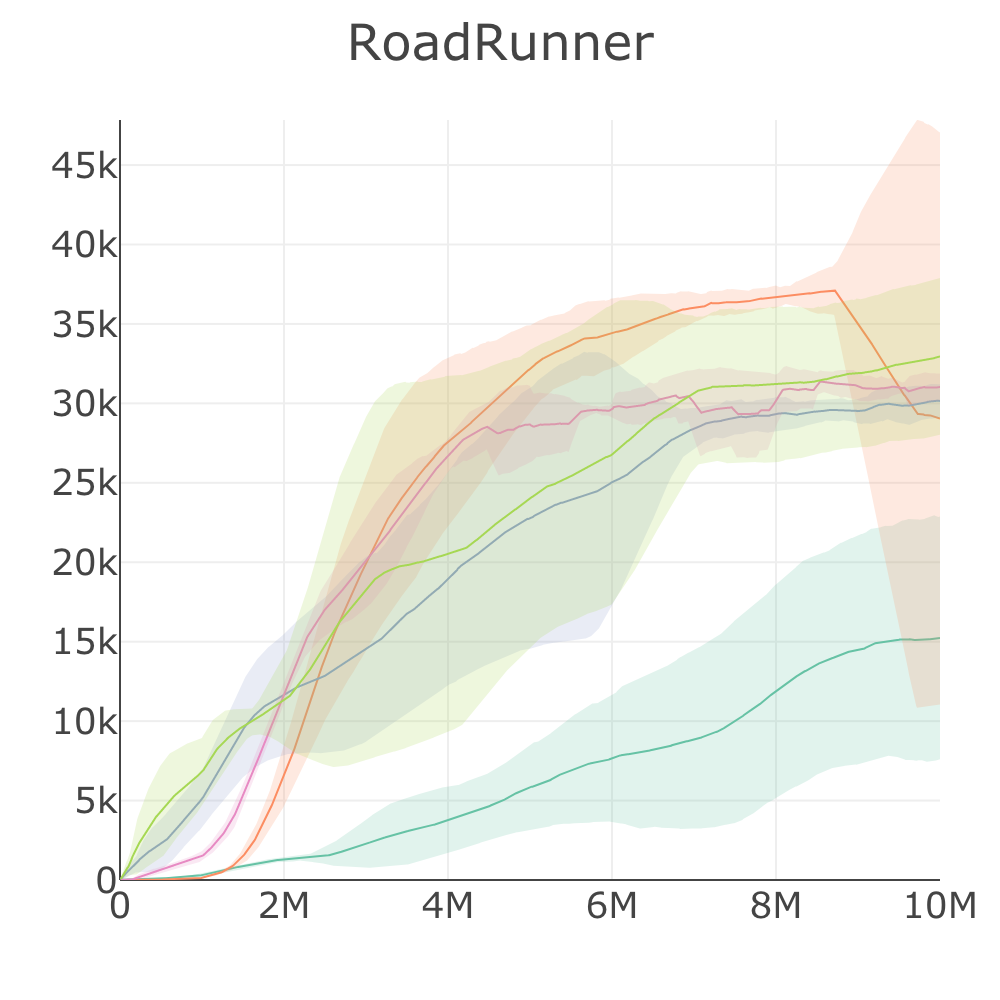} & 
\includegraphics[width=1.22in]{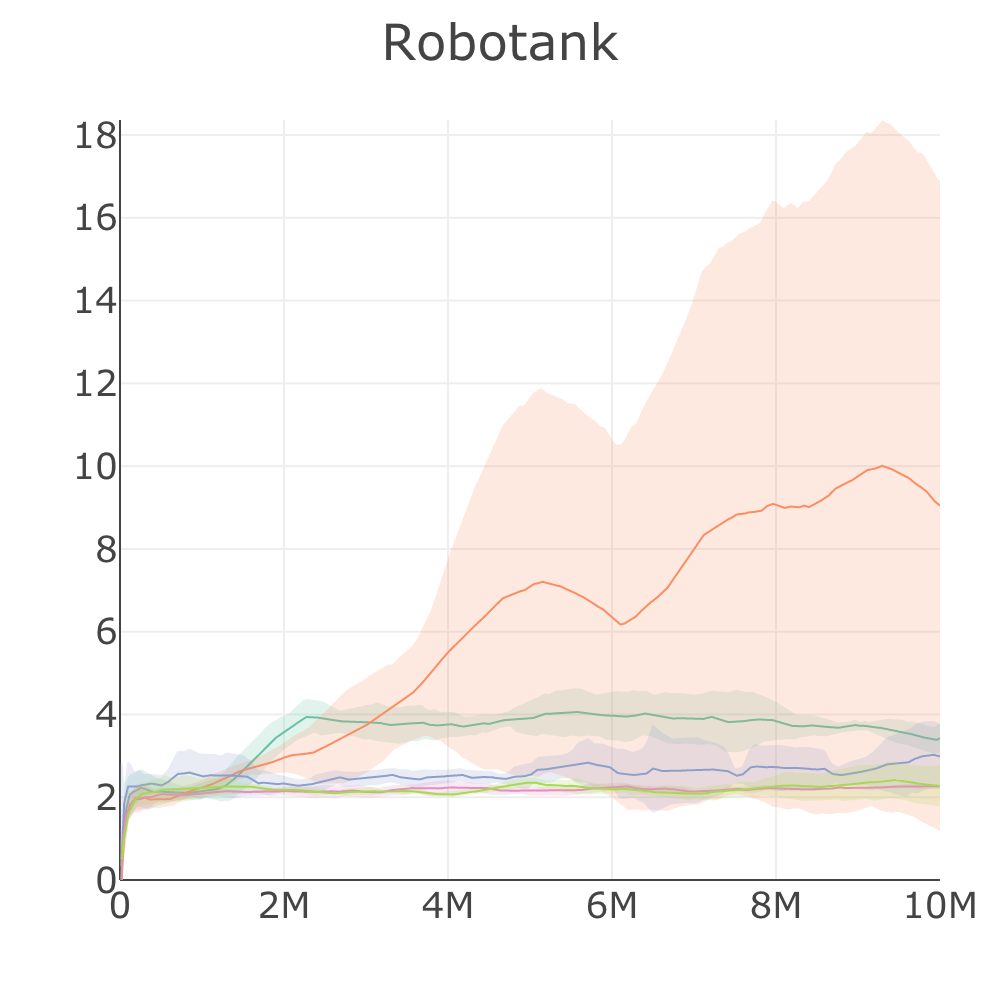}\\
\includegraphics[width=1.22in]{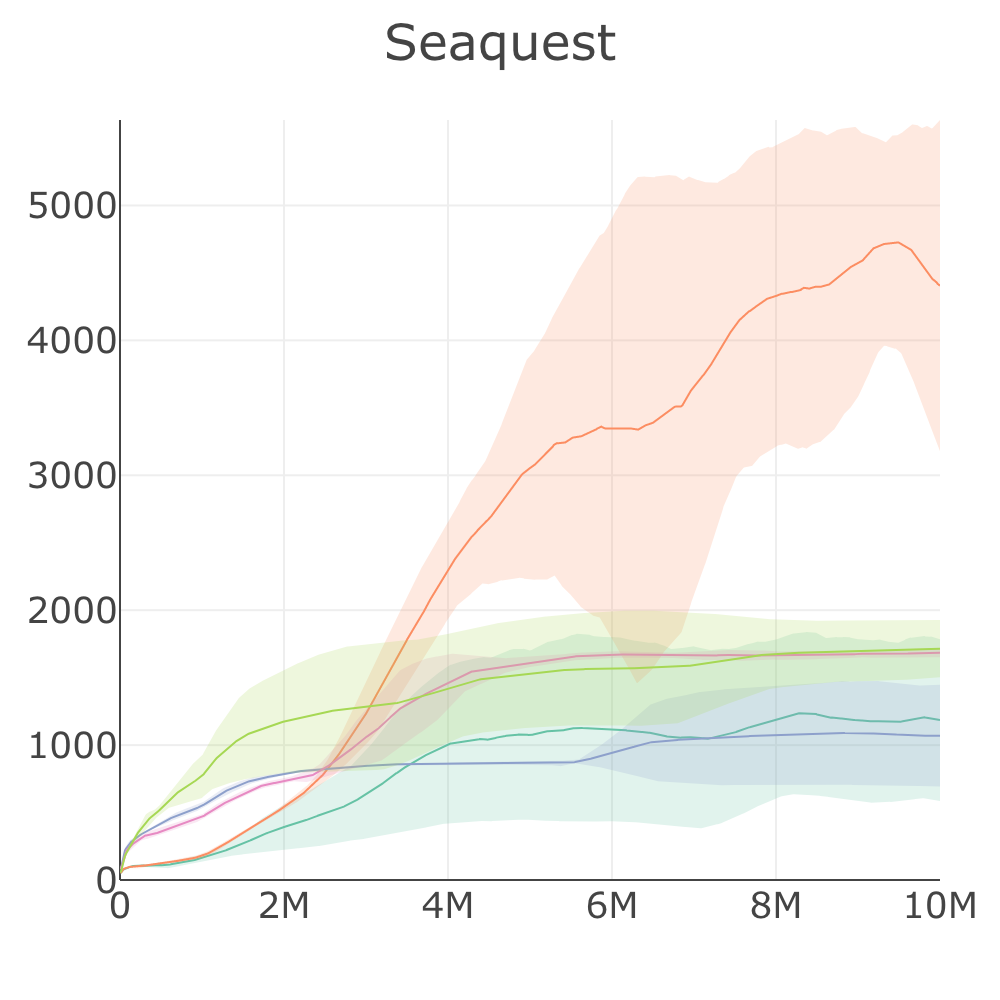} & 
\includegraphics[width=1.22in]{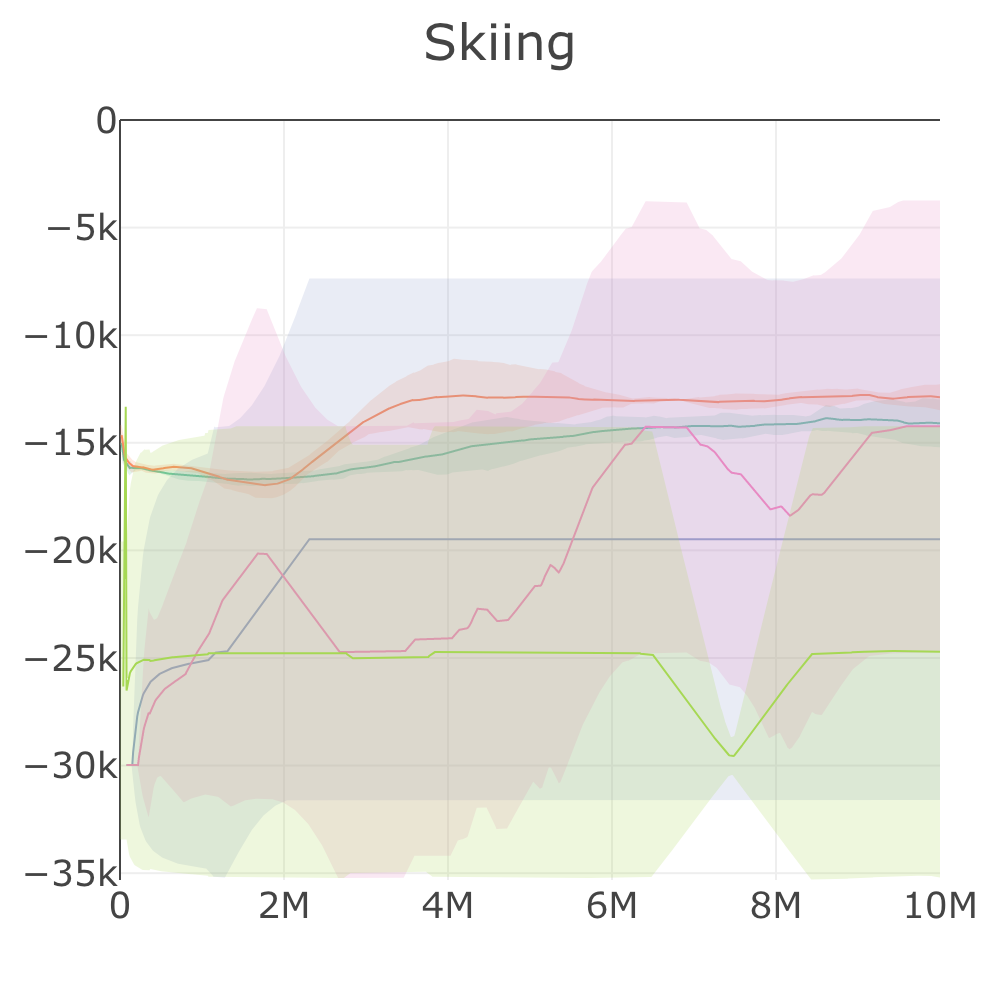} & 
\includegraphics[width=1.22in]{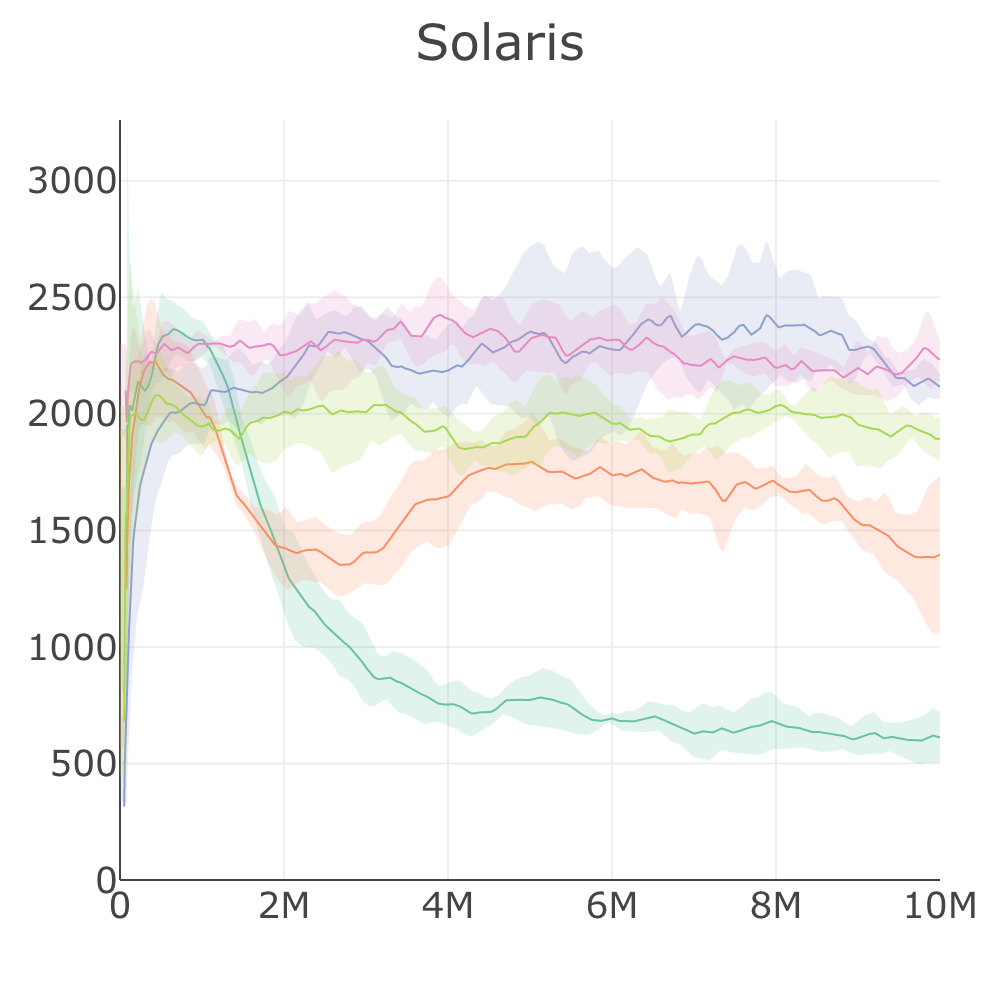} & 
\includegraphics[width=1.22in]{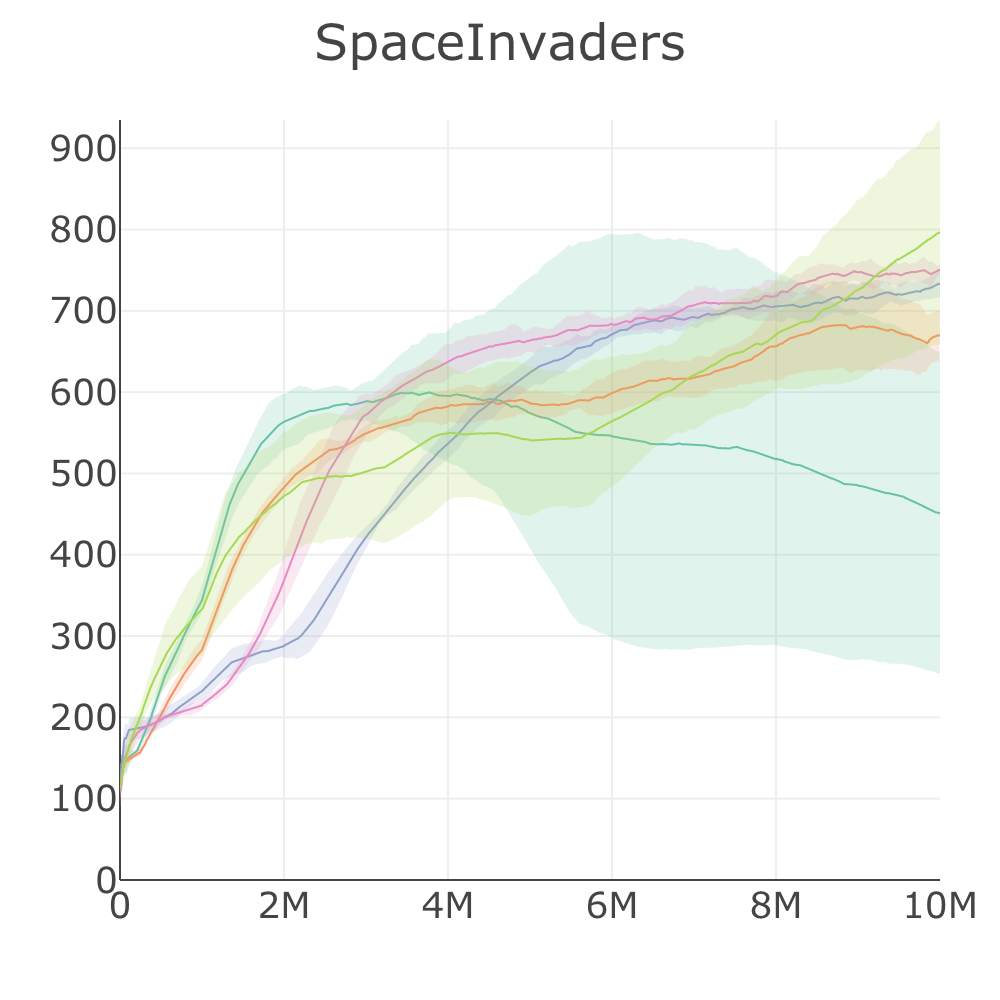}\\
\includegraphics[width=1.22in]{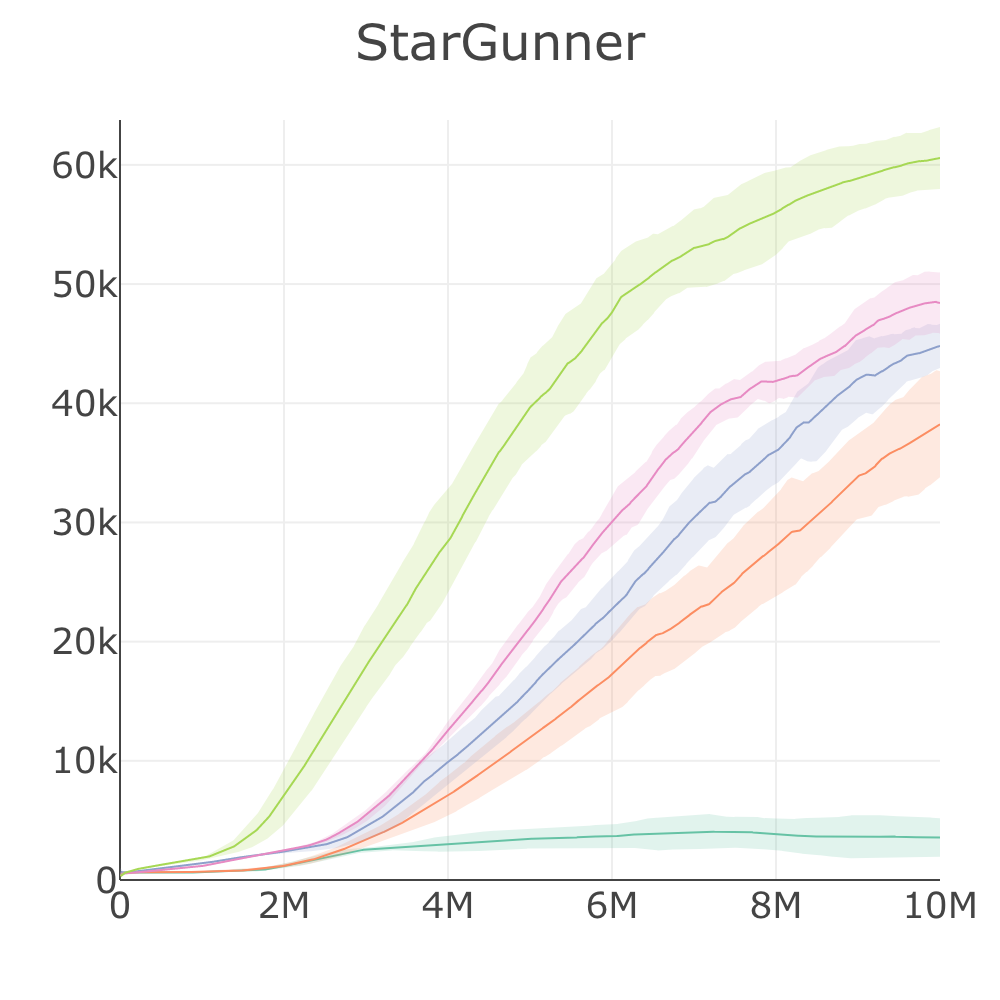} & 
\includegraphics[width=1.22in]{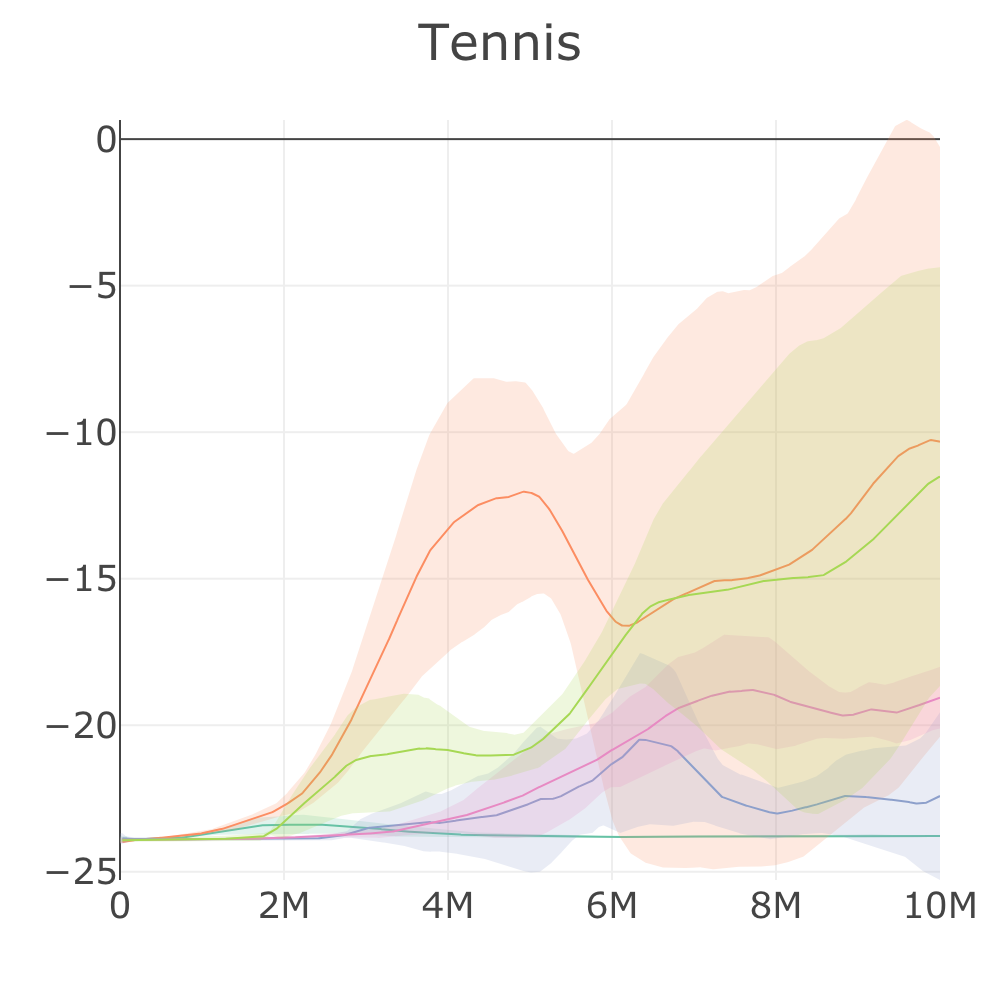} & 
\includegraphics[width=1.22in]{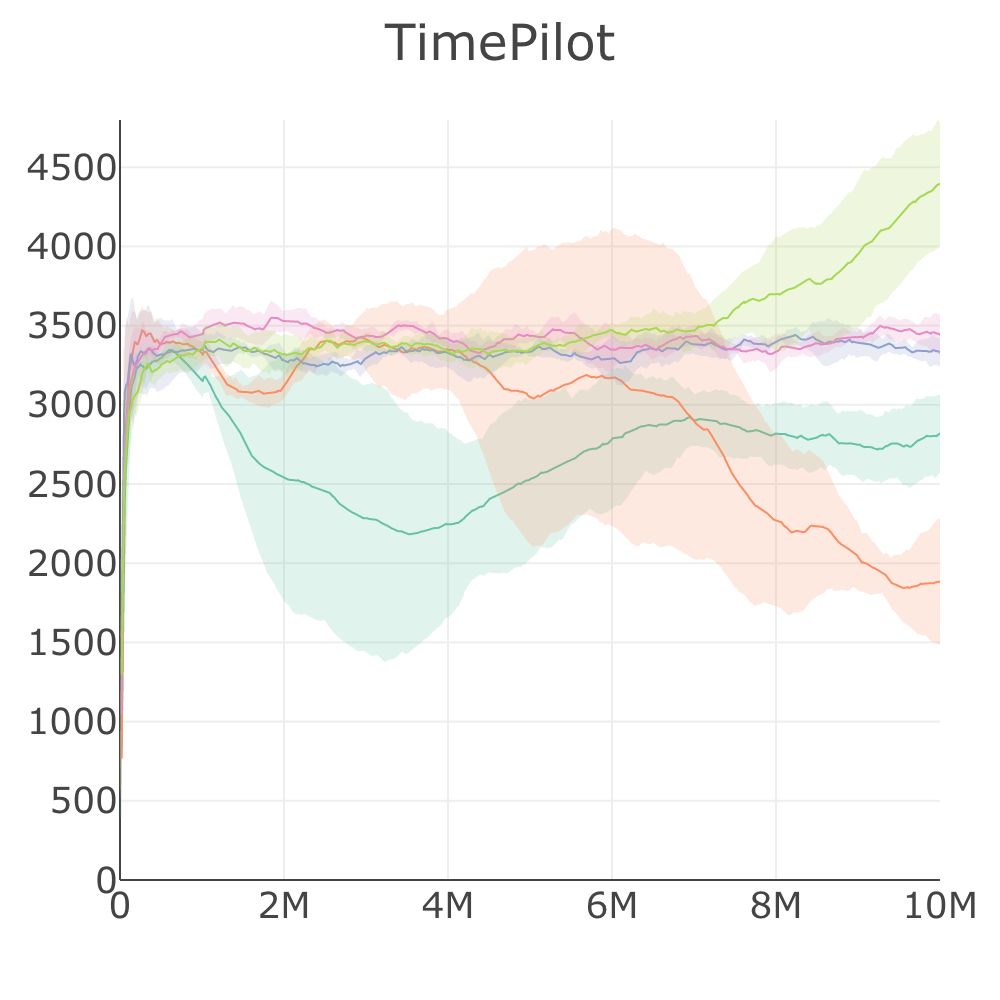} & 
\includegraphics[width=1.22in]{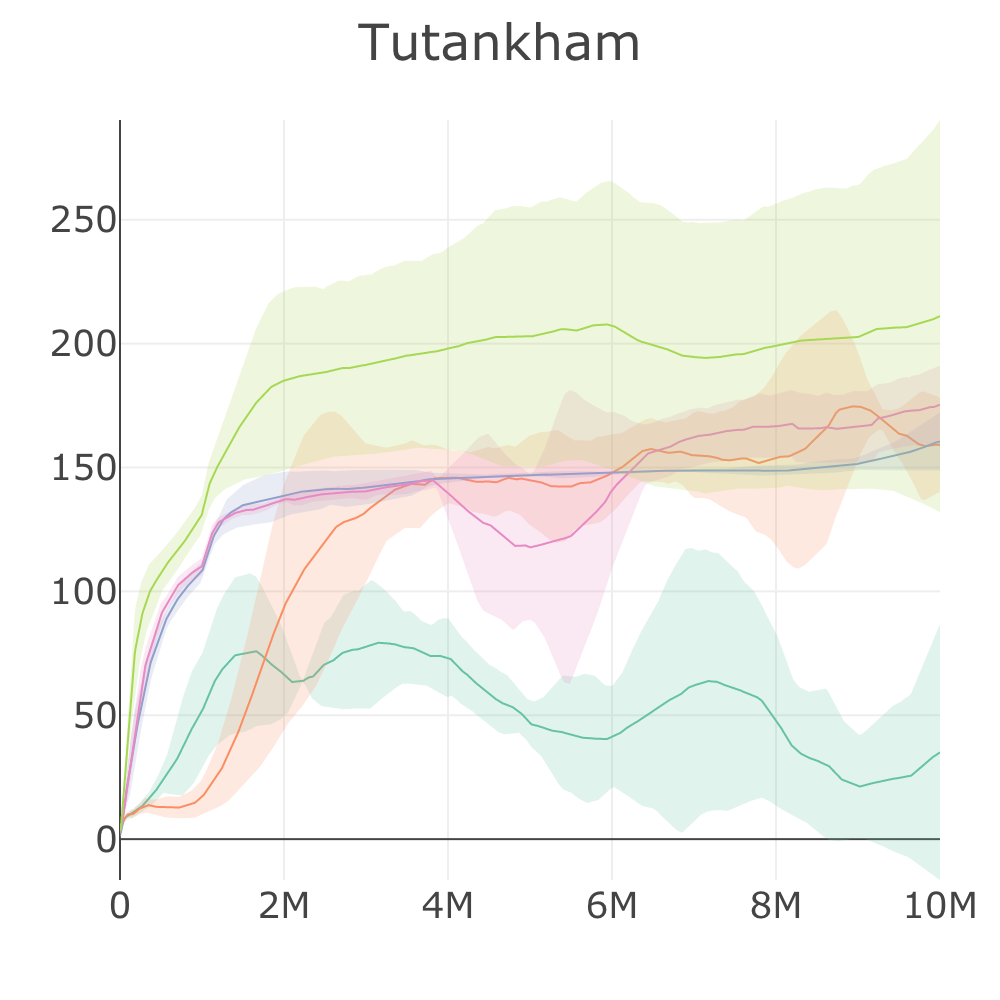}\\
\includegraphics[width=1.22in]{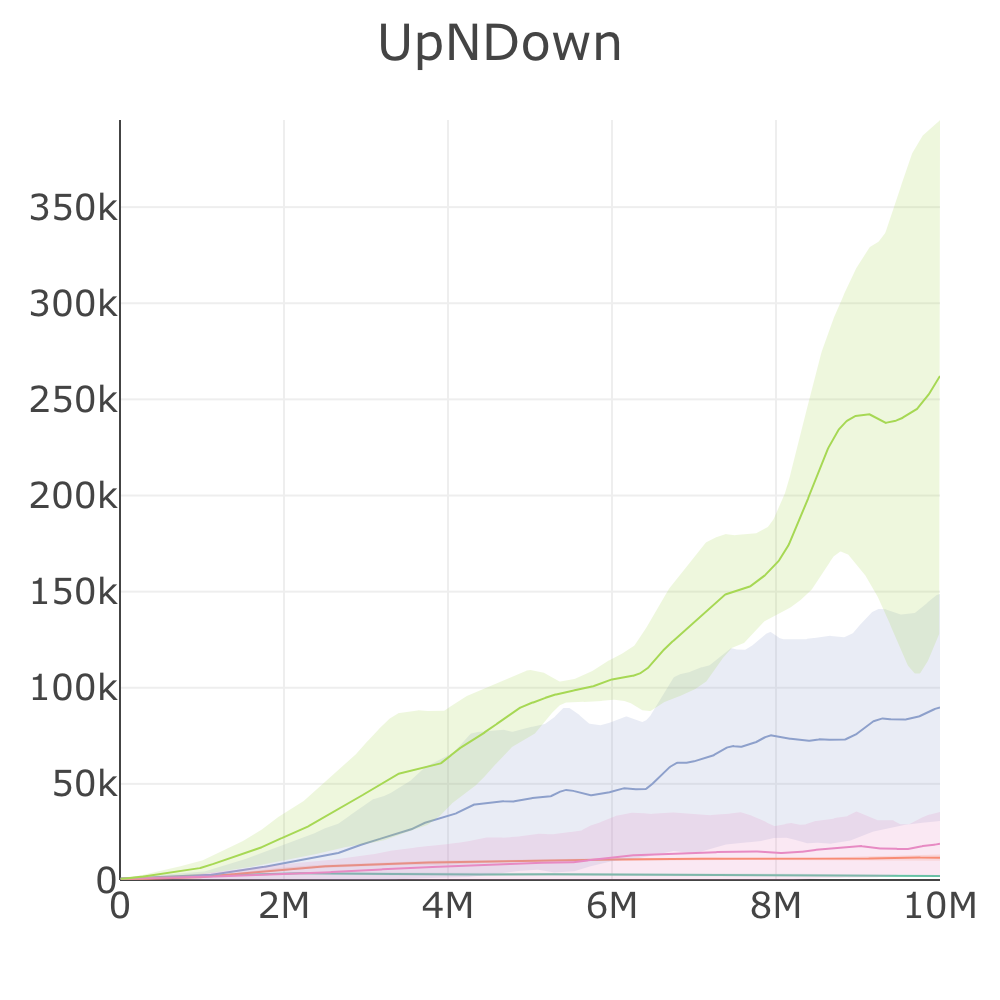} & 
\includegraphics[width=1.22in]{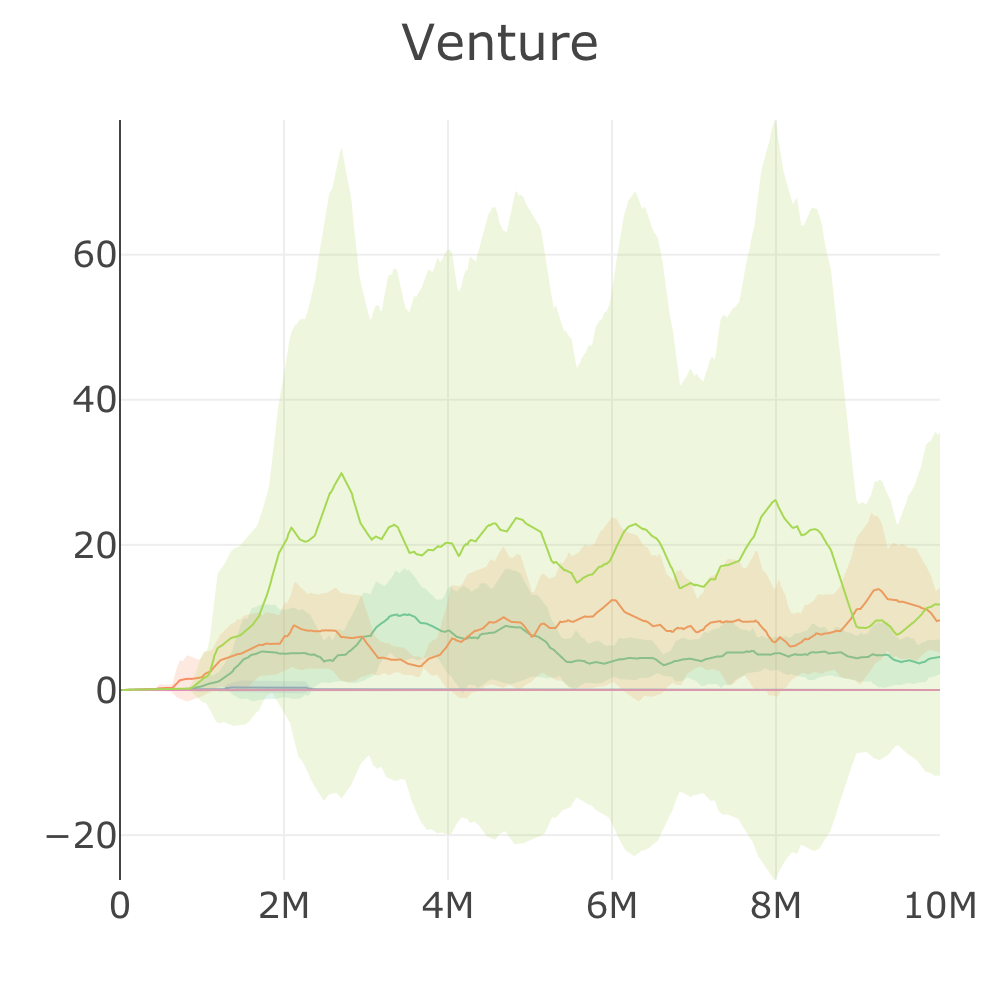} & 
\includegraphics[width=1.22in]{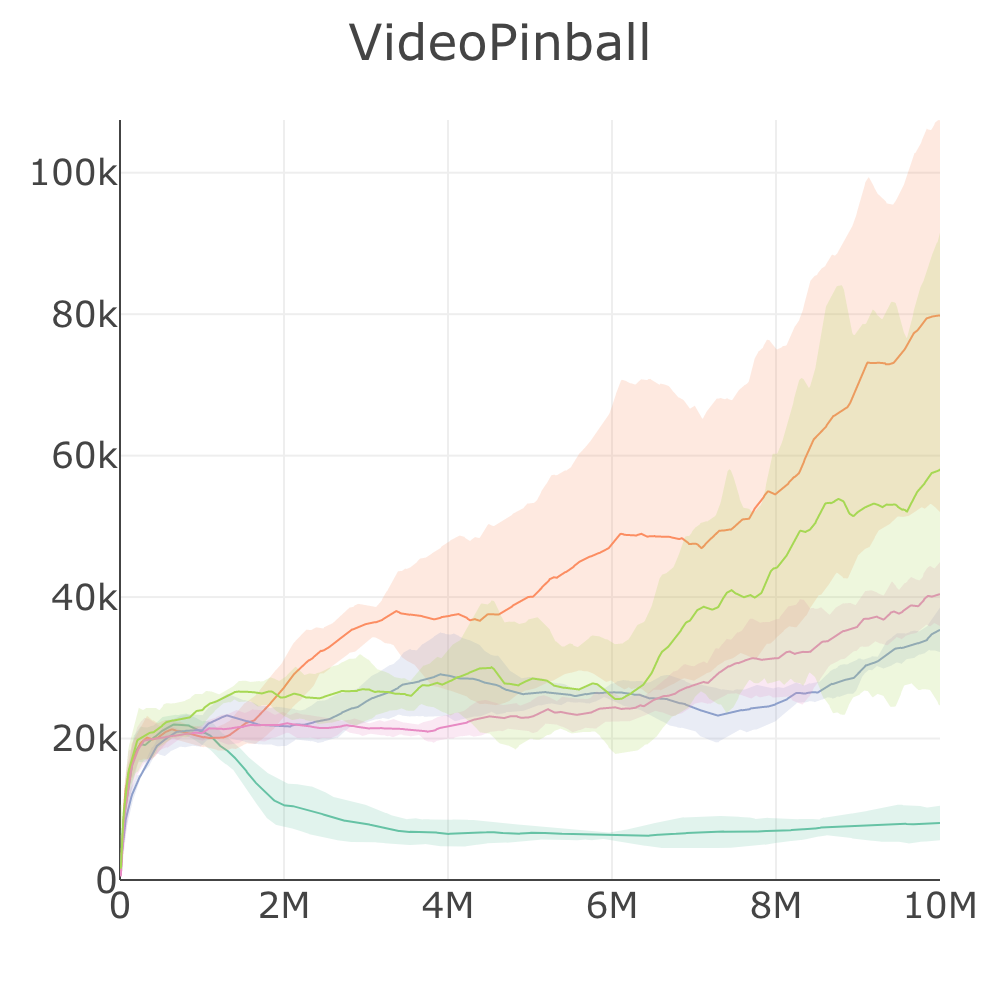} & 
\includegraphics[width=1.22in]{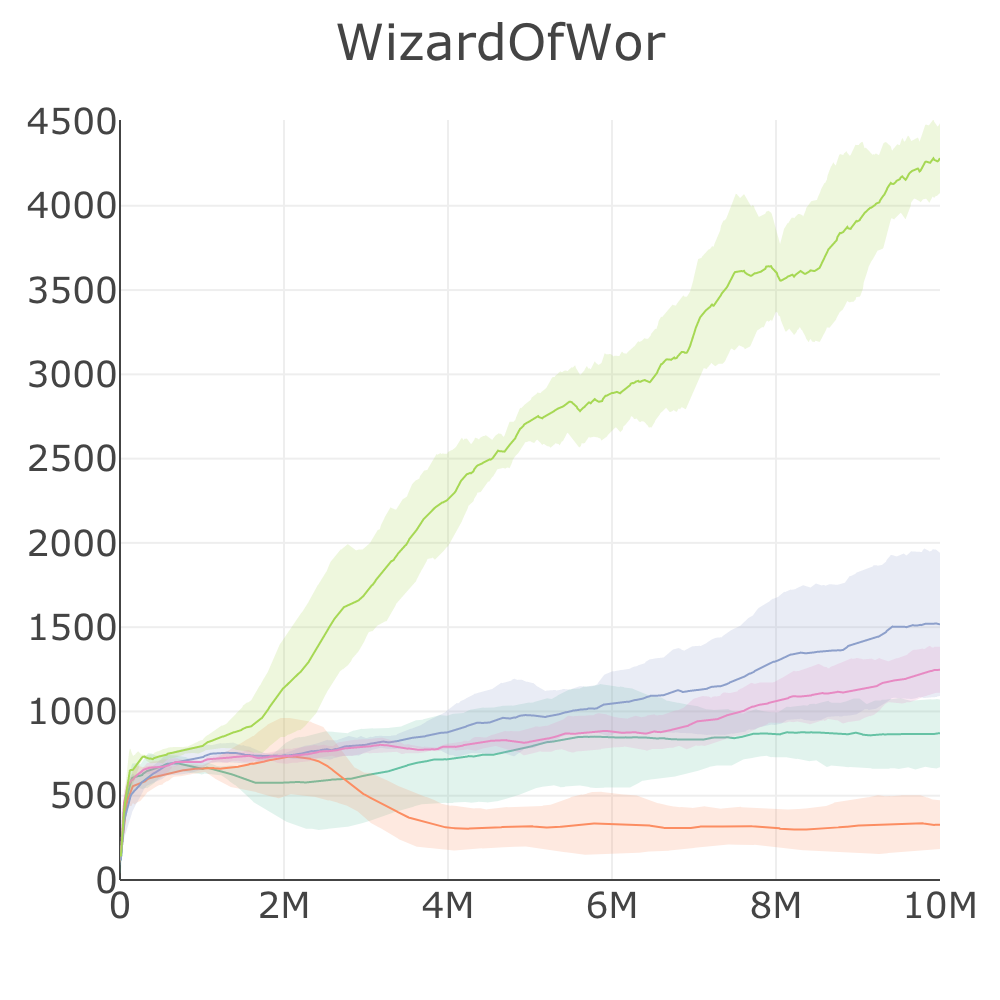}\\
\includegraphics[width=1.22in]{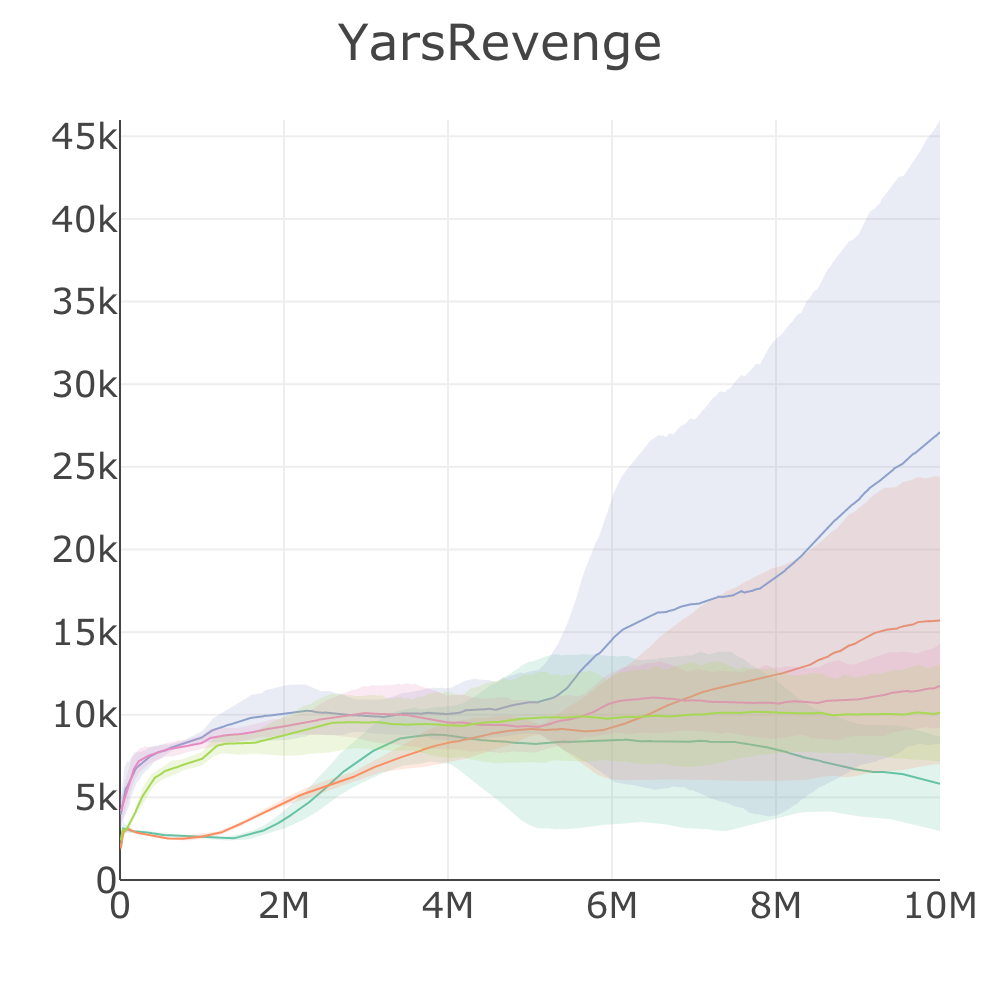} & 
\includegraphics[width=1.22in]{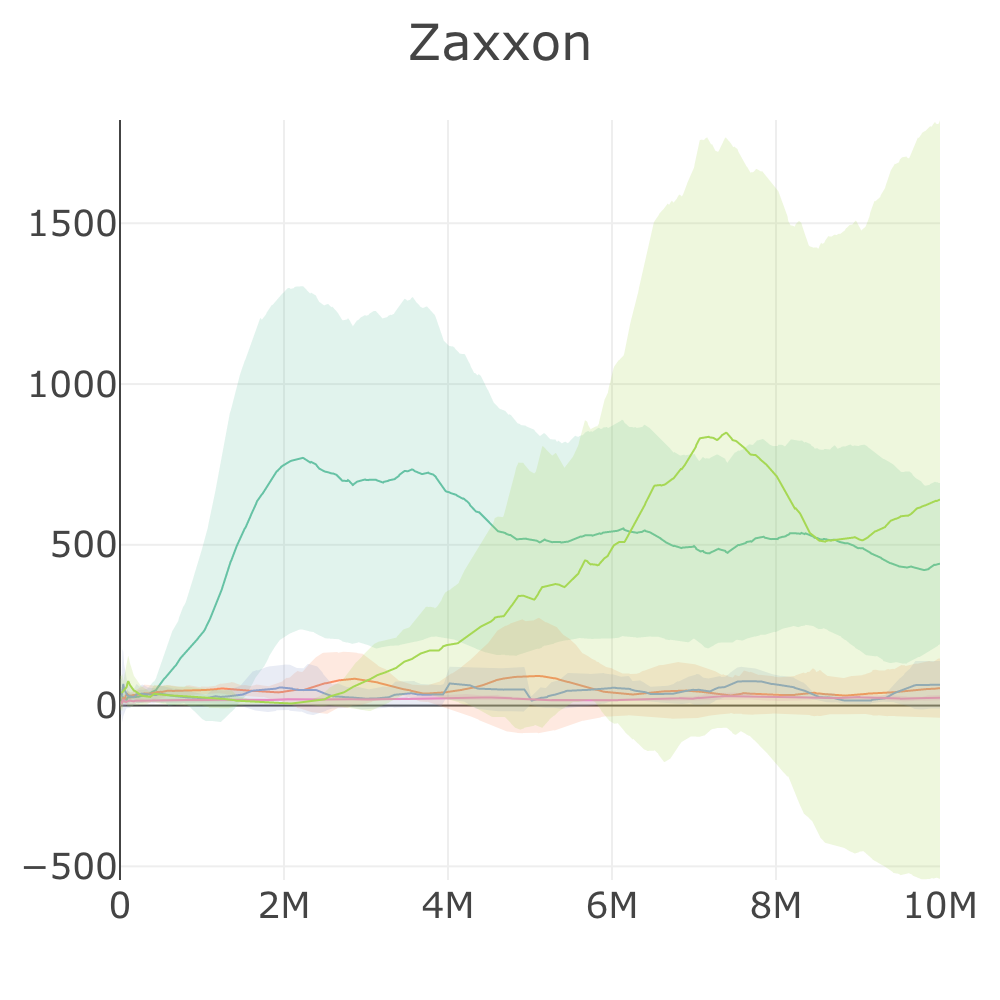}
\end{longtable}

\subsection{Example Spec files}\label{sec:sup_spec}

\begin{lstlisting}[frame=lines,caption=The Double DQN + PER spec file for Atari games]
{
  "ddqn_per_atari": {
    "agent": [{
      "name": "DoubleDQN",
      "algorithm": {
        "name": "DoubleDQN",
        "action_pdtype": "Argmax",
        "action_policy": "epsilon_greedy",
        "explore_var_spec": {
          "name": "linear_decay",
          "start_val": 1.0,
          "end_val": 0.01,
          "start_step": 10000,
          "end_step": 1000000
        },
        "gamma": 0.99,
        "training_batch_iter": 1,
        "training_iter": 4,
        "training_frequency": 4,
        "training_start_step": 10000
      },
      "memory": {
        "name": "PrioritizedReplay",
        "alpha": 0.6,
        "epsilon": 0.0001,
        "batch_size": 32,
        "max_size": 200000,
        "use_cer": false,
      },
      "net": {
        "type": "ConvNet",
        "conv_hid_layers": [
          [32, 8, 4, 0, 1],
          [64, 4, 2, 0, 1],
          [64, 3, 1, 0, 1]
        ],
        "fc_hid_layers": [256],
        "hid_layers_activation": "relu",
        "init_fn": null,
        "batch_norm": false,
        "clip_grad_val": 10.0,
        "loss_spec": {
          "name": "SmoothL1Loss"
        },
        "optim_spec": {
          "name": "Adam",
          "lr": 2.5e-5,
        },
        "lr_scheduler_spec": null,
        "update_type": "replace",
        "update_frequency": 1000,
        "gpu": true
      }
    }],
    "env": [{
      "name": "${env}",
      "frame_op": "concat",
      "frame_op_len": 4,
      "reward_scale": "sign",
      "num_envs": 16,
      "max_t": null,
      "max_frame": 1e7
    }],
    "body": {
      "product": "outer",
      "num": 1
    },
    "meta": {
      "distributed": false,
      "eval_frequency": 10000,
      "log_frequency": 10000,
      "rigorous_eval": 0,
      "max_session": 4,
      "max_trial": 1
    },
    "spec_params": {
      "env": [
        "BreakoutNoFrameskip-v4", "PongNoFrameskip-v4", "QbertNoFrameskip-v4", "SeaquestNoFrameskip-v4"
      ]
    }
  },
}
\end{lstlisting}

\begin{lstlisting}[frame=lines,caption=The PPO spec file for Atari games]
{
  "ppo_atari": {
    "agent": [{
      "name": "PPO",
      "algorithm": {
        "name": "PPO",
        "action_pdtype": "default",
        "action_policy": "default",
        "explore_var_spec": null,
        "gamma": 0.99,
        "lam": 0.70,
        "clip_eps_spec": {
          "name": "no_decay",
          "start_val": 0.10,
          "end_val": 0.10,
          "start_step": 0,
          "end_step": 0
        },
        "entropy_coef_spec": {
          "name": "no_decay",
          "start_val": 0.01,
          "end_val": 0.01,
          "start_step": 0,
          "end_step": 0
        },
        "val_loss_coef": 0.5,
        "time_horizon": 128,
        "minibatch_size": 256,
        "training_epoch": 4
      },
      "memory": {
        "name": "OnPolicyBatchReplay",
      },
      "net": {
        "type": "ConvNet",
        "shared": true,
        "conv_hid_layers": [
          [32, 8, 4, 0, 1],
          [64, 4, 2, 0, 1],
          [32, 3, 1, 0, 1]
        ],
        "fc_hid_layers": [512],
        "hid_layers_activation": "relu",
        "init_fn": "orthogonal_",
        "normalize": true,
        "batch_norm": false,
        "clip_grad_val": 0.5,
        "use_same_optim": false,
        "loss_spec": {
          "name": "MSELoss"
        },
        "actor_optim_spec": {
          "name": "Adam",
          "lr": 2.5e-4,
        },
        "critic_optim_spec": {
          "name": "Adam",
          "lr": 2.5e-4,
        },
        "lr_scheduler_spec": {
          "name": "LinearToZero",
          "frame": 1e7
        },
        "gpu": true
      }
    }],
    "env": [{
      "name": "${env}",
      "frame_op": "concat",
      "frame_op_len": 4,
      "reward_scale": "sign",
      "num_envs": 16,
      "max_t": null,
      "max_frame": 1e7
    }],
    "body": {
      "product": "outer",
      "num": 1
    },
    "meta": {
      "distributed": false,
      "eval_frequency": 10000,
      "log_frequency": 10000,
      "rigorous_eval": 0,
      "max_session": 4,
      "max_trial": 1
    },
    "spec_params": {
      "env": [
        "BreakoutNoFrameskip-v4", "PongNoFrameskip-v4", "QbertNoFrameskip-v4", "SeaquestNoFrameskip-v4"
      ]
    }
  },
}
\end{lstlisting}

\begin{lstlisting}[frame=lines,caption=The PPO spec file for Roboschool environments (excluding Humanoid)]
{
  "ppo_roboschool": {
    "agent": [{
      "name": "PPO",
      "algorithm": {
        "name": "PPO",
        "action_pdtype": "default",
        "action_policy": "default",
        "explore_var_spec": null,
        "gamma": 0.99,
        "lam": 0.95,
        "clip_eps_spec": {
          "name": "no_decay",
          "start_val": 0.20,
          "end_val": 0.20,
          "start_step": 0,
          "end_step": 0
        },
        "entropy_coef_spec": {
          "name": "no_decay",
          "start_val": 0.0,
          "end_val": 0.0,
          "start_step": 0,
          "end_step": 0
        },
        "val_loss_coef": 1.0,
        "time_horizon": 2048,
        "minibatch_size": 128,
        "training_epoch": 10
      },
      "memory": {
        "name": "OnPolicyBatchReplay",
      },
      "net": {
        "type": "MLPNet",
        "shared": false,
        "hid_layers": [256, 256],
        "hid_layers_activation": "relu",
        "init_fn": "orthogonal_",
        "clip_grad_val": 0.5,
        "use_same_optim": false,
        "loss_spec": {
          "name": "MSELoss"
        },
        "actor_optim_spec": {
          "name": "Adam",
          "lr": 3e-4,
        },
        "critic_optim_spec": {
          "name": "Adam",
          "lr": 3e-4,
        },
        "lr_scheduler_spec": null,
        "gpu": false
      }
    }],
    "env": [{
      "name": "${env}",
      "num_envs": 8,
      "max_t": null,
      "max_frame": 2e6
    }],
    "body": {
      "product": "outer",
      "num": 1
    },
    "meta": {
      "distributed": false,
      "log_frequency": 1000,
      "eval_frequency": 1000,
      "rigorous_eval": 0,
      "max_session": 4,
      "max_trial": 1
    },
    "spec_params": {
      "env": [
        "RoboschoolAnt-v1", "RoboschoolAtlasForwardWalk-v1", "RoboschoolHalfCheetah-v1", "RoboschoolHopper-v1", "RoboschoolInvertedDoublePendulum-v1", "RoboschoolInvertedPendulum-v1", "RoboschoolInvertedPendulumSwingup-v1", "RoboschoolReacher-v1", "RoboschoolWalker2d-v1"
      ]
    }
  },
}
\end{lstlisting}

\subsection{Key Implementation Lessons}

Implementing a number of well-known deep reinforcement learning algorithms taught us many important lessons. Such lessons are typically relegated to blog posts and personal conversations, but we feel it is worth sharing such knowledge in the research literature.

\begin{itemize}
    \item Put \textit{\textbf{everything}} that varies into the spec file. e.g. when implementing A2C we hardcoded state normalization in some places but not others. This extended the debugging process by days.
    \item Log policy entropy, $Q$ or $V$ function outputs and check that they change. Looking at $Q$ function values for a random agent helped us to track down an error due to state mutation.
    \item As with all Python code, watch out for mutation. For example, we noticed that the state returned from OpenAI baselines vector environment was mutated by preprocessing for the next time steps. This was because the state was tracked as internal variables and returned directly. To remedy this, the state must be returned as a copy.
    \item Start simple - implement parent algorithms first. e.g. REINFORCE, then A2C, then PPO or SAC, and ensure that each of these work in turn. PPO and SAC were straightforward to debug once A2C was working.
    \item Write tests: Tests have been invaluable in debugging and growing SLM Lab. 
    Especially useful are tests for tricky functions. For example, a test for the GAE calculation was essential for debugging A2C. Additionally, use Continuous Integration to automatically build and test every new code changes in a Pull Request, as is done in SLM Lab.
    \item Check input and output tensor shapes for important calculations. e.g. predicted and target values for Q or V. Broadcasting errors may lead to incorrect tensor shapes which fail silently.
    \item Check the size of the loss at the beginning of training. Is this comparable to other implementations? Tracking this value was essential when debugging DQN. It led us to a bug in the image permutation and we also stopped normalizing the image values as a result.
    \item Check that the main components of the computation graph are connected as expected. This can be implemented as a decorator around the training step and run when debugging. Conversely, be careful to detach target values.
    \item Visualize trained agents. We forgot to evaluate agents with reward clipping turned off, which led to very significant underestimation of performance for all environments except Pong. We finally realized training was working only after deciding to look at some policies playing the game.
\end{itemize}

\end{document}

%% file: math_commands.tex
\usepackage{amsmath,amsfonts,bm}

\def\eqref#1{equation~\ref{#1}}

\def\1{\bm{1}}

\DeclareMathAlphabet{\mathsfit}{\encodingdefault}{\sfdefault}{m}{sl}
\SetMathAlphabet{\mathsfit}{bold}{\encodingdefault}{\sfdefault}{bx}{n}

